\documentclass{article}
\usepackage[ruled,vlined,algo2e]{algorithm2e}
\usepackage[utf8]{inputenc}
\usepackage{amsfonts}
\usepackage{amssymb}
\usepackage{microtype}
\usepackage{graphicx}
\usepackage{booktabs} %
\usepackage{multirow}
\usepackage{rotating}

\usepackage[accepted]{icml2020}

\usepackage[utf8]{inputenc} %
\usepackage[T1]{fontenc}    %
\usepackage{hyperref}       %
\usepackage{url}            %
\usepackage{booktabs}       %
\usepackage{amsfonts}       %
\usepackage{nicefrac}       %
\usepackage{microtype}      %
\usepackage{graphicx}      
\usepackage{amsmath,amsthm,amssymb} 
\usepackage{dsfont}
\usepackage{float,adjustbox}
\usepackage{wrapfig,color}
\usepackage{subcaption}

\newcommand{\gan}{{InfoGAN-CR }}
\newcommand{\gannosp}{{InfoGAN-CR}}
\newcommand{\cL}{{\cal L}}

\newcommand{\cX}{{\cal X}}
\newcommand{\cQ}{{\cal Q}}
\newcommand{\reals}{{\mathbb R}}

\newcommand{\E}{{\mathbb E}}
\newcommand{\ones}{{\mathds 1}}
\newcommand{\id}{{\mathbb I}}
\newcommand{\Lcon}{\cL_{\rm c}}

\newtheorem{propo}{Proposition}

\newtheorem{thm}[propo]{Theorem}

\newtheorem{remark}[propo]{Remark}

\newcommand{\redca}[1]{\textcolor{black}{#1}}
\newcommand{\red}[1]{\textcolor{black}{#1}}

\newcommand{\dsprites}{dSprites}
\newcommand{\teapots}{3DTeapots}
\newcommand{\cdspriteslong}{Circular \dsprites{}}
\newcommand{\cdsprites}{C\dsprites{}}

\icmltitlerunning{Self-supervised Model Training and Selection for Disentangling GANs}

\begin{document}

\twocolumn[
\icmltitle{InfoGAN-CR and ModelCentrality: Self-supervised Model Training and Selection for Disentangling GANs}

\icmlsetsymbol{equal}{*}

\begin{icmlauthorlist}
\icmlauthor{Zinan Lin}{cmu}
\icmlauthor{Kiran K. Thekumparampil}{uiuc}
\icmlauthor{Giulia Fanti}{cmu}
\icmlauthor{Sewoong Oh}{uw}
\end{icmlauthorlist}

\icmlaffiliation{cmu}{Carnegie Mellon University}
\icmlaffiliation{uiuc}{University of Illinois at Urbana-Champaign}
\icmlaffiliation{uw}{University of Washington}
\icmlcorrespondingauthor{Zinan Lin}{zinanl@andrew.cmu.edu}
\icmlcorrespondingauthor{Kiran K. Thekumparampil}{thekump2@illinois.edu}
\icmlcorrespondingauthor{Giulia Fanti}{gfanti@andrew.cmu.edu}
\icmlcorrespondingauthor{Sewoong Oh}{sewoong@cs.washington.edu}

\icmlkeywords{Machine Learning, ICML}

\vskip 0.3in
]

\printAffiliationsAndNotice{}  %

\begin{abstract}

Disentangled generative models map a latent code vector to a target space, while enforcing that
a subset of the learned latent codes are interpretable and associated with distinct properties of the target distribution. 
Recent advances  have been dominated by Variational AutoEncoder (VAE)-based methods, while training disentangled generative adversarial networks (GANs) remains challenging. 
In this work, we show that the dominant challenges facing disentangled GANs can be mitigated through the use of self-supervision. 
We make two main contributions: 
first, we design a novel approach for training disentangled GANs with self-supervision. 
We propose {\em contrastive regularizer}, which is  inspired by a natural notion of disentanglement: latent traversal. 
This achieves higher disentanglement scores than state-of-the-art VAE- and GAN-based approaches.
Second, we propose an unsupervised model selection scheme called ModelCentrality, which uses generated synthetic samples to compute the medoid (multi-dimensional generalization of  median) of a collection of models. 
\redca{
The current common practice of hyper-parameter tuning 
requires using ground-truths samples, each labelled with known perfect disentangled latent codes. 
As real datasets are not equipped with such labels, we propose an unsupervised model selection scheme 
and  show that it finds  a model close to the best one, for both VAEs and GANs. 
Combining contrastive regularization with ModelCentrality, we improve upon the state-of-the-art disentanglement scores significantly, without accessing the supervised data.}

\end{abstract}

\section{Introduction}
\label{sec:intro}

The ability to learn low-dimensional, informative data representations
can greatly enhance the utility of data. 
 The notion of \emph{disentangled} representations in particular was  theoretically proposed in 
\cite{bengio2013representation,Rid16,HMP16} for diverse applications including supervised and reinforcement learning. 
A disentangled generative model takes a number of latent factors as inputs, with each factor controlling 
an interpretable aspect of the generated data.  
For example, in facial images, disentangled latent factors might control variations in eyes, noses, and hair. 

\begin{figure}[ht]
	\centering
	\includegraphics[width=0.23\textwidth]{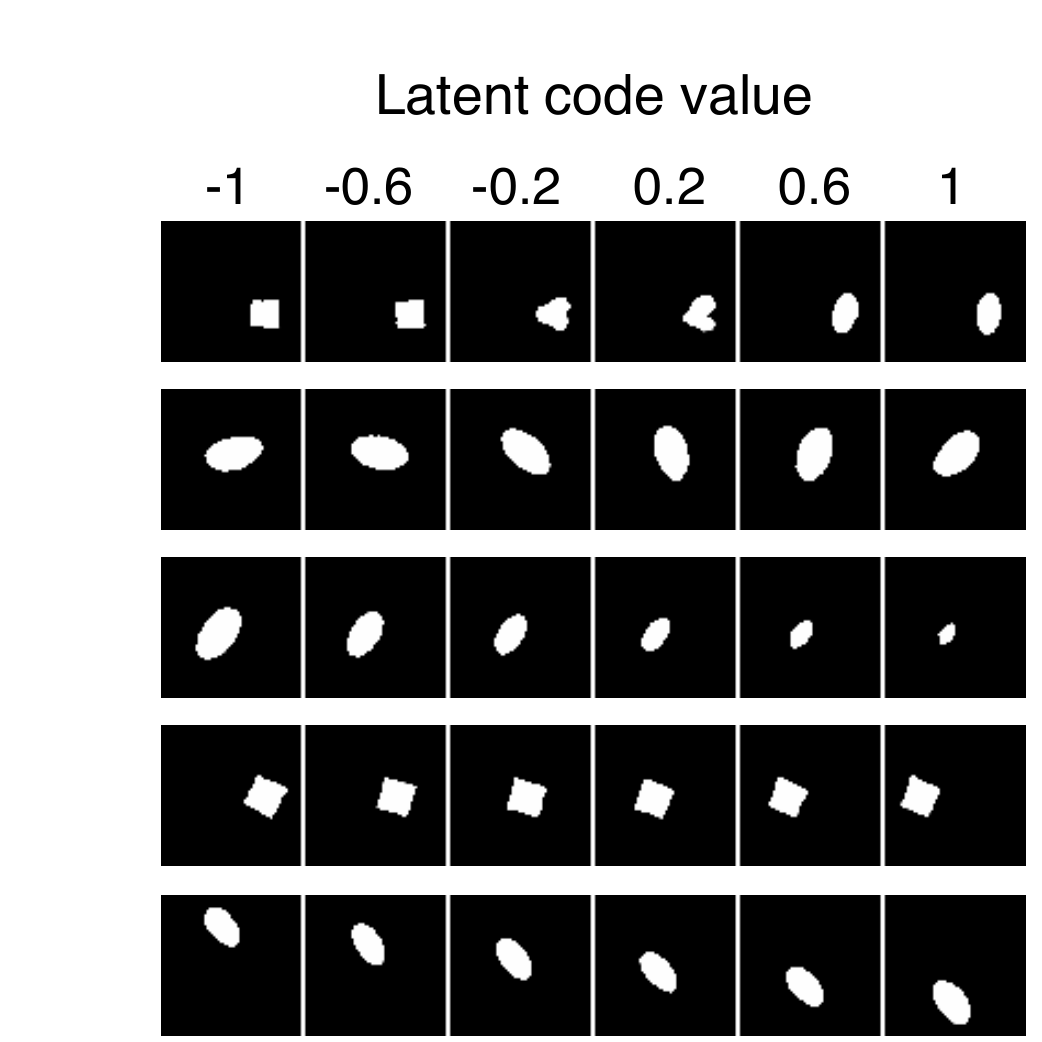}
	\put(-108,79){$c_1$}
	\put(-108,59){$c_2$}
	\put(-108,41){$c_3$}
	\put(-108,24){$c_4$}
	\put(-108,6){$c_5$}
	\put(2,79){shape}
	\put(2,59){rotation}
	\put(2,41){scale}
	\put(2,24){$x$-position}
	\put(2,6){$y$-position}
  \caption{Each row shows how the image changes 
	when traversing a single latent code under the proposed \gan architecture (\dsprites{} dataset, \S~\ref{sec:CRexp}). 
	Latent codes capture desired properties: \{shape, rotation, scale, x-pos, y-pos\}, of the image.  
	}
	\label{fig:latenttraversal}
\end{figure}
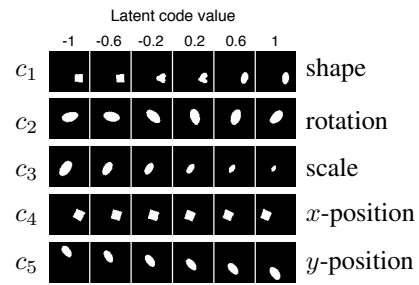
Most approaches for disentangling latent factors (or \emph{codes})
are based on the following natural intuition. 
We say a generative model has a better disentanglement if 
changing one latent code (while fixing other latent codes) 
makes a 
{\em noticeable} %
and {\em distinct} %
change in the generated sample (referred to as  ``informativeness'' and ``disentanglement'' in \cite{EW18}).   
Noticeable changes are desired as we want the latent codes to capture 
important characteristics of the image. 
Distinct changes are desired as we want each latent code to represent an aspect of the samples different from other latent codes.
\cite{EW18} also values ``completeness'', which refers to how much of the disentangling latent factors are covered by the learned model. 
As such,  disentanglement can be evaluated by traversing the latent space as in Figure \ref{fig:latenttraversal}: 
by fixing all latent codes except one, varying that code, and visualizing the resulting changes. 
Figure~\ref{fig:latenttraversal} illustrates how the latent codes $\{c_1,\ldots ,c_5\}$ of a successfully-trained generator 
capture noticeable and 
 distinct properties of the images.

Two main obstacles arise in the design of disentangled generative models:
(1) designing architectures that achieve good disentanglement \emph{and} good sample quality, and
(2) hyperparameter tuning and model selection given a fixed learning architecture. 

For the first problem, recent approaches to disentanglement have focused on adding carefully 
chosen regularizers to promote disentanglement, 
building upon the two popular deep generative models: 
Variational AutoEncoders (VAE) \cite{KW13} and 
Generative Adversarial Networks (GAN) \cite{GPM14}. 
Fundamental differences in these  two architectures led to 
the design of different regularizers. 
To achieve disentanglement in VAEs, a popular approach is to promote ``{\em uncorrelatedness}'' by regularizing with total correlation, as in $\beta$-VAE  and FactorVAE \cite{HMP16,KM18}. %
This approach has led to successful disentanglement scores, albeit at the cost of sample quality.
Disentangled GANs, on the other hand, add a secondary input of latent codes, which are meant to control  the underlying factors.
The loss function  then adds an extra regularizer to promote 
``{\em informativeness}'',  as proposed in  InfoGAN \cite{CDH16}. 
Despite improving sample quality, InfoGAN has lower disentanglement scores than its VAE-based counterparts,
which led to slow progress on GAN-based disentangled representation learning. 

The second problem, model selection, has received relatively less attention. 
Most prior work on disentanglement conducts hyperparameter tuning by cross-validating on a holdout dataset labelled with ground truth latent codes.
This significantly limits the validity of those training methods on real  datasets with unknown labels.
However, recent work has acknowledged the need for unsupervised model selection techniques and proposed an unsupervised approach \cite{duan2019heuristic}.
This approach was evaluated only on VAEs, and as we will show, it has poor performance on GAN-based models. 

In summary, the two principal challenges associated with the design of disentangled generative models are particularly pronounced for disentangled GANs.
This has contributed to a perception in the community that GANs are less well-suited to learning disentangled representations.

\noindent{\bf Main contributions.} 
In this paper, we show that self-supervision can mitigate both of these challenges for disentangled GANs, allowing their performance to far supersede state-of-the-art VAE-based methods. 
We make two primary contributions:

First, we design a novel architecture for training disentangled GANs, which we call InfoGAN-CR. 
InfoGAN-CR adds a {\em contrastive regularizer} (CR)  
that combines self-supervision with the most natural measure of disentanglement: latent traversal. 
\textcolor{black}{We create a self-supervised learning task of 
multi-way hypothesis tests over the latent codes and encouraging the generator to succeed at those tasks.} 
We provide experimental results showing that it achieves state-of-the-art disentanglement scores on benchmark tasks.

Second, we introduce a novel model selection scheme based on self-supervision, which we call {\em ModelCentrality}.
This builds upon a premise that well-disentangled models are close together, with the closeness measured by a popular disentanglement metric from \cite{KM18}. 
We verify this premise numerically
and define ModelCentrality as the medoid (multi-dimensional generalization of the median) of a set of models, computed under this disentanglement metric. 
\textcolor{black}{ModelCentrality assigns centrality scores to each trained model based on 
the self-supervised labels defined by the closeness to other models }. 
We demonstrate on benchmark datasets that \redca{ModelCentrality can be used for selecting both disentanlged GANs and VAEs.}
 Models trained with InfoGAN-CR and selected with ModelCentrality significantly outperform 
state-of-the-art baseline approaches, even those  that use supervised hyper-parameter tuning.

\noindent{\bf Related work.}
Learning a disentangled representation 
 was first demonstrated in  the 
{\em semisupervised setting}, where 
additional annotated data is available.  
This consists of examples from desired isolated latent factor traversals \cite{karaletsos2015bayesian,kulkarni2015deep,NPV17,LRJ18,watters2019spatial,locatello2019disentangling,chen2019weakly}.
However, as manual data annotation is costly, {\em unsupervised methods} for disentangling are desired.
Early approaches to unsupervised disentangling 
imposed uncorrelatedness by making it difficult to predict 
one representational unit from the rest \cite{Sch92}, 
disentangling higher order moments \cite{DCB12}, using factor analysis \cite{TSH13}, and applying group representations  \cite{CW14}.   
Breakthroughs in making these ideas scalable were achieved 
by $\beta$-VAE \cite{HMP16} for VAE-based methods, and InfoGAN \cite{CDH16} for GAN-based ones. 
Rapid progress in improving disentanglement was driven mainly by VAE-based methods, 
in %
a series of papers \cite{KM18,LBL18,CLG18,LRJ18,CHYVAE,EWJ18,GBV18,PL18,Dup18,AFL18,AFF18,szabo2017challenges,burgess2018understanding,jeong2019learning,li2019learning,caselles2019symmetry,tschannen2018recent}. 
Quantitative comparisons in these papers 
suggest that InfoGAN learns poorly-disentangled representations. 
This has led to a misconception that GAN-based methods are inherently bad at learning disentangled representations.

{
Concurrent and subsequent to our work, several other GAN-based disentangling frameworks have been proposed~\cite{IBGAN,liu2019oogan,lee2020high} and these work corroborate our finding that VAE-based approaches are not superior in disentangling.
Additionaly, various domain specific models have also been proposed for structured data such as sequences~\cite{hsu2017unsupervised}, images~\cite{awiszus2019learning,lee2018diverse}, video~\cite{xing2018deformable,denton2017unsupervised,hsieh2018learning}, shapes~\cite{aumentado2019geometric,lorenz2019unsupervised}, and state space~\cite{miladinovic2019disentangled}.
Several works have studied the use of disentangled representations in diverse topics such as transfer learning~\cite{DARLA}, hierarchical visual concepts learning~\cite{SCAN}, visual reasoning~\cite{van2019disentangled}, fairness of learning~\cite{locatello2019fairness,marx2019disentangling,creager2019flexibly}, computer vision~\cite{lee2018diverse,hsieh2018learning,singh2019finegan}, speech processing~\cite{hsu2017unsupervised}, robust learning~\cite{duan2019disentangled}.
}

\section{Background} 
\label{sec:info}

In this section, we give a brief overview of GANs and InfoGAN,
 introduced in \cite{CDH16}.  

\noindent{\bf Background on GAN.}
Generative Adversarial Networks (GANs) 
are a breakthrough method for training generative models \cite{GPM14}. 
A deep neural network generative model maps a latent 
code $z\in\reals^d$ to a desired distribution of the samples $x=G(z)$. 
$z$ is typically drawn from a Gaussian distribution with identity covariance or a uniform distribution.
No likelihood is available for ML training of the neural network $G$. 
GANs instead update  weights of 
a generator $G$ and discriminator $D$ using alternative gradient updates on the following {\em adversarial  loss}: 
\begin{eqnarray}
	\min_G \;  \max_D \;\; \cL_{\rm Adv}(D,G) \;.
	\label{eq:GAN}
\end{eqnarray}
The discriminator provides an approximate measure 
of how different the current generator distribution is from the distribution of the real data. 
For example, a common choice is $\cL_{\rm Adv}(D,G)=\E_{x\sim P_{\rm real}}[\log(D(x))] + \E_{P_G}[\log(1-D(x))]$, 
which provides an approximation of the 
Jensen-Shannon divergence between the real data distribution $P_{\rm data}$ and the current generator distribution $P_G$.

\noindent{\bf Background on InfoGAN.}
In order to achieve disentanglement, 
 InfoGAN proposes a regularizer based on mutual information. 
As the goal is not to disentangle all  latent codes, 
but rather to disentangle a subset, 
InfoGAN  \cite{CDH16}  proposed to first split the latent codes into two parts: 
the disentangled code vector $c\in\reals^k$ and the remaining code vector $z\in\reals^d$ that provides  additional randomness. 
InfoGAN   then uses the GAN loss with regularization to 
encourage informative latent codes $c$:
\begin{eqnarray}
	\min_G \;  \max_D \;\; \cL_{\rm Adv}(G,D) - \lambda  \, I(c; G(c,z)) \;,
	\label{eq:InfoGAN}
\end{eqnarray}
where 
$I(c; G(c,z))$ denotes the mutual information between 
the latent code $c$ and the sample $G(c,z)$ generated from that latent code, 
and $\lambda$ is a positive scalar coefficient. 
Notice that encouraging informativeness alone does not necessarily imply good disentanglement; a fully entangled representation can achieve infinite mutual information $I(c; G(c,z))$.
Despite this,  InfoGAN achieves reasonable performance in practice. 
Its decent empirical performance follows from  implementation  choices that promote stability and alter the InfoGAN objective, which we discuss in Appendix~\ref{app:infogan}.

\section{Self-supervision with Contrastive Regularizer}
\label{sec:CR}

Our proposed regularizer is inspired by the idea that
disentanglement 
should be measured via changes in the images when traversing the latent space. 
This is a popular interpretation of disentanglement, 
as evidenced by the widely-adopted visual evaluations (e.g.~Figure~\ref{fig:latenttraversal}). 
This suggests a natural disentanglement approach: 
run latent traversal experiments and encourage 
models that make \emph{distinct} changes. 

We design a regularizer, which we call a {\em Contrastive Regularizer} (CR), based on this insight. 
That is, we generate two (or more) images 
from the generator, while fixing one of the latent codes $c_i$ to be the same for both images. We draw the rest of the latent codes uniformly at  random, and  
let $(x,x')\sim Q^{(i)}$ denote the resulting distribution of paired samples when factor $c_i$ is fixed. 
We propose measuring the distinctness of this latent traversal with Jensen-Shannon divergence among $Q^{(i)}$'s 
defined as 
	\begin{eqnarray}
		d_{\rm JS}(Q^{(1)}, \ldots, Q^{(k)}) \triangleq \frac1k \sum_{i \in [k]} d_{\rm KL} \big( Q^{(i)} \,\big|\big|\, \bar{Q} \big) \,,
		\label{eq:JSregularizer}
	\end{eqnarray}
	where $\bar{Q}= (1/k){\sum_{j \in [k]} Q^{(j)}} $.
	This measures how different each latent code traversal 
	is. 
If we maximize this as a regularizer to 
the generator training, 
in the subsequent generator update, 
the $Q^{(i)}$'s will be forced to be as different as possible. 
This, in turn, forces the changes in the latent codes to make changes in the images that are noticeable and easy to distinguish from (the changes of) other latent codes. 

In general different ways of {\em coupling} the latent space of the paired images can be used, and we leave it as a design choice. For example, one could fix the rest of the codes to be the same and randomly sample $c_i$. This fits a more traditional definition of latent traversal. 
We provide practical guidelines in \S\ref{sec:CRexp}. 

\begin{figure}[t]
	\centering
	\includegraphics[width=2.2in]{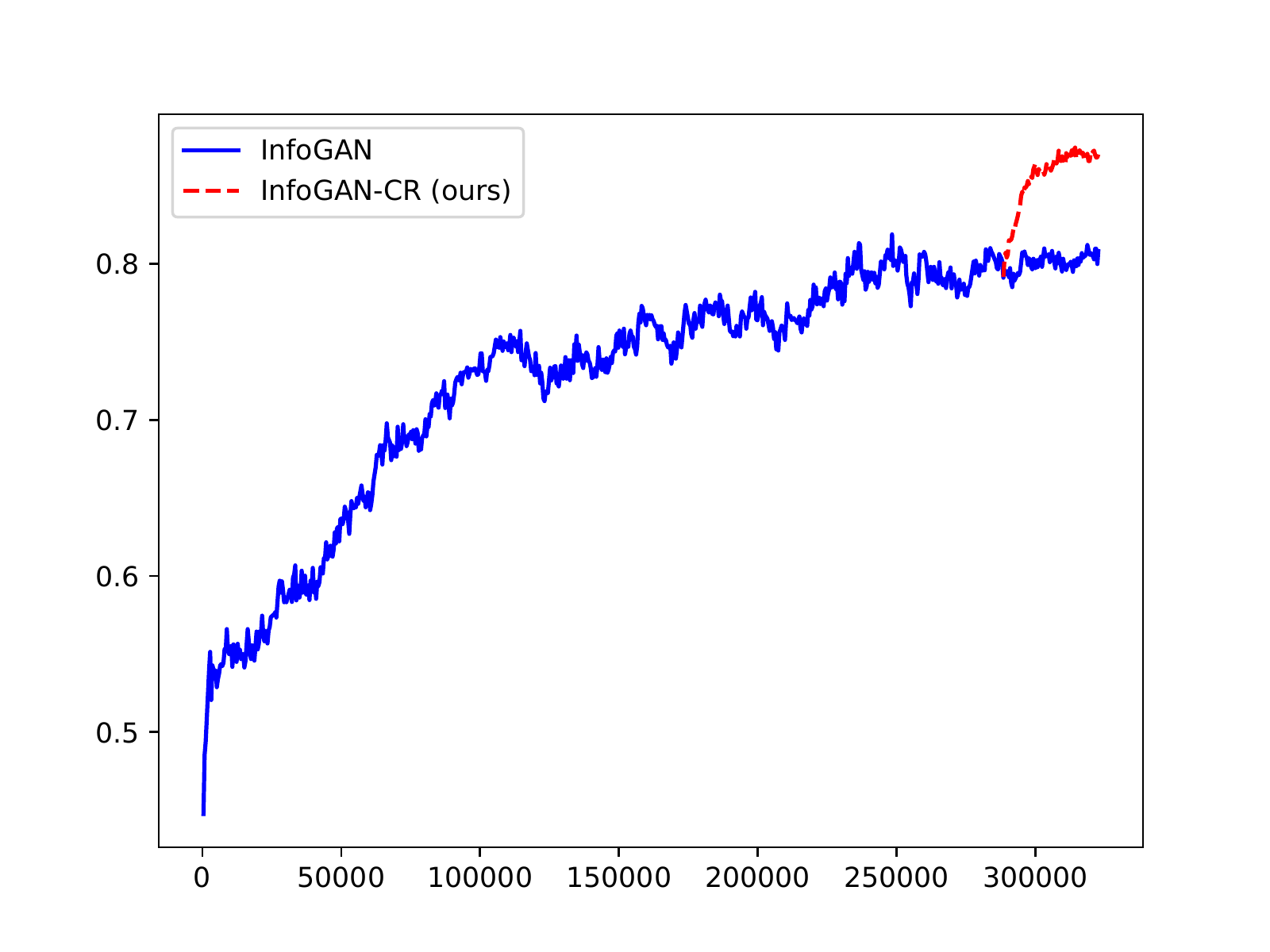}
	\put(-132,-5){Iterations (\# of minibatches)}
	\put(-145,113){FactorVAE disentanglement score}
	\caption{After 288,000 iterations, we continue training InfoGAN with(out) the proposed contrastive regularizer.  
	 The jump illustrates gains due to  CR regularization. Curves are averaged over 10 trials on the same data.}
	\label{fig:split}
\end{figure}

Before explaining how to implement our contrastive regularizer in \S\ref{sec:InfoGAN-CR}, we show in
Figure~\ref{fig:split} 
how it enhances disentanglement beyond vanilla InfoGAN. 
The blue curve shows the performance when we train a vanilla InfoGAN on the \dsprites{} dataset \cite{dSprites}  for 
28 epochs (322,560 iterations) total. %
To show the effect of the proposed CR regularizer, 
we take the model we just trained with InfoGAN at 
25 epochs (288,000 iterations), 
and keep training with an added CR-regularizer (red curve),  
precisely defined in Eq.~\eqref{eq:loss}. 
All other hyperparameters are  identical. 
We measure disentanglement using   
 the popular  metric of \cite{KM18} and defined in \S\ref{sec:CRexp}.
 The jump at epoch 28  suggests that contrastive regularization significantly enhances disentanglement, on top of what was achieved by InfoGAN regularizer alone.  
 
\subsection{Contrastive Regularizer Architecture} 
\label{sec:InfoGAN-CR}

\begin{figure}[h]
\centering
	\includegraphics[width=.4\textwidth]{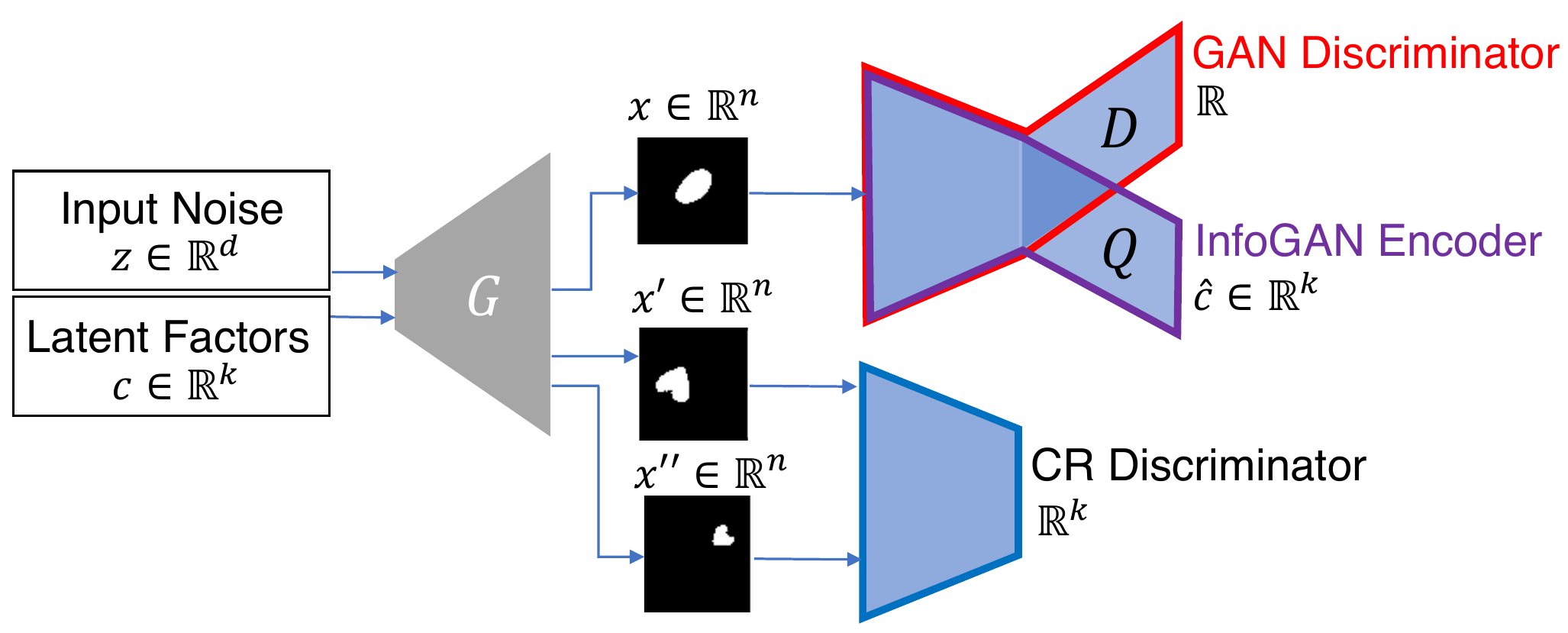}
	\put(-83,16){\small $H$}
  \caption{Like InfoGAN, InfoGAN-CR includes a GAN discriminator $D$ and an encoder $Q$, which share all convolutional layers and have separate fully-connected final layers. 
	In addition, the CR discriminator $H$ takes as input a pair of images $x$ and $x'$ that are generated by 
	sharing one fixed latent factor $c_i=c'_i$ for a randomly chosen $i\in [k] $, and randomly drawing the rest.  
	The discriminator is trained to correctly identify $i$, the index of the fixed factor. }
	\label{fig:CRarch}
\end{figure}

To approximate the Contrastive Regularizer in \eqref{eq:JSregularizer}, we introduce an additional discriminator $H:{\mathbb R}^{n}\times{\mathbb R}^{n}\to {\mathbb R}^k$ that performs multi-way hypothesis testing. 
We then justify its use via an equivalence in an ideal scenario in Theorem~\ref{rem:js}. 
Building upon InfoGAN's architecture (see \S\ref{sec:info} for details), 
we add contrastive regularization and refer 
to the resulting architecture as InfoGAN-CR, illustrated in Figure~\ref{fig:CRarch}.  
For non-negative scalars $\lambda$ and $\alpha$, this architecture is trained as 
\begin{eqnarray}
	\min_{G,H,Q} \;  \max_{D} \;\; \cL_{\rm Adv}(G,D) - \lambda \cL_{\rm Info}(G,Q) %
	- \alpha  \Lcon(G,H)
	\label{eq:InfoGAN-CR}
\end{eqnarray}

The pair of coupled images $x$ and $x'$ are 
generated according to a choice of a coupling that defines how to traverse the latent space. 
The discriminator $H$ tries to identify which code $i$ was shared between the paired images. 
Both the generator and the discriminator try to make the  
$k$-way hypothesis testing successful.
We use the standard cross entropy loss:
\begin{eqnarray} 
	 \Lcon(G,H)  =   \E_{I\sim{\rm U}([k]),(x,x')\sim Q^{(I)}} [  \langle I \,,\, \log H(x,x') \rangle ] \;,
	\label{eq:loss}
\end{eqnarray} 
where $Q^{(I)}$ denotes the joint distribution of the paired images, 
 $I$ denotes the one-hot encoding of the random index, and $H$ is a $k$-dimensional vector-valued neural network 
normalized to be $\langle \ones,H(x,x') \rangle=1$ for all $x$ and $x'$. 
This naturally encourages each latent code to make distinct and noticeable changes, hence promoting disentanglement. 
Further, the following theorem justifies the use of this architecture and loss. 
 We provide a proof in Appendix~\ref{sec:proof_rem2}. 

\begin{thm} 
	\label{rem:js} 
	When maximized over the class of all functions, 
	the maximum of Eq.~\eqref{eq:loss} is achieved by 
	$H(x,x') =  (1/Z_{x,x'}) \begin{bmatrix} Q^{(1)}(x,x') &, \cdots ,& Q^{(k)}(x,x') \end{bmatrix}  $ with a normalizing constant 
	$Z_{x,x'}=\sum_{i\in[k]} Q^{(i)}(x,x')$ and the maximum value is the generalized Jensen-Shannon divergence, 
	$$
	\max_H \cL_c (G,H) = d_{\rm JS}(Q^{(1)}, \ldots, Q^{(k)}) - \log k \;.$$ 
\end{thm} 

\begin{figure}[h]
		\centering
		\includegraphics[width=0.3\textwidth]{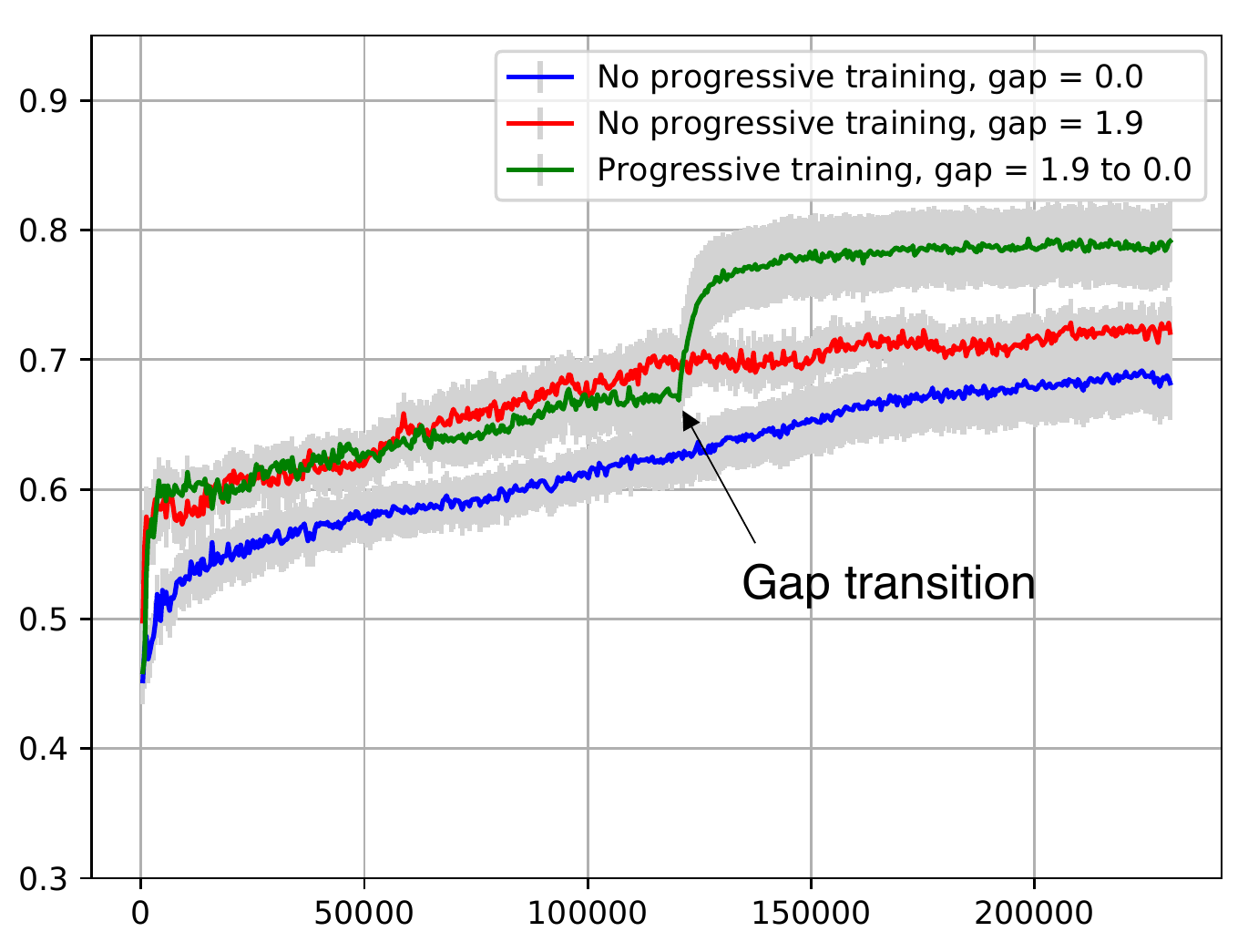}
		\put(-100,-7){Batch number}
		\put(-155,0){\rotatebox{90}{FactorVAE disentanglement}}
	\caption{ Reducing the contrastive gap  from  1.9 to 0 during training significantly improves FactorVAE scores.
	} \label{fig:prog}
\end{figure}

{\bf Progressive training.}
There are many ways to couple the latent variables. 
We prescribe progressively changing the hypotheses (or how we couple the images) during the course of the training, 
from easy to hard. 
The  hypotheses class we propose is as follows. 
First we draw a random index $I$ over $k$ indices, 
and sample the chosen latent code $c_I \in \reals$. 
Two images are generated with the same value of $c_I$;
the remaining factors are chosen independently at random. 
Letting $c_j^m$ denote the $j$th latent code for image $m\in \{1,2\}$, the  \emph{contrastive gap} is defined as $\min_{j\in [k]\setminus \{I\}} |c_j^1 - c_j^2|$.
\red{In Appendix \ref{app:latent-detail}, we discuss in more detail how we sample the latent codes for a given choice of a contrastive gap.}
The larger the contrastive gap, the more distinct the pair of samples. 
We gradually reduce the contrastive gap for progressive training (\S\ref{sec:design_exp}).
Figure~\ref{fig:prog} illustrates the power of progressive training on \dsprites{} dataset. 
For the `progressive  training' curve, we use a contrastive gap of 1.9 for 120,000 batches, and then introduce a (more aggressive) gap of 0. 
For the `no progressive training' curves, we use gap size of 0 or 1.9 for all 230,400 batches. 

\subsection{Empirical Evaluation of Contrastive Regularizer with Supervised Hyper-parameter Tuning} 
\label{sec:CRexp}

\begin{table*}[t]
    \begin{center}
    \scalebox{0.85}{
        \begin{tabular}{| c | c | l | l | l | l | l | l| l|}
            \hline
            & \textbf{Model} 
            & \textbf{FactorVAE} %
            & \textbf{DCI} %
            & \textbf{SAP} %
             & \textbf{Explicitness} %
            & \textbf{Modularity} %
            & \textbf{MIG} %
            & \textbf{BetaVAE} %
            \\ \hline
            
            \multirow{11}{*}{\textbf{VAE}}
            &VAE & 
                $0.63 \pm .06$ %
                & $0.30 \pm .10$ %
                &&&&0.10 %
                &\\
            &$\beta$-TCVAE & 
                $0.62 \pm .07$ %
                & $0.29 \pm .10$ %
                &&&& {\bf 0.45} %
                &\\
            &HFVAE & 
                $0.63 \pm .08$ %
                & $0.39 \pm .16$ %
                &&&& &\\
            &$\beta$-VAE & 
                $0.63 \pm .10$ %
                & $0.41 \pm .11$ %
                & 0.55 %
                &&&0.21 %
                &\\
            &CHyVAE & 
                0.77 %
                && &&&&\\
            &DIP-VAE &
                & &  0.53 %
                &&&&\\
            &FactorVAE & 
                0.82 %
                &&&&& 0.15 %
                &\\
            &FactorVAE (1.0)&  $0.79 \pm .01$  &  $0.67 \pm .03$  &  $0.47 \pm .03$  &  $0.78 \pm .01$  &  $0.79 \pm .01$  &  $0.27 \pm .03$  &  $0.79 \pm .02$\\
            &FactorVAE (10.0)&  $0.83 \pm .01$  &  $0.70 \pm .02$  &  $0.57 \pm .00$  &  $0.79 \pm .00$  &  $0.79 \pm .00$  &  $0.40 \pm .01$  &  $0.83 \pm .01$\\
            &FactorVAE (20.0)&  $0.83 \pm .01$  &  $0.72 \pm .02$  &  $0.57 \pm .00$  &  $0.79 \pm .00$  &  $0.79 \pm .01$  &  $0.40 \pm .01$  &  $0.85 \pm .00$\\
            &FactorVAE (40.0)&  $0.82 \pm .01$  &  ${\bf 0.74 \pm .01}$  &  $0.56 \pm .00$  &  $0.79 \pm .00$  &  $0.77 \pm .01$  &  ${ 0.43 \pm .01}$  &  $0.84 \pm .01$\\
\hline
                
            \multirow{4}{*}{\textbf{GAN}}
            & InfoGAN & 
                $0.59 \pm .70$ %
                & $0.41 \pm .05$ %
                &&&&0.05 %
                & \\
            & IB-GAN & 
                $0.80 \pm .07$ %
                & $0.67 \pm .07$ %
                &&&& & \\
            & InfoGAN (modified) & 
             $0.82 \pm 0.01$  &  $0.60 \pm 0.02$  &  $0.41 \pm 0.02$  &  $0.82 \pm 0.00$  &  $0.94 \pm 0.01$  &  $0.22 \pm 0.01$  &  $0.87 \pm 0.01$\\
            & InfoGAN-CR& 
             ${\bf 0.88 \pm 0.01}$  &  $0.71 \pm 0.01$  & $ {\bf0.58 \pm 0.01}$  & $ {\bf 0.85 \pm 0.00}$  &  ${\bf 0.96 \pm 0.00}$  &  $0.37 \pm 0.01$  &   ${\bf0.95 \pm 0.01}$\\
 \hline
        \end{tabular}
    }
    \end{center}
      	\caption{Comparisons of the popular disentanglement metrics on the \dsprites{} dataset. 
	A perfect disentanglement corresponds to 1.0 scores. 
	The proposed \gan achieves the highest  score on most cases, compared to 
	the best  reported result for each 
	baseline. 
	See Appendix \ref{app:infogan} for InfoGAN (modified). 
	\red{The InfoGAN (modified) and \gan rows are averaged over 50 runs. Appendix \ref{app:dsprites-reproducibility} gives more details on the reproducibility of the results.}
	We show in Table~\ref{tbl:MCdsprites} that with our proposed model selection scheme, we improve the performance even further. } 
  	\label{tbl:comp}
\end{table*}

For quantitative evaluation, we run experiments on synthetic datasets with pre-defined latent factors, including \dsprites{} \cite{dSprites} and \teapots{} \cite{EW18}.\footnote{The code for all experiments is available at \url{https://github.com/fjxmlzn/InfoGAN-CR}}
We evaluate %
disentanglement using the popular metrics from \cite{KM18,EW18,kumar2017variational,ridgeway2018learning,CLG18,HMP16}. 
For qualitative evaluation, we use our synthetic datasets as well as the CelebA dataset \cite{LLW15}.
More details on datasets and metrics can be found in Appendix~\ref{app:metrics}. 

It is typical in disentanglement literature to select hyperparameters in a supervised manner in synthetic datasets where ground truth disentanglement is known. 
We do the same in this section and choose hyperparameters of all the models we train (FactorVAE, InfoGAN modified, and InfoGAN-CR).  These are fair comparisons as all reported scores in this section are results of such hyperparameter tuning (some by us and some by the experimenters). 
However, this practice of supervised hyperparameter tuning is  problematic; we resolve this issue in \S\ref{sec:MC}. 
Perhaps surprisingly, we show that our {\em unsupervised model selection} finds a better model than that found via supervised hyperparameter tuning.

\subsubsection{\dsprites{} Dataset} 
\label{sec:design_exp}

We compute and/or reproduce disentanglement metrics for a number of protocols in Table \ref{tbl:comp}. We provide details of the experiments in Appendix~\ref{app:dsprites}, and focus on the interpretation of the results in this section. An example of latent traversal of the output of InfoGAN-CR is shown in Figure~\ref{fig:latenttraversal}. 

Contrastive regularization provides a clear gain in disentanglement, bringing \gannosp's FactorVAE score up to 0.90, higher than any baseline from the VAE or GAN literature. 
A similar trend holds for most of the metrics. 
We were made aware of independent work that proposes a special case of Contrastive Regularization in \cite{li2018unsupervised}; 
concretely, \cite{li2018unsupervised} fixes $\lambda=0$ in our loss \eqref{eq:InfoGAN-CR}, and also uses a special coupling that matches all but one latent code in $c$ for the matched pairs. 
This empirically achieves a lower FactorVAE scores (0.39 $\pm$ 0.02 standard error over 10 runs) than even vanilla InfoGAN. 
Note that this difference is not a matter of parameter tuning, but of the loss function and training mechanism; indeed, in our own preliminary trials, we found that training a CR-regularizer without the InfoGAN loss, as in \cite{li2018unsupervised}, achieved similarly poor performance. 
\red{The choice of coupling in our contrastive regularizer, the progressive training we propose, and the InfoGAN loss are all critical in achieving the improved the performance, as described in Appendix \ref{app:onlycr}.} 
Hence, we do not consider it as a  baseline moving forward.

\begin{table*}[h]
    \begin{center}
    \scalebox{0.9}{
        \begin{tabular}{ | c | l | l | l | l | l | l| l |}
            \hline
             \textbf{Model} 
            & \textbf{FactorVAE} %
            & \textbf{DCI} %
            & \textbf{SAP} %
            & \textbf{Explicitness} %
            & \textbf{Modularity} %
            & \textbf{MIG} %
            & \textbf{BetaVAE} %
            \\ \hline

            FactorVAE &  $0.79 \pm .03$  &  $0.55 \pm .04$  &  $0.49 \pm .05$  &  ${\bf 0.84 \pm .01}$  &  $0.72 \pm .02$  &  $0.24 \pm .03$  &  ${\bf 0.94 \pm .02}$\\

             InfoGAN (modified)&  $0.76 \pm .06$  &  $0.62 \pm .06$  &  ${\bf 0.57 \pm .06}$  &  $0.82 \pm .04$  &  ${\bf 0.98 \pm .01}$  &  $0.34 \pm .04$  &  $0.90 \pm .07$\\
             InfoGAN-CR %
            &  ${\bf 0.82 \pm .02}$  &  ${\bf 0.66 \pm .01}$  &  $0.53 \pm .02$  &  $0.81 \pm .01$  &  $0.97 \pm .00$  &  ${\bf 0.38 \pm .02}$  &  $0.89 \pm .02$\\
            \hline
        \end{tabular}
    }
    \end{center}

    \caption{ Comparisons of the popular disentanglement metrics on the \teapots{}. 
    We show in Table~\ref{tbl:MCteapot} that with our proposed model selection scheme, we achieve the best performance on all metrics.
    }
    \label{tbl:teapot}
\end{table*}

\subsubsection{\teapots{} Dataset}
We ran \gan on the \teapots{} dataset from  \cite{EW18},
with images of teapots in various orientations and colors generated by the renderer in \cite{MW16}. 
Details on this point, our implementation, and additional plots appear in Appendix \ref{app:teapots}. 
Table \ref{tbl:teapot} shows the disentanglement scores of FactorVAE and InfoGAN compared to \gannosp.
While the results with this {\em supervised hyperparameter tuning} are mixed (none of the methods dominate), we show in \S\ref{sec:MC}, Table~\ref{tbl:MCteapot} that, perhaps surprisingly, our proposed {\em unsupervised model selection} finds a model that dominates all baseline algorithms.

\begin{figure*}[h]
	\begin{center}
		\includegraphics[width=4.8in]{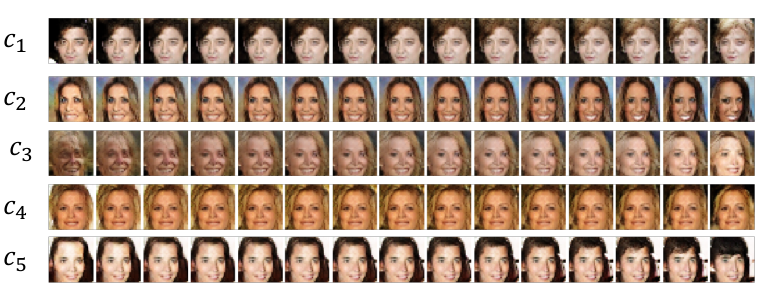}
		\put(0,112){Hair color}
		\put(0,85){Azimuth}
		\put(0,63){Lighting}
		\put(0,37){Background}
		\put(0,15){Bangs}
	\end{center}
	\caption{Latent traversal for CelebA dataset, using \gannosp.
	}
	\label{fig:celeba_traversal}
	
\end{figure*}

\subsubsection{CelebA Dataset} 
We train \gan on the CelebA dataset of 202,599 celebrity facial images. 
Since these images do not have known continuous latent factors, we cannot compute the disentanglement metric. 
We therefore evaluate this dataset qualitatively by producing latent traversals, as seen in Figure  \ref{fig:celeba_traversal}.
Details of this experiment are included in Appendix \ref{app:celeba}.

\section{ModelCentrality: Self-supervised Model Selection}
\label{sec:MC}
\begin{figure}[t]
\centering
	\includegraphics[width=.28\textwidth]{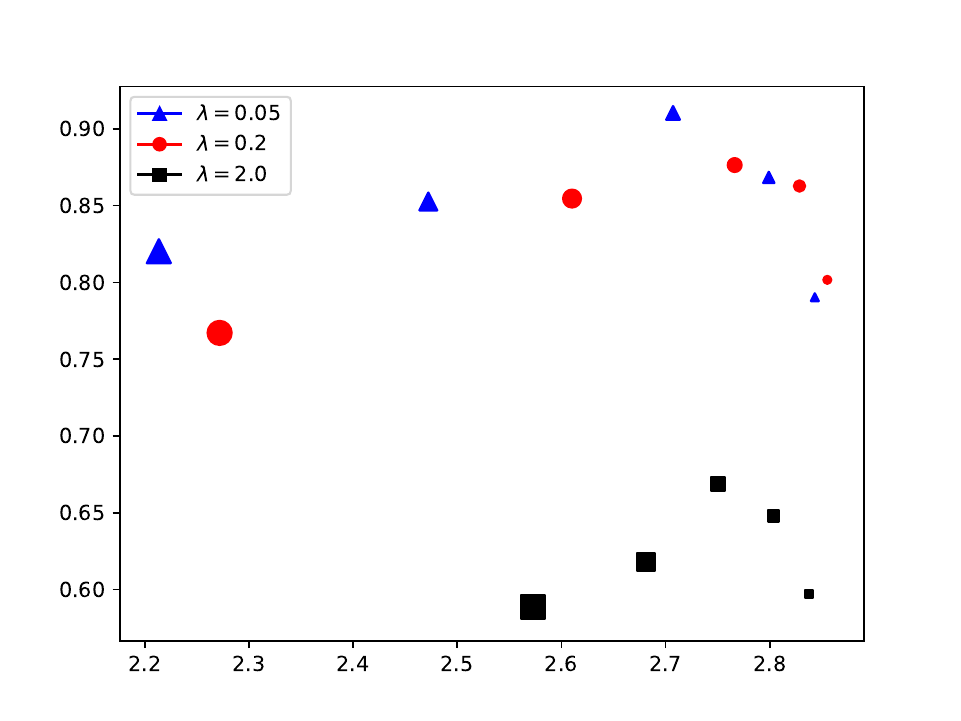}
	\put(-99,-3){Inception Score}
	\put(-21,54){$\alpha=0$}
	\put(-26,78){$\alpha=1$}
	\put(-40,88){$\alpha=2$}
	\put(-85,75){$\alpha=4$}
	\put(-110,64){$\alpha=8$}
	\put(-142,21){\rotatebox{90}{Disentanglement}}
\caption{ Inception score and FactorVAE score achieved by various hyper-parameters in \eqref{eq:InfoGAN-CR}. 
The size of each point denotes $\alpha \in \{0,1,2,4,8\}$, in the order of increasing size. 
We explicitly label this for $\lambda=0.05$ (blue triangles).}
\label{fig:varyalpha}
\end{figure}
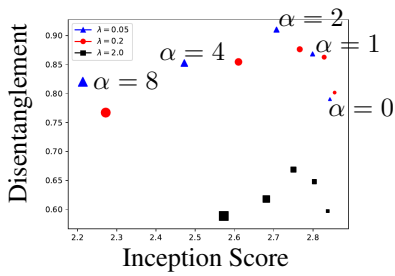

The achievable scores in 
Table~\ref{tbl:comp} are a consequence of {\em supervised} hyper-parameter tuning, for both our models and all baseline models. 
As shown in Figure~\ref{fig:varyalpha}, 
the designer runs experiments with multiple hyper-parameters---whose performance could vary significantly---and chooses one hyper-parameter that gives the best average performance.
This approach is supervised, as performance evaluation requires access to a synthetic data generator with access to the  ground truth disentangled codes. 

Supervised  hyper-parameter tuning is problematic, as $(i)$ in important real-world applications we do not have ground truth data, and $(ii)$ a more complex model with a larger space to tune could get better scores by an extensive search. 
To this end, we propose a novel unsupervised model selection scheme called  \emph{ModelCentrality} that bypasses both of these concerns. 

\subsection{ModelCentrality} 

Suppose there is a notion of an optimal disentanglement that we want to discover from the data. 
We start from a premise that well-disentangled  models should be close to that optimal model, and hence also close to each other. 
To measure similarity between  models, we borrow insights from a long line of research in measuring disentanglement.  
\red{In  particular, prior work suggests that models with good disentanglement metrics (e.g.~those in Table~\ref{tbl:comp}) tend to exhibit qualitatively good disentanglement properties, e.g., via latent traversals.
This suggests that disentanglement scores can be used to measure how close the disentangled latent codes of one model are to the latent codes of another.}

\red{ 
Consider a trained model $G:{\mathbb R}^k\to {\mathbb R}^n$ 
(here we only consider the model as mapping a disentangled latent code $c\in{\mathbb R}^k$ to the image $x\in {\mathbb R}^n$ and treat the $z\in{\mathbb R}^d$ as an inherent randomness in the generative model). 
Existing metrics also require the corresponding {\em encoder} 
 $Q:{\mathbb R}^n \to {\mathbb R}^k$ that maps samples to estimated disentangled latent factors. 
For example, the popular FactorVAE metric of \cite{KM18} of a trained generative model $G_i$ measures how well its encoder $Q_i$ can estimate, from real samples, the true latents of real samples (for example the ground truths dSprites dataset with also the true disentangled latent factors).}

\red{ 
Instead of the original FactorVAE score, which requires supervision from the training data with ground truths latent codes, 
we use other trained models as a surrogate for the ground truths. 
ModelCentrality treats the distribution of another model $G_j$ as the ground truth.
Given two trained models: $G_i$ and $G_j$, we can measure how well the encoder $Q_i$ can estimate, from the generated samples of model $G_j$, the learned latents of the generated samples of model $G_j$. 
Hence, we can compute the similarity from $G_j$ to $G_i$ by (1) generating samples using the target model $G_j$; 
(2) passing those samples through the encoder $Q_i$ of model $G_i$ to estimate its latents, 
(3) using these estimated latents to evaluate the FactorVAE metric by using the latents generated by target model $G_j$ as ground truths.
}
\red{This similarity metric is an instance of self-supervision, as we treat one model as the \emph{target label}  and no ground-truth labels are needed. }

\red{Given a pool of $N$ trained generative models, we compute $A_{ij}$ as the disentanglement score achieved by model $G_i$ treating model 
$G_j$ as the target model with the way mentioned above.}
\red{Then we define a symmetric similarity matrix $B\in{\mathbb R}^{N\times N}$, where the similarity between a model $i$ and model $j$ is denoted by $B_{ij}$, and is computed as $B_{ij} = (1/2)(A_{ij}+A_{ji})$}.
In our experiments, we choose FactorVAE score as the disentanglement metric, because it is popular and robust, 
but we will show that FactorVAE-based  ModelCentrality predicts all other scores 
accurately  in Figure~\ref{fig:metric_corr_dSprites_infogancr}.

\begin{figure}[h]
    \centering
    \includegraphics[width=0.6\linewidth]{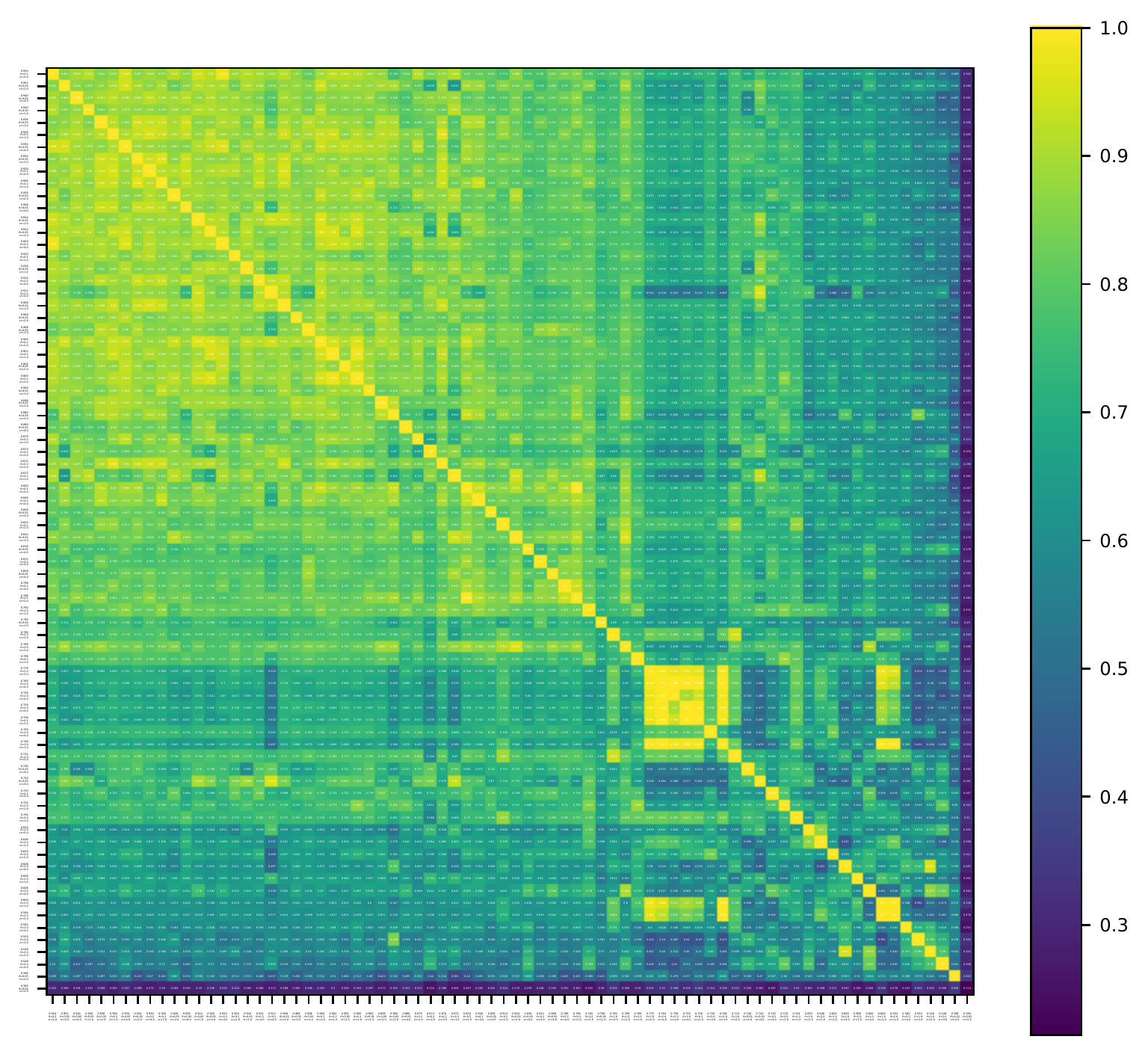}
    \caption{Heat map of the matrix $B$ used to compute ModelCentrality for each InfoGAN-CR model trained with \dsprites{} dataset. Each row/column corresponds to one trained model, which are  sorted according to FactorVAE score on the ground truth factors (computed with the supervised ground truths  dSprites dataset). Top-left  is the highest FactorVAE scoring model. }
    \label{fig:model_corr_dSprites_infogancr}
\end{figure}

We experimentally confirm our premise that good models are close to each other  
in Figure \ref{fig:model_corr_dSprites_infogancr}, which illustrates the similarity matrix $B$. 
\red{The rows/columns of this matrix are sorted by FactorVAE score, computed on the ground truth disentanglement factors.}
As expected, models that are better disentangled (as measured by FactorVAE score) are closer to each other (top-left), 
and the models that are not disentangled are far from other models (bottom-right).

This observation naturally suggests using some notion of a {\em central model} in our pool as the best model.  
We propose a measure of ModelCentrality based on the medoid of models with respect to the similarity matrix $B$. 
We define the ModelCentrality  of a model $i$ as 
$s_i = \frac{1}{n-1} \sum_{j\neq i} B_{ij}$. 
We then select the model with the largest ModelCentrality, which coincides with the medoid in the pool of models. 
The pseudocode for computing ModelCentrality is given in Algorithm~\ref{alg:MC}. 

\redca{Besides model selection, ModelCentrality can be used for other tasks, as $s_i$ provides a quantitative evaluation of the $i$-th model. For example, ModelCentrality can rank the models according to $s_i$. It can also be used for hyper-parameter selection by averaging the $s_i$'s of the models trained with the same hyper-parameter, and selecting the best hyper-parameter.}

\begin{algorithm}[h]
	\caption{ModelCentrality}
	\label{alg:MC}
	\SetAlgoLined
	\KwIn{$N$ pairs of generative models and latent code encoder: $(G_1,Q_1),...,(G_N,Q_N)$, \newline 
	 supervised disentanglement metric $f:{\rm encoder}\times {\rm model} \to {\mathbb R}$}
	\KwOut{the estimated best model $G^*$}
	Initialize a zero matrix: $A\in\mathbb{R}^{N \times N }$
	
	\For{$i, j=1 \rightarrow N$}{
					$A_{ij} \leftarrow f(Q_i,G_j)$
	}
	$B \leftarrow (A+A^T)/2$
	
	\For{$i=1 \rightarrow N$}{
			$s_i \leftarrow (\sum_{j\not=i} B_{ij})/(N-1)$
	}
	$k\leftarrow\arg\max_i s_i$
	
	 $G^* \leftarrow G_k$
\end{algorithm}

\subsection{Comparison with State-of-the-art Model Selection} 
\label{sec:MCexp}
\red{
We compare our model selection approach with state-of-the-art schemes from \cite{duan2019heuristic}. 
The first scheme, UDR Lasso, defines a distance $A_{ij}$ from one model $i$ to another model $j$  as follows. 
Consider the encoder of model $i$ that maps an image to a latent code: $\hat{c}=Q_i(x) \in{\mathbb R}^k$.
A linear regressor is trained with Lasso to predict $Q_i(x)$ from $Q_j(x)$ using samples $\{x^{(\ell)} \in {\mathbb R}^n \}_{\ell\in S_{\rm train}}$ from the training dataset.
If two models are identical, then the resulting (matrix valued) Lasso regressor will be a permutation matrix.
Otherwise, a formula is applied to give a score \cite{duan2019heuristic}. 
The second approach,  UDR Spearman, uses  similar approach, except instead of training a Lasso regressor, the Spearman correlation coefficient  is computed.
}

\begin{figure}[h]
    \centering
    \includegraphics[width=0.8\linewidth]{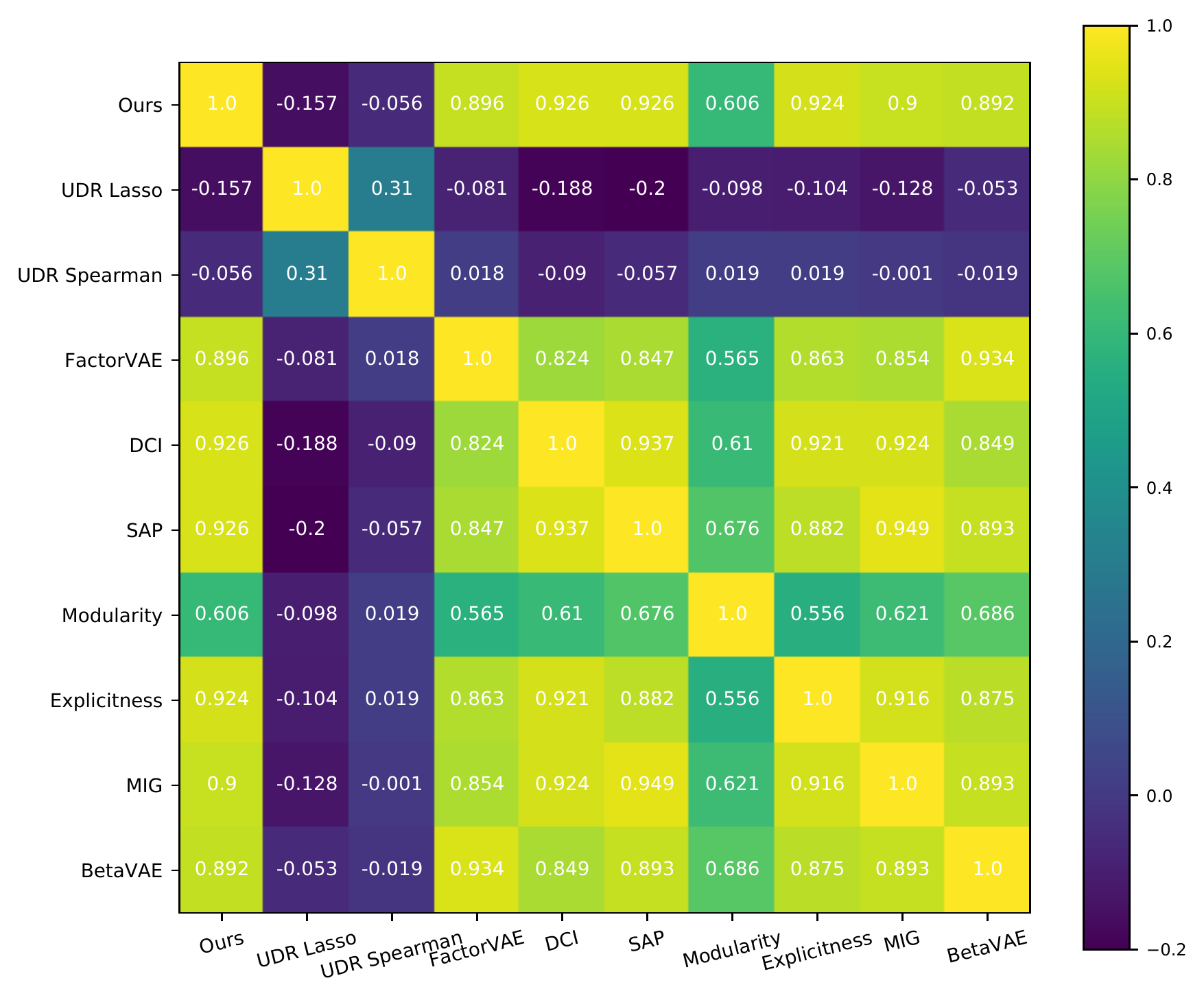}
    \put(-145,152){Spearman rank correlation}
    \caption{ \redca{The rank correlation of the metrics on \dsprites{} dataset. }The first row/column is our ModelCentrality, which is highly correlated with all other disentanglement scores (row/column 4-10). Competing  schemes of UDR Lasso and UDR Spearman are nearly  uncorrelated. }
    \label{fig:metric_corr_dSprites_infogancr}
\end{figure}

\begin{table*}[h!]
    \begin{center}
    \scalebox{0.93}{
        \begin{tabular}{ | c | l | l | l | l | l | l| l|}
            \hline
             \textbf{Model} 
            & \textbf{FactorVAE}
            & \textbf{DCI} 
            & \textbf{SAP} 
             & \textbf{Explicitness} 
            & \textbf{Modularity} 
            & \textbf{MIG} 
            & \textbf{BetaVAE}  \\ \hline

            FactorVAE with &&&&&&&\\
            UDR Lasso&  $0.81 \pm .00$  &  $0.70 \pm .01$  &  $0.56 \pm .00$  &  $0.79 \pm .00$  &  $0.78 \pm .00$  &  ${\bf 0.40 \pm .00}$ &  $0.84 \pm .00$\\
            UDR Spearman&  $0.79 \pm .00$  &   ${\bf 0.73 \pm .00}$  &  $0.53 \pm .01$  &  $0.79 \pm .00$  &  $0.77 \pm .00$  &  ${\bf 0.40 \pm .01}$ &  $0.79 \pm .00$\\
            ModelCentrality&  ${\bf 0.84 \pm .00}$  &  ${\bf 0.73 \pm .01}$  &  ${\bf 0.58 \pm .00}$  &  ${\bf 0.80 \pm .00}$  &  ${\bf 0.82 \pm .00}$  &  $0.37 \pm .00$  &  ${\bf 0.86 \pm .00}$\\\hline
            \redca{Best model in the pool} &  $0.88$  &  $nan$  &  $0.58$  &  $0.79$  &  $0.79$  &  $0.39$  &  $0.83$\\
            \hline\hline

            InfoGAN-CR with&&&&&&&\\
            UDR Lasso&  $0.86 \pm .01$  &  $0.68 \pm .01$  &  $0.49 \pm .01$  &  $0.84 \pm .00$  &  $0.96 \pm .00$  &  $0.30 \pm .01$ & $0.92 \pm .01$\\
            UDR Spearman&  $0.84 \pm .01$  &  $0.67 \pm .01$  &  $0.53 \pm .01$  &  $0.84 \pm .00$  &  $0.96 \pm .00$  &  $0.31 \pm .01$  &  $0.90 \pm .01$\\
            ModelCentrality &  ${\bf 0.92 \pm .00}$  &  ${\bf 0.77 \pm .00}$  &  ${\bf 0.65 \pm .00}$  &  ${\bf 0.87 \pm .00}$  &  ${\bf 0.99 \pm .00}$  &  ${\bf 0.45 \pm .00}$ & ${\bf 0.99 \pm .00}$\\\hline
            \redca{Best model in the pool} &  $0.95$  &  $0.77$  &  $0.65$  &  $0.88$  &  $0.99$  &  $0.46$  &  $0.99$\\\hline
        \end{tabular}
    }
    \end{center}
    \caption{\redca{On the \dsprites{} dataset},  models selected with  ModelCentrality outperform those selected with UDR Lasso and UDR Spearman, for both FactorVAE and InfoGAN-CR respectively. Further, this outperforms the \emph{hyper-parameter} tuned models supervised by the groundtruths disentangled codes reported in Table~\ref{tbl:comp}. \redca{Results of the model with the best FactorVAE score computed by ground truth disentangled code are also included for reference (row ``best model in the pool'').}}
    \label{tbl:MCdsprites}
\end{table*}

\begin{table*}[h!]
    \begin{center}
        \begin{tabular}{| c | l | l | l | l | l | l| l |}
            \hline
            \textbf{Model} 
            & \textbf{FactorVAE} 
            & \textbf{DCI} 
            & \textbf{SAP} 
            & \textbf{Explicitness} 
            & \textbf{Modularity} 
            & \textbf{MIG} 
            & \textbf{BetaVAE} \\ \hline
             ModelCentrality  &  ${1.00}$  &  ${0.75} $  &  ${ 0.77} $  &  ${ 0.92} $  &  ${1.00} $  &  ${ 0.53} $  &  ${ 1.00} $\\
            \redca{Best model in the pool}&  $1.00$  &  $0.85$  &  $0.89$  &  $0.92$  &  $1.00$  &  $0.47$  &  $1.00$\\\hline
        \end{tabular}
    \end{center}
    \caption{\redca{On the \teapots{} dataset, InfoGAN-CR models selected with ModelCentrality has close performance to the model with the best groundtruth FactorVAE score and DCI score.
    The standard errors of ModelCentrality are  less than 0.01 and we omit them in this table. }}
    \label{tbl:MCteapot} 
\end{table*}

In  our experiments, we  compare ModelCentrality to UDR Lasso and UDR Spearman on InfoGAN-CR and FactorVAE models trained on \dsprites{} and \teapots{} datasets.  
\redca{On the \dsprites{} dataset}, we first generated $N=76$ InfoGAN-CR models from a grid of hyper-parameters. 
Figure \ref{fig:metric_corr_dSprites_infogancr} shows the Spearman rank correlations between models selected different metrics (including two UDR approaches and ModelCentrality). 
\red{
To produce this figure, we start with trained models $m_1,\ldots, m_{76}$, and a list of disentanglement metrics $f_1,\ldots, f_{10}$,  including ModelCentrality, UDR (Spearman and Lasso), and an assortment of other disentanglement  metrics.  
Then for the $i$th metric  $f_i$, we compute $v_i = [f_i(m_1),\ldots, f_i(m_{76})] \in {\mathbb R}^{76}$.
Note that all the metrics, except ModelCentrality and UDR, require the access to ground truths latent factors (and hence are supervised). 
Finally, the $(i,j)$th entry of Figure \ref{fig:metric_corr_dSprites_infogancr} is the Spearman rank correlation coefficient between vectors $v_i$ and $v_j$.
}

\red{
Figure \ref{fig:metric_corr_dSprites_infogancr} illustrates two points.
First, ModelCentrality is not closely correlated with UDR Spearman  or UDR Lasso, since the 2nd and 3rd rows/columns have low correlation  coefficients.  
Second, ModelCentrality \emph{is} closely correlated with the remaining disentanglement metrics.
}
In Appendix~\ref{app:model_centrality}, we show a more detailed  statistics of the scores, and show that a similar results hold when selecting FactorVAE models and also under \teapots{} dataset. 
\red{
This suggests that choosing a model with maximum ModelCentrality tends to maximize existing disentanglement metrics, {without} requiring access to ground truth labels---an intuition that we confirm  in Tables~\ref{tbl:MCdsprites} and \ref{tbl:MCteapot}.}
Perhaps surprisingly, not only does ModelCentrality outperform UDR schemes, but it also selects models that outperform 
(a set of) models trained with a supervised hyper-parameter tuning from literature and from our experiments in \S\ref{sec:CRexp}. 
 Notice the subtle difference in ModelCentrality producing a single model versus 
 supervised hyper-parameter tuning producing a hyper-parameter for training a set of models.  
 \redca{In fact, as shown in Table \ref{tbl:MCdsprites} and Table \ref{tbl:MCteapot}, the model selected with ModelCentrality has very close performance to the model with the best ground truth FactorVAE score. Under some metrics other than FactorVAE score, the model selected with ModelCentrality is even better. }

\red{
A natural question is why ModelCentrality outperforms UDR. }
Several aspects of UDR Lasso contribute to its unreliability. 
$(i)$ Lasso involves a hyper-parameter, which can significantly change the resulting score. 
 $(ii)$ Lasso is restricted to linear relations, whereas two perfectly disentangled models can have highly non-linear relations. 
 $(iii)$ In addition, UDR Lasso does not generalize to discrete latent codes. 
UDR Spearman uses the Spearman's rank correlation in place of Lasso, and is reported to be inferior to UDR Lasso \cite{duan2019heuristic}. 
Notice that UDR schemes  inherit the issues present in the  disentanglement scores of DCI \cite{EW18},  
from which the UDR schemes are derived. 
The proposed ModelCentrality is derived from FactorVAE scores \cite{KM18}, which is popular, principled, and demonstrated to be a stable measure of disentanglement.

\section{Conclusion} 
\label{sec:conclusion}

This work makes two contributions. 
First, we introduce InfoGAN-CR, a new architecture for training disentangled GANs. 
Next, we introduce ModelCentrality, a new framework for selecting disentangled models. 
Numerical results in Tables~\ref{tbl:comp},  \ref{tbl:teapot}, \ref{tbl:MCdsprites}, and \ref{tbl:MCteapot} confirm 
that InfoGAN-CR together with ModelCentrality achieves the best disentanglement across all metrics in the literature.  
This is surprising because hyper-parameter tuning in the literature is typically supervised:  oracle access to the ground truth disentangled latent codes 
is needed. 
Instead, our proposed ModelCentrality is unsupervised, yet reliably selects a superior model.
While ModelCentrality can be used to select both GAN and VAE based models, 
ModelCentrality with InfoGAN-CR improves upon ModelCentrality with other state-of-the-art methods, including VAE-based ones. 
Unlike other VAE-based methods, our approach seamlessly generalizes to semi-supervised settings. 
If we have paired examples where one latent code has been changed, e.g., a person with and without glasses, 
this can be readily incorporated in our architecture. 
Hence, one way to interpret our approach is as a self-supervised training from unsupervised data. 

In addition, we experimentally find that CR substantially increases the disentanglement capabilities of InfoGAN, 
but does not appear to affect the state-of-the-art VAEs (Appendix~\ref{app:tc}). 
Similarly, we experimentally show that 
the total correlation regularization, a popular technique for disentangling VAEs, 
do not improve disentanglement in GAN training. 
This suggests that disentangling 
 VAEs and GANs require fundamentally different techniques.
 The proposed CR regularization could be used in any application of disentangled GANs, 
e.g., hierarchical image representation or reinforcement learning. 
Understanding this phenomenon analytically is an interesting direction for future work,
and may give rise to a more general understanding of how to design regularizers for GANs as opposed to VAEs. 

Another key question  is to understand disentanglement in challenging datasets, 
compared to those studied in the literature as a benchmark.
We study two such datasets. 
The first one studies three dimensional rotations on the \teapots{} dataset in Appendix \ref{app:rotation}. 
Existing training datasets includes only a subset of the full rotations, making disentanglement substantially easy. 
When training data is drawn from complete set of rotations in 3-D space, several challenges arise. 
The usual rotations along the three standard basis vectors do not commute, hence do not disentangle. 
We can find a commutative coordinate system, but it is not uniquely defined. 
Our preliminary experiments suggest that %
current state-of-the-art methods fail to learn a disentangled representation. 
The second one studies two dimensional polar coordinate system using a novel dataset (\cdspriteslong) in Appendix~\ref{sec:cdsprites}. 
State-of-the-art methods fail to learn the disentangled representation of the polar coordinates.

\section*{Acknowledgements}
\redca{We thank Hyunjik Kim for a productive discussion on the experimental evaluation.}
We thank Qian Ge for valuable discussions.
This work used the Extreme Science and Engineering Discovery Environment (XSEDE), which is supported by NSF grant OCI-1053575. 
It used the Bridges system, which is supported by NSF award number ACI-1445606, at the Pittsburgh Supercomputing Center (PSC). 
This work is partially supported by funding from Siemens and AWS cloud computing credits from Amazon.
Giulia Fanti acknowledges support from a Google Faculty Research Award and a JP Morgan Chase Faculty Research Award.
Sewoong Oh acknowledges funding from Google Faculty Research Award, Intel Future Wireless
Systems Research Award, and NSF awards CCF-1927712, CCF-1705007, IIS-1929955, and CNS-2002664.

\newpage
\bibliography{_gan}

\begin{thebibliography}{72}
\providecommand{\natexlab}[1]{#1}
\providecommand{\url}[1]{\texttt{#1}}
\expandafter\ifx\csname urlstyle\endcsname\relax
  \providecommand{\doi}[1]{doi: #1}\else
  \providecommand{\doi}{doi: \begingroup \urlstyle{rm}\Url}\fi

\bibitem[Ainsworth et~al.(2018{\natexlab{a}})Ainsworth, Foti, and Fox]{AFF18}
Ainsworth, S.~K., Foti, N.~J., and Fox, E.~B.
\newblock Disentangled vae representations for multi-aspect and missing data.
\newblock \emph{arXiv preprint arXiv:1806.09060}, 2018{\natexlab{a}}.

\bibitem[Ainsworth et~al.(2018{\natexlab{b}})Ainsworth, Foti, Lee, and
  Fox]{AFL18}
Ainsworth, S.~K., Foti, N.~J., Lee, A. K.~C., and Fox, E.~B.
\newblock oi-{VAE}: Output interpretable {VAE}s for nonlinear group factor
  analysis.
\newblock In \emph{Proceedings of the 35th International Conference on Machine
  Learning}, volume~80, pp.\  119--128, 2018{\natexlab{b}}.

\bibitem[Aizerman(1964)]{aizerman1964theoretical}
Aizerman, M.~A.
\newblock Theoretical foundations of the potential function method in pattern
  recognition learning.
\newblock \emph{Automation and remote control}, 25:\penalty0 821--837, 1964.

\bibitem[Ansari \& Soh(2018)Ansari and Soh]{CHYVAE}
Ansari, A.~F. and Soh, H.
\newblock Hyperprior induced unsupervised disentanglement of latent
  representations.
\newblock \emph{arXiv preprint arXiv:1809.04497}, 2018.

\bibitem[Aumentado-Armstrong et~al.(2019)Aumentado-Armstrong, Tsogkas, Jepson,
  and Dickinson]{aumentado2019geometric}
Aumentado-Armstrong, T., Tsogkas, S., Jepson, A., and Dickinson, S.
\newblock Geometric disentanglement for generative latent shape models.
\newblock In \emph{Proceedings of the IEEE International Conference on Computer
  Vision}, pp.\  8181--8190, 2019.

\bibitem[Awiszus et~al.(2019)Awiszus, Ackermann, and
  Rosenhahn]{awiszus2019learning}
Awiszus, M., Ackermann, H., and Rosenhahn, B.
\newblock Learning disentangled representations via independent subspaces.
\newblock In \emph{Proceedings of the IEEE International Conference on Computer
  Vision Workshops}, pp.\  0--0, 2019.

\bibitem[Bengio et~al.(2013)Bengio, Courville, and
  Vincent]{bengio2013representation}
Bengio, Y., Courville, A., and Vincent, P.
\newblock Representation learning: A review and new perspectives.
\newblock \emph{IEEE transactions on pattern analysis and machine
  intelligence}, 35\penalty0 (8):\penalty0 1798--1828, 2013.

\bibitem[Brock et~al.(2018)Brock, Donahue, and Simonyan]{BIGGAN}
Brock, A., Donahue, J., and Simonyan, K.
\newblock Large scale gan training for high fidelity natural image synthesis.
\newblock \emph{arXiv preprint arXiv:1809.11096}, 2018.

\bibitem[Burgess et~al.(2018)Burgess, Higgins, Pal, Matthey, Watters,
  Desjardins, and Lerchner]{burgess2018understanding}
Burgess, C.~P., Higgins, I., Pal, A., Matthey, L., Watters, N., Desjardins, G.,
  and Lerchner, A.
\newblock Understanding disentangling in beta-vae.
\newblock \emph{arXiv preprint arXiv:1804.03599}, 2018.

\bibitem[Caselles-Dupr{\'e} et~al.(2019)Caselles-Dupr{\'e}, Ortiz, and
  Filliat]{caselles2019symmetry}
Caselles-Dupr{\'e}, H., Ortiz, M.~G., and Filliat, D.
\newblock Symmetry-based disentangled representation learning requires
  interaction with environments.
\newblock In \emph{Advances in Neural Information Processing Systems}, pp.\
  4608--4617, 2019.

\bibitem[Chen \& Batmanghelich(2019)Chen and Batmanghelich]{chen2019weakly}
Chen, J. and Batmanghelich, K.
\newblock Weakly supervised disentanglement by pairwise similarities.
\newblock \emph{arXiv preprint arXiv:1906.01044}, 2019.

\bibitem[Chen et~al.(2018)Chen, Li, Grosse, and Duvenaud]{CLG18}
Chen, T.~Q., Li, X., Grosse, R., and Duvenaud, D.
\newblock Isolating sources of disentanglement in variational autoencoders.
\newblock \emph{arXiv preprint arXiv:1802.04942}, 2018.

\bibitem[Chen et~al.(2016)Chen, Duan, Houthooft, Schulman, Sutskever, and
  Abbeel]{CDH16}
Chen, X., Duan, Y., Houthooft, R., Schulman, J., Sutskever, I., and Abbeel, P.
\newblock Infogan: Interpretable representation learning by information
  maximizing generative adversarial nets.
\newblock In \emph{Advances in Neural Information Processing Systems}, pp.\
  2172--2180, 2016.

\bibitem[Cohen \& Welling(2014)Cohen and Welling]{CW14}
Cohen, T.~S. and Welling, M.
\newblock Transformation properties of learned visual representations.
\newblock \emph{arXiv preprint arXiv:1412.7659}, 2014.

\bibitem[Creager et~al.(2019)Creager, Madras, Jacobsen, Weis, Swersky, Pitassi,
  and Zemel]{creager2019flexibly}
Creager, E., Madras, D., Jacobsen, J.-H., Weis, M.~A., Swersky, K., Pitassi,
  T., and Zemel, R.
\newblock Flexibly fair representation learning by disentanglement.
\newblock \emph{arXiv preprint arXiv:1906.02589}, 2019.

\bibitem[Denton et~al.(2017)]{denton2017unsupervised}
Denton, E.~L. et~al.
\newblock Unsupervised learning of disentangled representations from video.
\newblock In \emph{Advances in neural information processing systems}, pp.\
  4414--4423, 2017.

\bibitem[Desjardins et~al.(2012)Desjardins, Courville, and Bengio]{DCB12}
Desjardins, G., Courville, A., and Bengio, Y.
\newblock Disentangling factors of variation via generative entangling.
\newblock \emph{arXiv preprint arXiv:1210.5474}, 2012.

\bibitem[Duan et~al.(2019{\natexlab{a}})Duan, Watters, Matthey, Burgess,
  Lerchner, and Higgins]{duan2019heuristic}
Duan, S., Watters, N., Matthey, L., Burgess, C.~P., Lerchner, A., and Higgins,
  I.
\newblock A heuristic for unsupervised model selection for variational
  disentangled representation learning.
\newblock \emph{arXiv preprint arXiv:1905.12614}, 2019{\natexlab{a}}.

\bibitem[Duan et~al.(2019{\natexlab{b}})Duan, Min, Li, Cai, Xu, and
  Ni]{duan2019disentangled}
Duan, Z., Min, M.~R., Li, L.~E., Cai, M., Xu, Y., and Ni, B.
\newblock Disentangled deep autoencoding regularization for robust image
  classification.
\newblock \emph{arXiv preprint arXiv:1902.11134}, 2019{\natexlab{b}}.

\bibitem[Dupont(2018)]{Dup18}
Dupont, E.
\newblock Joint-vae: Learning disentangled joint continuous and discrete
  representations.
\newblock \emph{arXiv preprint arXiv:1804.00104}, 2018.

\bibitem[Eastwood \& Williams(2018)Eastwood and Williams]{EW18}
Eastwood, C. and Williams, C.~K.
\newblock A framework for the quantitative evaluation of disentangled
  representations.
\newblock 2018.

\bibitem[Esmaeili et~al.(2018)Esmaeili, Wu, Jain, Bozkurt, Siddharth, Paige,
  Brooks, Dy, and van~de Meent]{EWJ18}
Esmaeili, B., Wu, H., Jain, S., Bozkurt, A., Siddharth, N., Paige, B., Brooks,
  D.~H., Dy, J., and van~de Meent, J.-W.
\newblock Structured disentangled representations.
\newblock \emph{stat}, 1050:\penalty0 12, 2018.

\bibitem[Gao et~al.(2018)Gao, Brekelmans, Ver~Steeg, and Galstyan]{GBV18}
Gao, S., Brekelmans, R., Ver~Steeg, G., and Galstyan, A.
\newblock Auto-encoding total correlation explanation.
\newblock \emph{arXiv preprint arXiv:1802.05822}, 2018.

\bibitem[Goodfellow et~al.(2014)Goodfellow, Pouget-Abadie, Mirza, Xu,
  Warde-Farley, Ozair, Courville, and Bengio]{GPM14}
Goodfellow, I., Pouget-Abadie, J., Mirza, M., Xu, B., Warde-Farley, D., Ozair,
  S., Courville, A., and Bengio, Y.
\newblock Generative adversarial nets.
\newblock In \emph{Advances in neural information processing systems}, pp.\
  2672--2680, 2014.

\bibitem[Hajek(2015)]{hajek2015random}
Hajek, B.
\newblock \emph{Random processes for engineers}.
\newblock Cambridge university press, 2015.

\bibitem[Heusel et~al.(2017)Heusel, Ramsauer, Unterthiner, Nessler, and
  Hochreiter]{HRU17}
Heusel, M., Ramsauer, H., Unterthiner, T., Nessler, B., and Hochreiter, S.
\newblock Gans trained by a two time-scale update rule converge to a local nash
  equilibrium.
\newblock In \emph{Advances in Neural Information Processing Systems 30}, pp.\
  6629--6640. 2017.

\bibitem[Higgins et~al.(2016)Higgins, Matthey, Pal, Burgess, Glorot, Botvinick,
  Mohamed, and Lerchner]{HMP16}
Higgins, I., Matthey, L., Pal, A., Burgess, C., Glorot, X., Botvinick, M.,
  Mohamed, S., and Lerchner, A.
\newblock beta-vae: Learning basic visual concepts with a constrained
  variational framework.
\newblock 2016.

\bibitem[Higgins et~al.(2017)Higgins, Pal, Rusu, Matthey, Burgess, Pritzel,
  Botvinick, Blundell, and Lerchner]{DARLA}
Higgins, I., Pal, A., Rusu, A.~A., Matthey, L., Burgess, C.~P., Pritzel, A.,
  Botvinick, M., Blundell, C., and Lerchner, A.
\newblock Darla: Improving zero-shot transfer in reinforcement learning.
\newblock \emph{arXiv preprint arXiv:1707.08475}, 2017.

\bibitem[Higgins et~al.(2018{\natexlab{a}})Higgins, Amos, Pfau, Racaniere,
  Matthey, Rezende, and Lerchner]{higgins2018towards}
Higgins, I., Amos, D., Pfau, D., Racaniere, S., Matthey, L., Rezende, D., and
  Lerchner, A.
\newblock Towards a definition of disentangled representations.
\newblock \emph{arXiv preprint arXiv:1812.02230}, 2018{\natexlab{a}}.

\bibitem[Higgins et~al.(2018{\natexlab{b}})Higgins, Sonnerat, Matthey, Pal,
  Burgess, Bo{\v{s}}njak, Shanahan, Botvinick, Hassabis, and Lerchner]{SCAN}
Higgins, I., Sonnerat, N., Matthey, L., Pal, A., Burgess, C.~P., Bo{\v{s}}njak,
  M., Shanahan, M., Botvinick, M., Hassabis, D., and Lerchner, A.
\newblock Scan: Learning hierarchical compositional visual concepts.
\newblock 2018{\natexlab{b}}.

\bibitem[Hsieh et~al.(2018)Hsieh, Liu, Huang, Fei-Fei, and
  Niebles]{hsieh2018learning}
Hsieh, J.-T., Liu, B., Huang, D.-A., Fei-Fei, L.~F., and Niebles, J.~C.
\newblock Learning to decompose and disentangle representations for video
  prediction.
\newblock In \emph{Advances in Neural Information Processing Systems}, pp.\
  517--526, 2018.

\bibitem[Hsu et~al.(2017)Hsu, Zhang, and Glass]{hsu2017unsupervised}
Hsu, W.-N., Zhang, Y., and Glass, J.
\newblock Unsupervised learning of disentangled and interpretable
  representations from sequential data.
\newblock In \emph{Advances in neural information processing systems}, pp.\
  1878--1889, 2017.

\bibitem[Jeon et~al.(2018)Jeon, Lee, and Kim]{IBGAN}
Jeon, I., Lee, W., and Kim, G.
\newblock Ib-gan: Disentangled representation learning with information
  bottleneck gan.
\newblock 2018.

\bibitem[Jeong \& Song(2019)Jeong and Song]{jeong2019learning}
Jeong, Y. and Song, H.~O.
\newblock Learning discrete and continuous factors of data via alternating
  disentanglement.
\newblock \emph{arXiv preprint arXiv:1905.09432}, 2019.

\bibitem[Karaletsos et~al.(2015)Karaletsos, Belongie, and
  R{\"a}tsch]{karaletsos2015bayesian}
Karaletsos, T., Belongie, S., and R{\"a}tsch, G.
\newblock Bayesian representation learning with oracle constraints.
\newblock \emph{arXiv preprint arXiv:1506.05011}, 2015.

\bibitem[Kim \& Mnih(2018)Kim and Mnih]{KM18}
Kim, H. and Mnih, A.
\newblock Disentangling by factorising.
\newblock In \emph{International Conference on Machine Learning}, 2018.

\bibitem[Kingma \& Welling(2013)Kingma and Welling]{KW13}
Kingma, D.~P. and Welling, M.
\newblock Auto-encoding variational bayes.
\newblock \emph{arXiv preprint arXiv:1312.6114}, 2013.

\bibitem[Kulkarni et~al.(2015)Kulkarni, Whitney, Kohli, and
  Tenenbaum]{kulkarni2015deep}
Kulkarni, T.~D., Whitney, W.~F., Kohli, P., and Tenenbaum, J.
\newblock Deep convolutional inverse graphics network.
\newblock In \emph{Advances in neural information processing systems}, pp.\
  2539--2547, 2015.

\bibitem[Kumar et~al.(2017)Kumar, Sattigeri, and
  Balakrishnan]{kumar2017variational}
Kumar, A., Sattigeri, P., and Balakrishnan, A.
\newblock Variational inference of disentangled latent concepts from unlabeled
  observations.
\newblock \emph{arXiv preprint arXiv:1711.00848}, 2017.

\bibitem[Lee et~al.(2018)Lee, Tseng, Huang, Singh, and Yang]{lee2018diverse}
Lee, H.-Y., Tseng, H.-Y., Huang, J.-B., Singh, M., and Yang, M.-H.
\newblock Diverse image-to-image translation via disentangled representations.
\newblock In \emph{Proceedings of the European conference on computer vision
  (ECCV)}, pp.\  35--51, 2018.

\bibitem[Lee et~al.(2020)Lee, Kim, Hong, and Lee]{lee2020high}
Lee, W., Kim, D., Hong, S., and Lee, H.
\newblock High-fidelity synthesis with disentangled representation.
\newblock \emph{arXiv preprint arXiv:2001.04296}, 2020.

\bibitem[Li et~al.(2018)Li, Tang, and He]{li2018unsupervised}
Li, Z., Tang, Y., and He, Y.
\newblock Unsupervised disentangled representation learning with analogical
  relations.
\newblock \emph{arXiv preprint arXiv:1804.09502}, 2018.

\bibitem[Li et~al.(2019)Li, Tang, Li, and He]{li2019learning}
Li, Z., Tang, Y., Li, W., and He, Y.
\newblock Learning disentangled representation with pairwise independence.
\newblock In \emph{Proceedings of the AAAI Conference on Artificial
  Intelligence}, volume~33, pp.\  4245--4252, 2019.

\bibitem[Lin et~al.(2017)Lin, Khetan, Fanti, and Oh]{LKFO17}
Lin, Z., Khetan, A., Fanti, G., and Oh, S.
\newblock Pacgan: The power of two samples in generative adversarial networks.
\newblock \emph{arXiv preprint arXiv:1712.04086}, 2017.

\bibitem[Liu et~al.(2019)Liu, Zhu, Fu, de~Melo, and Elgammal]{liu2019oogan}
Liu, B., Zhu, Y., Fu, Z., de~Melo, G., and Elgammal, A.
\newblock Oogan: Disentangling gan with one-hot sampling and orthogonal
  regularization.
\newblock \emph{arXiv preprint arXiv:1905.10836}, 2019.

\bibitem[Liu et~al.(2015)Liu, Luo, Wang, and Tang]{LLW15}
Liu, Z., Luo, P., Wang, X., and Tang, X.
\newblock Deep learning face attributes in the wild.
\newblock In \emph{Proceedings of the IEEE International Conference on Computer
  Vision}, pp.\  3730--3738, 2015.

\bibitem[Locatello et~al.(2018)Locatello, Bauer, Lucic, Gelly, Sch{\"o}lkopf,
  and Bachem]{LBL18}
Locatello, F., Bauer, S., Lucic, M., Gelly, S., Sch{\"o}lkopf, B., and Bachem,
  O.
\newblock Challenging common assumptions in the unsupervised learning of
  disentangled representations.
\newblock \emph{arXiv preprint arXiv:1811.12359}, 2018.

\bibitem[Locatello et~al.(2019{\natexlab{a}})Locatello, Abbati, Rainforth,
  Bauer, Sch{\"o}lkopf, and Bachem]{locatello2019fairness}
Locatello, F., Abbati, G., Rainforth, T., Bauer, S., Sch{\"o}lkopf, B., and
  Bachem, O.
\newblock On the fairness of disentangled representations.
\newblock In \emph{Advances in Neural Information Processing Systems}, pp.\
  14584--14597, 2019{\natexlab{a}}.

\bibitem[Locatello et~al.(2019{\natexlab{b}})Locatello, Tschannen, Bauer,
  R{\"a}tsch, Sch{\"o}lkopf, and Bachem]{locatello2019disentangling}
Locatello, F., Tschannen, M., Bauer, S., R{\"a}tsch, G., Sch{\"o}lkopf, B., and
  Bachem, O.
\newblock Disentangling factors of variation using few labels.
\newblock \emph{arXiv preprint arXiv:1905.01258}, 2019{\natexlab{b}}.

\bibitem[Lopez et~al.(2018)Lopez, Regier, Jordan, and Yosef]{LRJ18}
Lopez, R., Regier, J., Jordan, M.~I., and Yosef, N.
\newblock Information constraints on auto-encoding variational bayes.
\newblock In Bengio, S., Wallach, H., Larochelle, H., Grauman, K.,
  Cesa-Bianchi, N., and Garnett, R. (eds.), \emph{Advances in Neural
  Information Processing Systems 31}, pp.\  6117--6128. 2018.

\bibitem[Lorenz et~al.(2019)Lorenz, Bereska, Milbich, and
  Ommer]{lorenz2019unsupervised}
Lorenz, D., Bereska, L., Milbich, T., and Ommer, B.
\newblock Unsupervised part-based disentangling of object shape and appearance.
\newblock In \emph{Proceedings of the IEEE Conference on Computer Vision and
  Pattern Recognition}, pp.\  10955--10964, 2019.

\bibitem[Marx et~al.(2019)Marx, Phillips, Friedler, Scheidegger, and
  Venkatasubramanian]{marx2019disentangling}
Marx, C., Phillips, R., Friedler, S., Scheidegger, C., and Venkatasubramanian,
  S.
\newblock Disentangling influence: Using disentangled representations to audit
  model predictions.
\newblock In \emph{Advances in Neural Information Processing Systems}, pp.\
  4498--4508, 2019.

\bibitem[Matthey et~al.(2017)Matthey, Higgins, Hassabis, and
  Lerchner]{dSprites}
Matthey, L., Higgins, I., Hassabis, D., and Lerchner, A.
\newblock dsprites: Disentanglement testing sprites dataset.
\newblock 2017.
\newblock https://github.com/deepmind/dsprites-dataset/.

\bibitem[Miladinovi{\'c} et~al.(2019)Miladinovi{\'c}, Gondal, Sch{\"o}lkopf,
  Buhmann, and Bauer]{miladinovic2019disentangled}
Miladinovi{\'c}, {\DJ}., Gondal, M.~W., Sch{\"o}lkopf, B., Buhmann, J.~M., and
  Bauer, S.
\newblock Disentangled state space representations.
\newblock \emph{arXiv preprint arXiv:1906.03255}, 2019.

\bibitem[Miyato et~al.(2018)Miyato, Kataoka, Koyama, and Yoshida]{MKKY18}
Miyato, T., Kataoka, T., Koyama, M., and Yoshida, Y.
\newblock Spectral normalization for generative adversarial networks.
\newblock In \emph{International Conference on Learning Representations
  (ICLR)}, 2018.

\bibitem[Moreno et~al.(2016)Moreno, Williams, Nash, and Kohli]{MW16}
Moreno, P., Williams, C.~K., Nash, C., and Kohli, P.
\newblock Overcoming occlusion with inverse graphics.
\newblock In \emph{European Conference on Computer Vision}, pp.\  170--185.
  Springer, 2016.

\bibitem[Narayanaswamy et~al.(2017)Narayanaswamy, Paige, van~de Meent,
  Desmaison, Goodman, Kohli, Wood, and Torr]{NPV17}
Narayanaswamy, S., Paige, T.~B., van~de Meent, J.-W., Desmaison, A., Goodman,
  N., Kohli, P., Wood, F., and Torr, P.
\newblock Learning disentangled representations with semi-supervised deep
  generative models.
\newblock In \emph{Advances in Neural Information Processing Systems 30}, pp.\
  5925--5935. 2017.

\bibitem[Pfister et~al.(2018)Pfister, B{\"u}hlmann, Sch{\"o}lkopf, and
  Peters]{PBS18}
Pfister, N., B{\"u}hlmann, P., Sch{\"o}lkopf, B., and Peters, J.
\newblock Kernel-based tests for joint independence.
\newblock \emph{Journal of the Royal Statistical Society: Series B (Statistical
  Methodology)}, 80\penalty0 (1):\penalty0 5--31, 2018.

\bibitem[Pineau \& Lelarge(2018)Pineau and Lelarge]{PL18}
Pineau, E. and Lelarge, M.
\newblock Infocatvae: Representation learning with categorical variational
  autoencoders.
\newblock \emph{arXiv preprint arXiv:1806.08240}, 2018.

\bibitem[Ridgeway(2016)]{Rid16}
Ridgeway, K.
\newblock A survey of inductive biases for factorial representation-learning.
\newblock \emph{arXiv preprint arXiv:1612.05299}, 2016.

\bibitem[Ridgeway \& Mozer(2018)Ridgeway and Mozer]{ridgeway2018learning}
Ridgeway, K. and Mozer, M.~C.
\newblock Learning deep disentangled embeddings with the f-statistic loss.
\newblock In \emph{Advances in Neural Information Processing Systems}, pp.\
  185--194, 2018.

\bibitem[Salimans et~al.(2016)Salimans, Goodfellow, Zaremba, Cheung, Radford,
  and Chen]{SGZ16}
Salimans, T., Goodfellow, I., Zaremba, W., Cheung, V., Radford, A., and Chen,
  X.
\newblock Improved techniques for training gans.
\newblock In \emph{Advances in Neural Information Processing Systems}, pp.\
  2234--2242, 2016.

\bibitem[Schmidhuber(1992)]{Sch92}
Schmidhuber, J.
\newblock Learning factorial codes by predictability minimization.
\newblock \emph{Neural Computation}, 4\penalty0 (6):\penalty0 863--879, 1992.

\bibitem[Singh et~al.(2019)Singh, Ojha, and Lee]{singh2019finegan}
Singh, K.~K., Ojha, U., and Lee, Y.~J.
\newblock Finegan: Unsupervised hierarchical disentanglement for fine-grained
  object generation and discovery.
\newblock In \emph{Proceedings of the IEEE Conference on Computer Vision and
  Pattern Recognition}, pp.\  6490--6499, 2019.

\bibitem[Srivastava et~al.(2017)Srivastava, Valkov, Russell, Gutmann, and
  Sutton]{SVR17}
Srivastava, A., Valkov, L., Russell, C., Gutmann, M., and Sutton, C.
\newblock Veegan: Reducing mode collapse in gans using implicit variational
  learning.
\newblock \emph{arXiv preprint arXiv:1705.07761}, 2017.

\bibitem[Szab{\'o} et~al.(2017)Szab{\'o}, Hu, Portenier, Zwicker, and
  Favaro]{szabo2017challenges}
Szab{\'o}, A., Hu, Q., Portenier, T., Zwicker, M., and Favaro, P.
\newblock Challenges in disentangling independent factors of variation.
\newblock \emph{arXiv preprint arXiv:1711.02245}, 2017.

\bibitem[Szegedy et~al.(2016)Szegedy, Vanhoucke, Ioffe, Shlens, and
  Wojna]{SVI16}
Szegedy, C., Vanhoucke, V., Ioffe, S., Shlens, J., and Wojna, Z.
\newblock Rethinking the inception architecture for computer vision.
\newblock In \emph{Proceedings of the IEEE conference on computer vision and
  pattern recognition}, pp.\  2818--2826, 2016.

\bibitem[Tang et~al.(2013)Tang, Salakhutdinov, and Hinton]{TSH13}
Tang, Y., Salakhutdinov, R., and Hinton, G.
\newblock Tensor analyzers.
\newblock In \emph{International Conference on Machine Learning}, pp.\
  163--171, 2013.

\bibitem[Tschannen et~al.(2018)Tschannen, Bachem, and
  Lucic]{tschannen2018recent}
Tschannen, M., Bachem, O., and Lucic, M.
\newblock Recent advances in autoencoder-based representation learning.
\newblock \emph{arXiv preprint arXiv:1812.05069}, 2018.

\bibitem[van Steenkiste et~al.(2019)van Steenkiste, Locatello, Schmidhuber, and
  Bachem]{van2019disentangled}
van Steenkiste, S., Locatello, F., Schmidhuber, J., and Bachem, O.
\newblock Are disentangled representations helpful for abstract visual
  reasoning?
\newblock In \emph{Advances in Neural Information Processing Systems}, pp.\
  14222--14235, 2019.

\bibitem[Watters et~al.(2019)Watters, Matthey, Burgess, and
  Lerchner]{watters2019spatial}
Watters, N., Matthey, L., Burgess, C.~P., and Lerchner, A.
\newblock Spatial broadcast decoder: A simple architecture for learning
  disentangled representations in vaes.
\newblock \emph{arXiv preprint arXiv:1901.07017}, 2019.

\bibitem[Xing et~al.(2018)Xing, Gao, Han, Zhu, and Wu]{xing2018deformable}
Xing, X., Gao, R., Han, T., Zhu, S.-C., and Wu, Y.~N.
\newblock Deformable generator network: Unsupervised disentanglement of
  appearance and geometry.
\newblock \emph{arXiv preprint arXiv:1806.06298}, 2018.

\end{thebibliography}
\bibliographystyle{icml2020}

\clearpage
\newpage
\onecolumn
\appendix
\section*{Appendix}
\section{Implicit Bias in InfoGAN}
\label{app:infogan}

\noindent
{\bf Practical implementation of InfoGAN loss and the resulting implicit bias.} 
Let $P_{c,x}$ denote the joint distribution of the latent code $c$ and 
 the generated image  $x=G(c,z)$. 
 Two structural assumptions are enforced in \cite{CDH16} 
 to make the maximization of $I(c;x)$ tractable. 
First, to make optimization of the mutual information tractable, 
all practical implementations of 
InfoGAN replace $I(c;x)$ with a 
variational lower bound $\max_{Q_{c|x}} \cL_{\rm Info}(G,Q)$. 
Here $Q_{c|x}$ is an auxiliary conditional distribution, 
which is maximized over the 
InfoGAN regularizer defined as 
\begin{align} 
	&\cL_{\rm Info}(G,Q) \; \triangleq\;  \E_{c\sim P_c, z\sim P_z,x\sim G(z,c)} \left [\log Q_{c|x} \right ] \;,
	\label{eq:InfoGANloss}
\end{align} 
where $P_c$ and $P_z$ denote the distributions chosen as priors. 
When this lower bound is maximized over $Q_{c|x}$, it acts as a surrogate for mutual information:  rearranging the terms gives
\begin{align} 
	\cL_{\rm Info}(G,Q) & \, =  I(c;x) - H(c) - \E_{x\sim P_x} [ d_{\rm KL}( P_{c|x}\| Q_{c|x})  ] \;, \nonumber
\end{align} 
and  $\max_{Q_{c|x}} \cL_{\rm Info}(G,Q) = I(c;x)-H(c)$ (see \cite{CDH16} for a derivation). 
However, this maximization is a functional optimization 
over an infinite dimensional function $Q_{c|x}$, which is intractable. 
To make this tractable, 
the optimization is done over a restricted family of Gaussian distributions in \cite{CDH16}, 
which are factorized (or independent) Gaussian distributions
 (see $\cQ$ in Remark~\ref{rem:InfoGAN}).
$Q_{c|x}$ can then be parametrized by the conditional means and variances, 
which is now  finite dimensional functions, 
and one can use deep neural networks to approximate them. 
Note that  the Shannon entropy $H(c)$ is a constant that 
does not depend on  $G(\cdot,\cdot)$ or $Q_{c|x}$.

 Next, if this maximum over $Q_{c|x}$ has been achieved, then 
 notice that in the generator update, the generator  
 tries to minimize 
 $\min_G  \cL_{\rm Adv}(G,D) - \lambda   (I(c;x)-H(c))$. 
 This is problematic as the $G$ update will increase $I(c,x)$ unboundedly, tending to infinity 
 (even if $Q_{c|x}$ is restricted to factorized Gaussian class). 
The maximum value of $I(c;x)=\infty$ is achieved, for example, 
if $Q_{c_i|x}$ has variance zero for some $i$. 
 This problem is not just theoretical.
In Appendix~\ref{sec:InfoGAN_covariance},  we confirm experimentally that training InfoGAN with an unfactorized $Q_{c|x}$ leads to training instability.
 To avoid such catastrophic failures, \cite{CDH16} forces
 $Q_{c|x}$ to have an identity covariance matrix.  
 This restricts the class of $Q_{c|x}$ that we search over, and 
 forces the $\cL_{\rm Info}(G,Q)$ to be bounded, and the $G$ and $Q_{c|X}$ updates to be well-defined. 
 In summary,  for stability and efficiency of training, 
\cite{CDH16} restricted $Q_{c|X}$ to be a
 {\em factorized Gaussian} with  {\em identity covariance}. 
We show next that this choice creates an implicit bias.

\begin{remark}
	\label{rem:InfoGAN}
	If optimized over  a class of factorized Gaussian conditional distributions with unit variances, 
	i.e.~$\cQ=\{Q_{c|x} \,|\, Q_{c|x}=Q_{c_1|x} \times \cdots \times Q_{c_k|x} \}$ and 
	$Q_{c_i|x} =(1/\sqrt{2\pi})\exp\{-(1/2)(c_i-\mu_i(x))^2\}$ for some $\mu_i(x)\in\reals$ for all $i\in[k]$ and $x\in\cX$, 
	the maximum of the InfoGAN loss in Eq.~\eqref{eq:InfoGANloss} has the following implicit bias:   
	\begin{eqnarray}
		\max_{Q_{c|x} \in \cQ} \cL_{\rm Info}(G, Q) \; = \;I (x;c) -H(c)  - \underbrace{\E\Big[ \log \frac{P_{c|x} }{ N_{c_1|\nu_1(x)}  \cdots  N_{c_k|\nu_k(x)} } \Big]}_{\text{implicit bias}}   \;,
		\label{eq:bias}
	\end{eqnarray}
	where $P_{c,x}$ is the joint distribution between the latent code $c$ and the generated image $x=G(c,z)$, 
	and $N_{c_i|\nu_i(x)} = 1/\sqrt{2\pi}\exp\{ -(1/2)(c_i-\nu_i(x))^2\}$ with $\nu_i=\E_{P_{c_i|x}}[c_i]$ 
	is the best one-dimensional Gaussian approximation of $P_{c_i|x}$.
\end{remark}

We provide a proof in Appendix \ref{sec:InfoGANloss}.
The above implicit bias keeps the loss bounded, so it is necessary. 
On the other hand, it might have undesired and unintended consequences in terms of learning a disentangled deep generative model. 
In this paper,  we therefore introduce a new regularizer  to explicitly encourage disentanglement during  InfoGAN  training.

\noindent{\bf Improving InfoGAN performance via stable training.}
Before introducing our proposed regularizer, 
note that
several VAE-based  architectures 
claim to outperform InfoGAN by significant margins \cite{KM18,HMP16,CLG18}. 
This series of empirical  results has created a misconception that 
InfoGAN is fundamentally limited in achieving disentanglement, which has been reinforced in 
following literature \cite{IBGAN,CHYVAE}, which refer to those initial results.  
We show that 
the previously-reported inferior empirical performance of InfoGAN is due to 
poor architectural and hyperparameter choices in training. 
We take the same implementation reported to have bad performance (disentanglement score of 0.59 in Table~\ref{tbl:comp}). 
We then apply recently-proposed (but now popular) techniques for stabilizing GAN training and 
achieve a performance comparable to the best reported results of competing approaches 
(disentanglement score of 0.83 in Table~\ref{tbl:comp}). 
We provide more supporting experimental  results in Section \ref{sec:CRexp}. 
Concretely, 
we start from the  implementation in \cite{KM18}. 
We then apply spectral normalization to the adversarial loss discriminator \cite{MKKY18}, 
with a choice of cross entropy loss, and 
 use two time-scale update rule \cite{HRU17} with an appropriate choice of the learning rate.  
These choices are motivated by similar choices and  successes of 
\cite{BIGGAN} in scaling GAN to extremely large datasets. 
 Implementation details are in Appendix \ref{app:implementation},  
and we also submit the code for reproducibility. 
Hence, the  starting  point  for our design is to achieve even better  disentanglement than a properly-trained version of InfoGAN.

\subsection{Proof of Remark \ref{rem:InfoGAN}}
\label{sec:InfoGANloss}
Notice that $\cL_{\rm Info}(G,Q) \leq I(c,G(z,c))$.
To understand why this works, let us decompose this lower bound further:
\begin{align*}
&\cL_{\rm Info}(G,Q) \\  
      &= \E_{c\sim P_c, z\sim P_z,x\sim G(z,c)} \left [\log Q_{c|x} \right ] \\
      &= \E_{(x, c) \sim P(x, c)} \left [\log Q_{c|x} \right ] \\
      &= I(x; c) - H(c) + \E_{(x, c) \sim P(x, c)}\bigg[\log\frac{Q_{c|x}P_x}{P_x P_{c|x}}\bigg] \\
      &= I(x; c) - H(c) + \E_{(x, c) \sim P(x, c)}\bigg[\log\frac{Q_{c|x}}{P_{c|x}}\bigg] \\
      &= I(c;x) - H(c) - \E_{x\sim P_x} [ d_{\rm KL}( P_{c|x}\| Q_{c|x})  ]
\end{align*}

We can further simplify and maximize the last term, which is the only one dependent on $Q$ as,
\begin{align*}
&\min_{Q\in\cQ} \; \E_{x\sim P_x} [ d_{\rm KL}( P_{c|x}\| Q_{c|x})] \\ 
&= \min_{Q\in\cQ} \; \E_{c,x\sim P_{c,x} } \Big[ \log\Big( \frac{P_{c|x}}{Q_{c|x}} \Big)  \Big]  \\
&= \min_{Q\in\cQ} \; \E_{c,x\sim P_{c,x} } \Big[ \log\Big( \frac{P_{c|x}}{ N_{c_1|\nu_1(x)}  \cdots  N_{c_k|\nu_k(x)} } \Big)  + \log\Big( \frac{ N_{c_1|\nu_1(x)}  \cdots  N_{c_k|\nu_k(x)} }{Q_{c|x}} \Big) \Big]  \\
&= \E_{c,x\sim P_{c,x} } \Big[ \log\Big( \frac{P_{c|x}}{ N_{c_1|\nu_1(x)}  \cdots  N_{c_k|\nu_k(x)} } \Big) \,\Big]
\;+\; 
\min_{Q\in\cQ} \;  \E_{c,x\sim P_{c,x} } \Big[  \log\Big( \frac{ N_{c_1|\nu_1(x)}  \cdots  N_{c_k|\nu_k(x)} }{Q_{c|x}} \Big) \Big]  \\
& = \E_{c,x\sim P_{c,x} } \Big[ \log\Big( \frac{P_{c|x}}{ N_{c_1|\nu_1(x)}  \cdots  N_{c_k|\nu_k(x)} } \Big) \,\Big]\;,
\end{align*}
where the last equality follows from the fact that 
any $Q\in\cQ$ can be parametrized by $(\mu_1(x),\ldots,\mu_k(x) )$ as 
$Q_{c|x} = Q_{c_1|x} \cdots Q_{c_k|x}$ with 
$Q_{c_i|x} = (1/\sqrt{2\pi}) \exp \{-(1/2)(c_i - \mu(x))^2\}$, in which case 
\begin{eqnarray*}
	\min_{Q\in\cQ} \;  \E_{c,x\sim P_{c,x} } \Big[  \log\Big( \frac{ N_{c_1|\nu_1(x)}  \cdots  N_{c_k|\nu_k(x)} }{Q_{c|x}} \Big) \Big]  
	&=& 	\min_{\{\mu_i(x)\}_{i=1}^k} \;  \E_{x\sim P_{x} } \Big[ \sum_{i\in[k]} \big( \mu_i(x) - \nu_i(x) \big)^2 \Big] \\
	&\geq & 0 \;,
\end{eqnarray*}
and the minimum of zero can be achieved by $\mu_i(x) = \nu_i(x)$, where $\nu_i(x) = \E_{P_{c|x}}[c_i]$.

\section{InfoGAN with Non-factorizing Decoder}
\label{sec:InfoGAN_covariance}
\label{sec:InfoGANfullcov}
In this section we verify that the InfoGAN trained with non-factorizing multi-variate Gaussian distribution $Q(c|x)$ is unstable. Specifically, we train an InfoGAN with a decoder distribution of $Q(c | x) = \mathcal{N}(\mu(x), \Sigma(x))$, where $\mathcal{N}$ is the multivariate Gaussian distribution with mean $\mu(x) \in \reals^{k}$, and full covariance matrix $\Sigma(x) \in \reals^{k \times k}$. These parameters of the distribution are modelled as neural network functions of $x$, where we explicitly enforce the positive semi-definiteness of $\Sigma(x)$. This is less restrictive than the factorizing decoder distribution, $Q(c|x) = \prod_{i \in [k]} Q(c_i|x) = \prod_{i \in [k]} \mathcal{N}(\mu_i(x), 1)$ with its fixed diagonal covariance matrix $\id$, in the standard InfoGAN (see Remark \ref{rem:InfoGAN}).

\begin{figure}[h]
  \centering
  \vspace{-0.3cm}
  \includegraphics[width=2.4in]{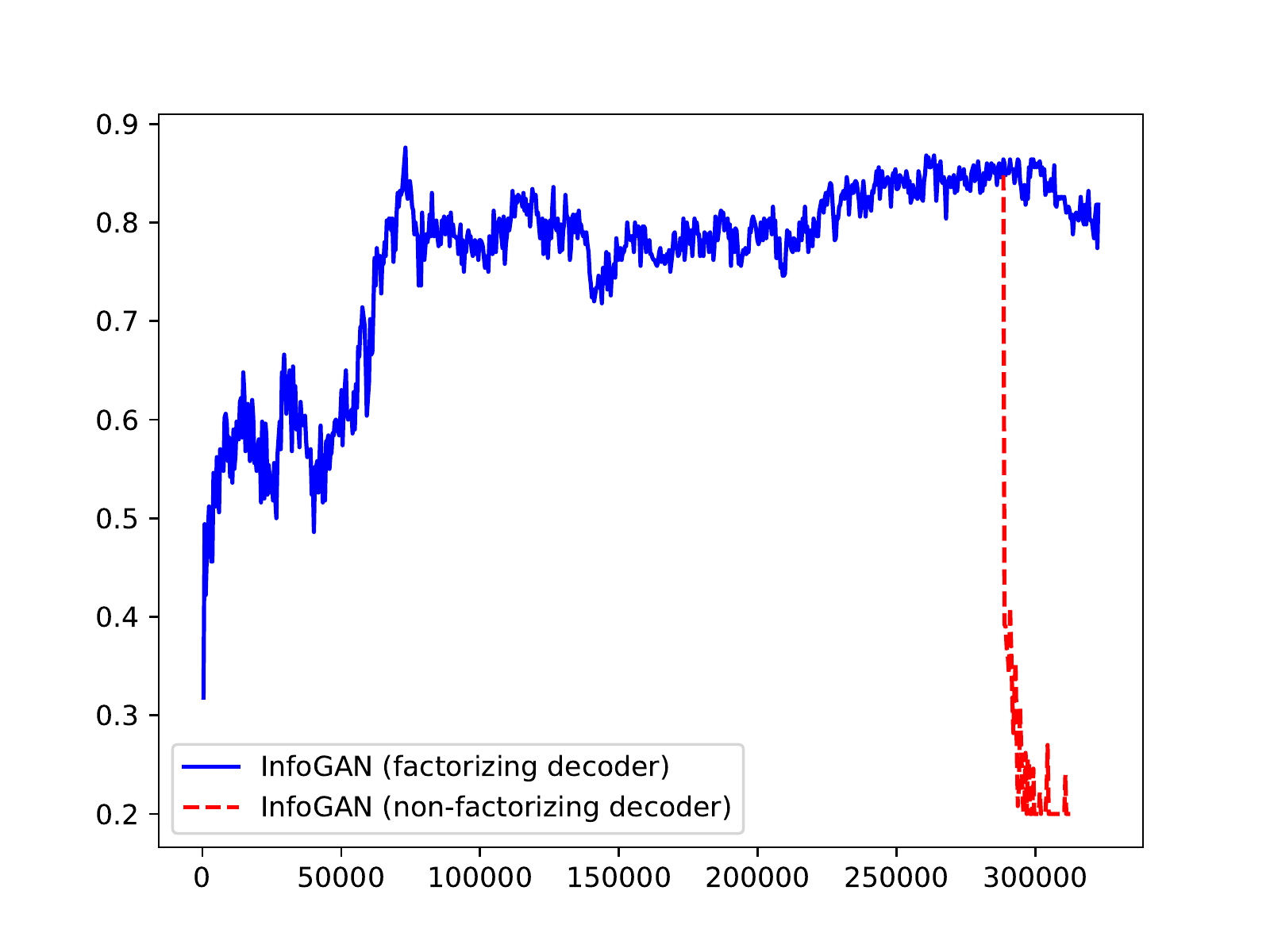}
  \put(-145,-5){Iterations (\# of minibatches)}
  \put(-175,30){\rotatebox{90}{Disentanglement}}
  \caption{After 288,000 iterations, we continue training an InfoGAN with/without the non-factorizing decoder distribution $Q(c|x)$ (Section~\ref{sec:InfoGANfullcov}). Non-factorizing decoder is high unstable and its training terminates early after reaching a low values of $0.2$.}
  \label{fig:split_fullcov}
\end{figure}

Figure~\ref{fig:split_fullcov} shows the degradation in disentanglement due to the non-factorizing decoder. Similar to the experiment in Figure~\ref{fig:split}, we first train the standard InfoGAN ($\lambda = 0.2$) with factorizing decoder distribution $Q(c|x) = \prod_{i \in [k]} \mathcal{N}(\mu_i(x), 1)$ on \dsprites{} dataset for 25 epochs (288,000 batches) and from this point onwards we train two different versions of this model for 3 more epochs: one where we continue training the standard InfoGAN (blue solid curve) and another version where we replace the non-factorizing decoder distribution with the factorizing decoder distribution $Q(c | x) = \mathcal{N}(\mu(x), \Sigma(x))$ (red dashed curve). We plot the FactorVAE score, as defined in Section~\ref{app:metrics}.

We see that the non-factorizing decoder is highly unstable and its performance drops (from that of factorizing decoder) to $0.2$ (minimum possible value for 5 latent codes) and its training terminates early because the learnt covariance matrix $\Sigma(x)$ becomes rank deficient.

\section{Proof of Remark~\ref{rem:js}}
\label{sec:proof_rem2}

\begin{proof}
The solution to the following optimization problem is  
$H(x,x') =  (1/Z_{x,x'}) \begin{bmatrix} Q^{(1)}(x,x') &, \cdots ,& Q^{(k)}(x,x') \end{bmatrix}  $ with a normalizing constant 
  $Z_{x,x'}=\sum_{i\in[k]} Q^{(i)}(x,x')$. 
\begin{eqnarray*} 
  \arg \max_H &&  \E_{I \sim {\rm Unif}([k]),(x,x')\sim Q^{(I)}} [  \langle I \,,\, \log H(x,x') \rangle ] \;, \\
  \text{subject to} && \langle \ones , H(x,x') \rangle =1 \text{, for all } (x,x')\;.
\end{eqnarray*} 
This follows from the fact that the gradient of the Lagrangian is 
$Q^{(i)}(x,x')/H_i(x,x')  - \mu_{x,x'} $ where $\mu_{x,x'}$ is the Lagrangian multiplier, 
and setting it to zero gives 
the desired maximizer. 
Plugging this back into the objective function, we get that 
\begin{align*}
&\max_{||H(x,x')||_1 = 1}\; \mathbb{E} [  \langle I \,,\, \log H(x,x') \rangle ] \\
&\;\;= \frac1k \sum_{i \in [k]} \mathbb{E}_{(x,x') \sim Q^{(i)}}\Bigg[\log \frac{Q^{(i)}(x,x')/k}{\sum_{j \in [k]} Q^{(j)}(x,x')/k}\Bigg] \nonumber \\
&= \frac1k \sum_{i \in [k]} d_\mathrm{KL} \Bigg( Q^{(i)} \,\Big|\Big|\, \frac{\sum_{j \in [k]} Q^{(j)}}k \Bigg) - \log k \nonumber \\
&= d_\mathrm{JS}(Q^{(1)}, \cdots ,Q^{(k)}) - \log k\,.
\end{align*}
\end{proof}

\red{
\section{Implementation Details of Contrastive  Latent Sampling}
\label{app:latent-detail}
The steps for sampling $c^1$ and $c^2$ given a contrastive gap $g$ is as follows:
\begin{enumerate}
	\item Sample $I$ from Uniform$\{1,2,...,k\}$, the index of the fixed latent.
	\item Sample the latent value $c^1_I$ = $c^2_I$ from Uniform$[-1, 1]$.
	\item For any other index, $j$ (not $I$):
	\begin{enumerate}
		\item Independently sample $c^1_j$ and $c^2_j$ from Uniform$[-1 + g/2, 1 - g/2]$.
		\item If $c^1_j$ is greater than $c^2_j$, add $g/2$ to $c^1_j$, subtract $g/2$ from $c^2_j$; else subtract $g/2$ from $c^1_j$, add $g/2$ to $c^2_j$.
	\end{enumerate}
\end{enumerate}
This ensures that we uniformly sample from the latent space where $c^1_I=c^2_I$ and $\min_{j \in \{1,2,...,k\}\setminus \{I\}} |c^1_j - c^2_j| \geq g$. 
}

\section{Evaluating Performance}
\label{app:metrics}

We use the following metrics to evaluate various aspects of the trained latent representation:  
disentanglement, independence,  and generated image quality. 

\paragraph{Disentanglement} We use the popular disentanglement metric proposed in \cite{KM18}. 
This metric is defined for datasets with known ground truth factors and is computed as follows. 
First, generate data points where the $i$th factor is fixed, but the other factors are varying uniformly at random, for a randomly-selected $i$. 
Pass each sample through the encoder, and normalize each resulting dimension by its empirical variance. 
Take the resulting dimension $j$ with the smallest variance. 
In a setting with perfect disentanglement, the variance in the $j$th dimension would be 0.
Each sample's encoding generates a `vote' $j$; we take the majority vote as the final output of the classifier; if the classifier is correct, it should map to $i$. 
The disentanglement metric is the error rate of this classifier, taken over many independent trials of this procedure. 
In our experiment, for each fixed factor index $i$, we generate 100 groups of images, where each group has 100 images with the same value at the $i$th index. 

One challenge is computing the disentanglement metric for FactorVAE when trained with more latent codes $k$ than there are latent factors (let $\hat k$ denote the true number of factors). 
For instance, \cite{KM18} uses $c\in \mathbb R^{10}$ for datasets with only five latent codes. 
To account for this, \cite{KM18} first removes all latent codes that have collapsed to the prior (i.e., $Q_{c_j|x}=P_{c_j}$); they then use the majority vote on the remaining factors.
However, this approach can artificially change the metric if the number of factors for which the posterior does not equal the prior does not equal $\hat k$. 
Hence to measure the metric on FactorVAE (or more generally, cases where $k>\hat k$) on \teapots{} dataset, we first compute the $k \times \hat k$ metric matrix, find the maximum value of each row. 
We then take the top $\hat k$ among the $k$ maximum row values, and sum them up.

We additionally compute the (less common) disentanglement metric of \cite{EW18} (DCI). 
This metric first requires an estimate of the disentangled code $\tilde c$ from samples, for which we use our encoder.
Next, we train a regressor $f$ to predict ground truth code $\hat{c}$ from generated code $\tilde c$, so $\hat{c} = f(\tilde c)$. 
These regressors must also provide a matrix of relative importance $R$, such that $R_{ij}$ denotes the relative importance of $\tilde c_i$ in predicting $\hat c_j$.
Because of this requirement, Eastwood and Williams propose using regressors that provide importance scores, such as LASSO and random forests.
These restrictions limit the generality of the metric; nonetheless, we include it for completeness.
Given the relative importance $R$, a disentanglement score can be computed (see \cite{EW18}  for details).

Eastwood and Williams disentanglement metric is computed using the random forest regressor \cite{EW18}\footnote{https://github.com/cianeastwood/qedr/blob/master/quantify.ipynb}, as implemented in the \texttt{scikit-learn} library\footnote{https://scikit-learn.org/stable/modules/generated/sklearn.ensemble.RandomForestRegressor.html}.%
For \dsprites{} experiments, we use default values for all parameters, except for the \texttt{max\_depth} parameter for which we use the values: 4, 2, 4, 2, and 2, for the latent factors: shape, scale, rotation, $x$-position, and $y$-position respectively, as used by the IB-GAN paper \cite{IBGAN} (as per a private communication with its authors).
For \teapots{} experiments, we use the default cross-validation version of random forest regressor.

The other metrics we compute are 
SAP \cite{kumar2017variational}, 
Explicitness \cite{ridgeway2018learning},
Modularity \cite{ridgeway2018learning},
MIG \cite{CLG18}, and
BetaVAE \cite{HMP16}.

\paragraph{$d$-Variable Hilbert-Schmidt Independence Score (dHSIC)} The dHSIC score is an empirical, kernel-based measure of the total correlation of a multivariate random variable from samples \cite{PBS18}. 
It is used in \cite{LRJ18} to enforce independence among latent  factors.  
We use this metric to understand whether and how disentanglement is correlated with total correlation. 
Suppose we have $c\in\mathbb R^k$.
We want to compute the distance of the distribution $P_c$ from the product distribution of the marginals $\prod_i P_{c_i}$.
The dHSIC score over $n$ samples is computed as follows.
Consider a Gaussian kernel \cite{aizerman1964theoretical} with a median heuristic for bandwidth:
$$K_h(x_i,x_j) \;\;=\;\; e^{-\frac{\|x_i-x_j\|}{h^2}}\;,$$ 
where $h={\rm Median}(\{ \|x_i-x_j\| \}_{i\neq j})$. 
When there are $k$ latent codes $c_1,\ldots, c_k$ and $n$ samples, we have
\begin{align*}
 &d{\rm HSIC}_n \;=\; \frac{1}{n^2} \sum_{i, j\in[n]} \left ( \prod_{\ell=1}^k K_{h_{\ell}}(c_{\ell i},c_{\ell j})  \right )\\
& \;+\; \frac{1}{n^{2k}} \prod_{\ell=1}^k \sum_{i, j \in [n]}   K_{h_\ell}(c_{\ell i},c_{\ell j}) 
 \; -\;  \frac{2}{n^{k +1}}  \sum_{i\in[n] }  \prod_{\ell=1}^k
\sum_{j} K_{h_\ell}(c_{\ell i},c_{\ell j})   \;.
 \end{align*}

\paragraph{Image Quality}
\textit{Inception score} was first proposed in \cite{SGZ16} for labelled data, and is  computed as $\exp(\E_x d_{\rm KL}(p(y|\boldsymbol x) || p(y)))$, where 
$d_{\rm KL}$ denotes the Kullback-Liebler divergence of two distributions. 
The distribution $p(y|\boldsymbol x)$ was originally designed to be used with the Inception network \cite{SVI16}, but we instead use a pre-trained classifier for the dataset at hand. 
Notice that this metric does not require any information about the disentangled representation of a sample. 
Inception score is widely used in the GAN literature to evaluate data quality.

 Intuitively, inception score measures a combination of two effects: mode collapse and individual sample quality. 
To tease apart these effects, we compute two additional metrics. 
The first is \emph{reverse KL-divergence}, proposed in \cite{SVR17} to measure mode collapse in labelled data.
It measures the KL-divergence between the generated label distribution and the true distribution. 
The second is \emph{classifier confidence}, which we use as a proxy for sample quality.
Classifier confidence is measured as the max of the softmax layer of a pre-trained classifier; the higher this value, the more confident the classifier is in its output, which suggests the image quality is better.

\section{\dsprites{} Dataset}
\label{app:dsprites}

\begin{wrapfigure}{r}{0.2\textwidth}
	\centering
	\includegraphics[width=0.2\textwidth]{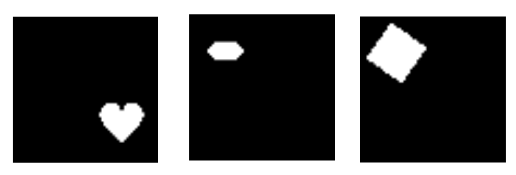}
	\caption{Example images from the \dsprites{} dataset.}
	\label{fig:dsprites}
\end{wrapfigure}

We begin with the synthetic \dsprites{} dataset \cite{dSprites}, commonly used to numerically compare disentangling methods. 
The dataset consists of 737,280 binary 64 $\times$ 64 images of 2D shapes generated from five ground truth independent latent factors: color, shape, scale, rotation, x and y position. 
All  combinations of latent factors are present in the dataset; some examples are
 illustrated in Figure \ref{fig:dsprites}.
Figure \ref{fig:latenttraversal} illustrates the latent traversal for \gannosp.
To generate this figure, we fix all latent factors except one $c_i$, and vary $c_i$ from -1 to 1 in evenly-spaced intervals. 
We observe that each of the five empirically-learned factors captures one true underlying factor, and the traversals span the full range of possibilities for each hidden factor.

\subsection{Details of Table \ref{tbl:comp} and Table \ref{tbl:MCdsprites}}

In Table~\ref{tbl:detailed1} (combining Table \ref{tbl:comp} and Table \ref{tbl:MCdsprites}), we show our main result of how InfoGAN-CR performs both $(a)$ 
when hyper-parameters are tuned via supervised selection with oracle access to a synthetic dataset generated with ground truth disentangled latent codes available to us, 
and $(b)$ with model selected with our proposed ModelCentrality. We use the following popular metrics: 
FactorVAE \cite{KM18}, DCI \cite{EW18}, SAP \cite{kumar2017variational}, Explicitness \cite{ridgeway2018learning}, Modularity \cite{ridgeway2018learning}, MIG \cite{CLG18}, and BetaVAE \cite{HMP16}. For baseline scores, $^\dagger$ indicates reported scores we copy from \cite{IBGAN}, $^\ddagger$ from \cite{CLG18}, $^\star$ from \cite{kumar2017variational}, $^\ast$ from \cite{CHYVAE}, and $^\triangle$ from \cite{KM18}. All the rest of the scores are run by us. 
The DCI metric uses random forest whose parameter is selected according to IB-GAN \cite{IBGAN}. DCI metric of some FactorVAE runs has NaN values, which we ignore when computing the mean and standard error. UDR and our model selection randomly select 80\% of other models for computing cross-metrics; and the results are averaged by 100 such random selection trials.

\begin{sidewaystable}
    \begin{center}
        \begin{tabular}{| c | c | l | l | l | l | l | l| l|}
            \hline
            & \textbf{Model} 
            & \textbf{FactorVAE} 
            & \textbf{DCI} 
            & \textbf{SAP} 
             & \textbf{Explicitness} 
            & \textbf{Modularity} 
            & \textbf{MIG} 
            & \textbf{BetaVAE} 
            \\ \hline
            
            \multirow{14}{*}{\textbf{VAE}}
            &VAE & 
                $0.63 \pm 0.06^\dagger$  & $0.30 \pm 0.10^\dagger$  &&&& $0.10^\ddagger$ & \\
            &$\beta$-TCVAE & 
                $0.62 \pm0.07^\dagger$ & $0.29 \pm 0.10^\dagger$  &&&& $0.45^\ddagger$   &\\
            &HFVAE & 
                $0.63 \pm 0.08^\dagger$ & $0.39 \pm 0.16^\dagger$  &&&& &\\
            &$\beta$-VAE & 
                $0.63 \pm0.10^\dagger$  & $0.41 \pm 0.11^\dagger$  & $0.55^\star$ &&& $0.21^\ddagger$ &\\
            &CHyVAE & 
                $0.77^\ast$  & &&&&&\\
            &DIP-VAE &
                & &  $0.53^\star$ &&&&\\
            &FactorVAE & 
                $0.82^\triangle$  &&&&& $0.15^\ddagger$  &\\
            &FactorVAE (1.0)&  $0.79 \pm 0.01$  &  $0.67 \pm 0.03$  &  $0.47 \pm 0.03$  &  $0.78 \pm 0.01$  &  $0.79 \pm 0.01$  &  $0.27 \pm 0.03$  &  $0.79 \pm 0.02$\\
            &FactorVAE (10.0)&  $0.83 \pm 0.01$  &  $0.70 \pm 0.02$  &  $0.57 \pm 0.00$  &  $0.79 \pm 0.00$  &  $0.79 \pm 0.00$  &  $0.40 \pm 0.01$  &  $0.83 \pm 0.01$\\
            &FactorVAE (20.0)&  $0.83 \pm 0.01$  &  $0.72 \pm 0.02$  &  $0.57 \pm 0.00$  &  $0.79 \pm 0.00$  &  $0.79 \pm 0.01$  &  $0.40 \pm 0.01$  &  $0.85 \pm 0.00$\\
            &FactorVAE (40.0)&  $0.82 \pm 0.01$  &  $0.74 \pm 0.01$  &  $0.56 \pm 0.00$  &  $0.79 \pm 0.00$  &  $0.77 \pm 0.01$  &  $0.43 \pm 0.01$  &  $0.84 \pm 0.01$\\
            & FactorVAE (UDR Lasso)&  $0.81 \pm 0.00$  &  $0.70 \pm 0.01$  &  $0.56 \pm 0.00$  &  $0.79 \pm 0.00$  &  $0.78 \pm 0.00$  &  $0.40 \pm 0.00$ &  $0.84 \pm 0.00$\\
            & FactorVAE (UDR Spearman)&  $0.79 \pm 0.00$  &  $0.73 \pm 0.00$  &  $0.53 \pm 0.01$  &  $0.79 \pm 0.00$  &  $0.77 \pm 0.00$  &  $0.40 \pm 0.01$ &  $0.79 \pm 0.00$\\
            & FactorVAE (our model selection) &  $0.84 \pm 0.00$  &  $0.73 \pm 0.01$  &  $0.58 \pm 0.00$  &  $0.80 \pm 0.00$  &  $0.82 \pm 0.00$  &  $0.37 \pm 0.00$  &  $0.86 \pm 0.00$\\\hline
                
            \multirow{7}{*}{\textbf{GAN}}
            & InfoGAN & 
                $0.59 \pm 0.70^\dagger$  & $0.41 \pm 0.05^\dagger$  &&&& $0.05^\ddagger$ &  \\
            & IB-GAN & 
                $0.80 \pm 0.07^\dagger$ & $0.67 \pm 0.07^\dagger$ &&&& &\\
            & InfoGAN (modified) & 
                $0.82 \pm 0.01$  &  $0.60 \pm 0.02$  &  $0.41 \pm 0.02$  &  $0.82 \pm 0.00$  &  $0.94 \pm 0.01$  &  $0.22 \pm 0.01$  &  $0.87 \pm 0.01$\\
            & InfoGAN-CR& 
                $0.88 \pm 0.01$  &  $0.71 \pm 0.01$  &  $0.58 \pm 0.01$  &  $0.85 \pm 0.00$  &  $0.96 \pm 0.00$  &  $0.37 \pm 0.01$  &  $0.95 \pm 0.01$\\
            & InfoGAN-CR (UDR Lasso)&  $0.86 \pm 0.01$  &  $0.68 \pm 0.01$  &  $0.49 \pm 0.01$  &  $0.84 \pm 0.00$  &  $0.96 \pm 0.00$  &  $0.30 \pm 0.01$ & $0.92 \pm 0.01$\\
            & InfoGAN-CR (UDR Spearman)&  $0.84 \pm 0.01$  &  $0.67 \pm 0.01$  &  $0.53 \pm 0.01$  &  $0.84 \pm 0.00$  &  $0.96 \pm 0.00$  &  $0.31 \pm 0.01$  &  $0.90 \pm 0.01$\\
            & InfoGAN-CR (our model selection)&  $\bf 0.92 \pm 0.00$  &  $\bf 0.77 \pm 0.00$  &  $\bf 0.65 \pm 0.00$  &  $\bf 0.87 \pm 0.00$  &  $\bf 0.99 \pm 0.00$  &  $\bf 0.45 \pm 0.00$ & $\bf 0.99 \pm 0.00$\\\hline
        \end{tabular}
    \end{center}
    \caption{ Contrastive regularizer, together with ModelCentrality for model selection, achieves significant improvements over baseline approaches on benchmark datasets and on popular disentanglement metrics.  }
    \label{tbl:detailed1}
\end{sidewaystable}

\begin{figure}[h]
	\centering
	\includegraphics[width=3in]{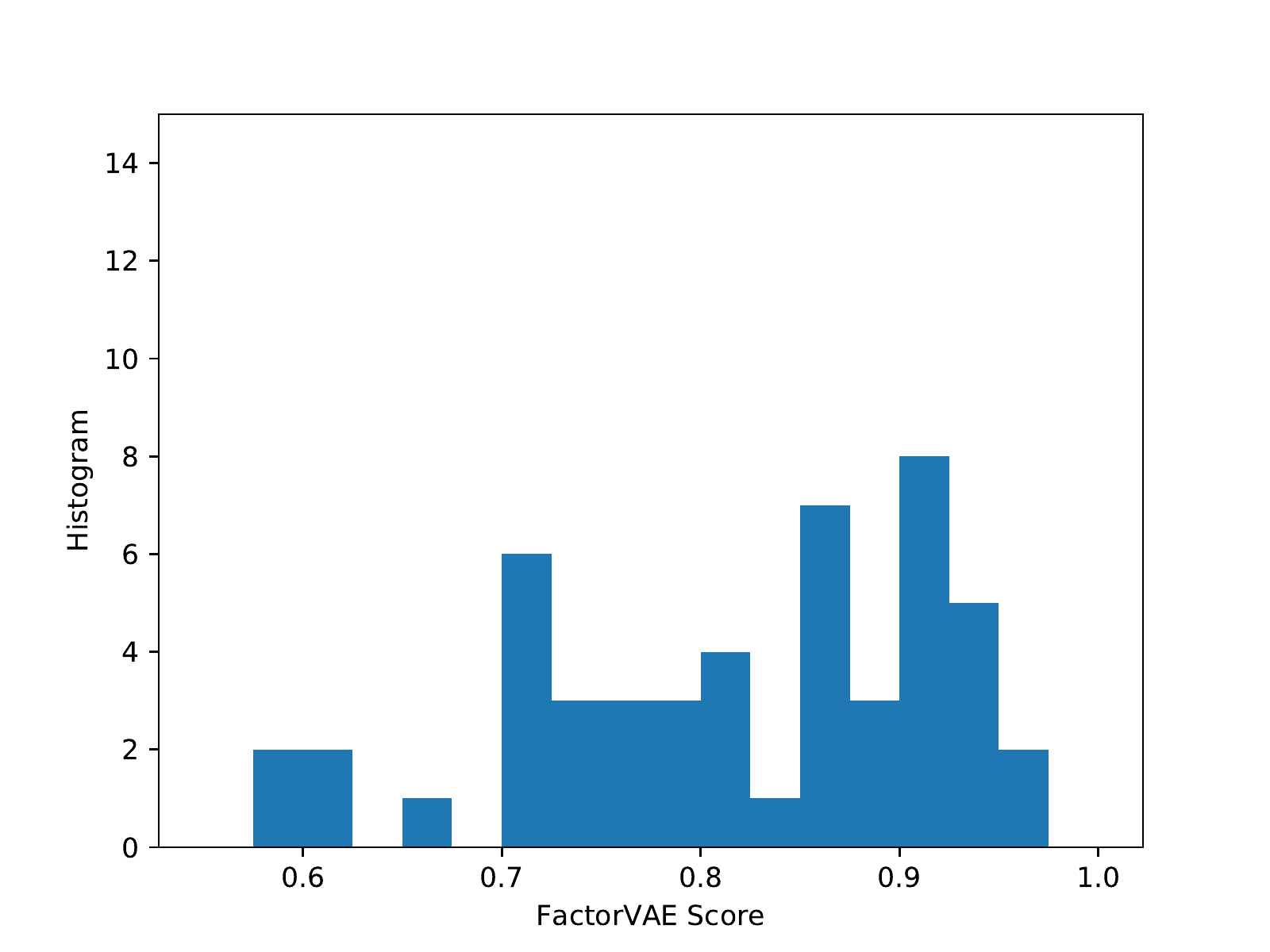}
	\caption{ Histogram of FactorVAE scores of InfoGAN (modified). 
	}
	\label{fig:factorvaehist-infogan}
\end{figure}

\begin{figure}[h]
	\centering
	\includegraphics[width=3in]{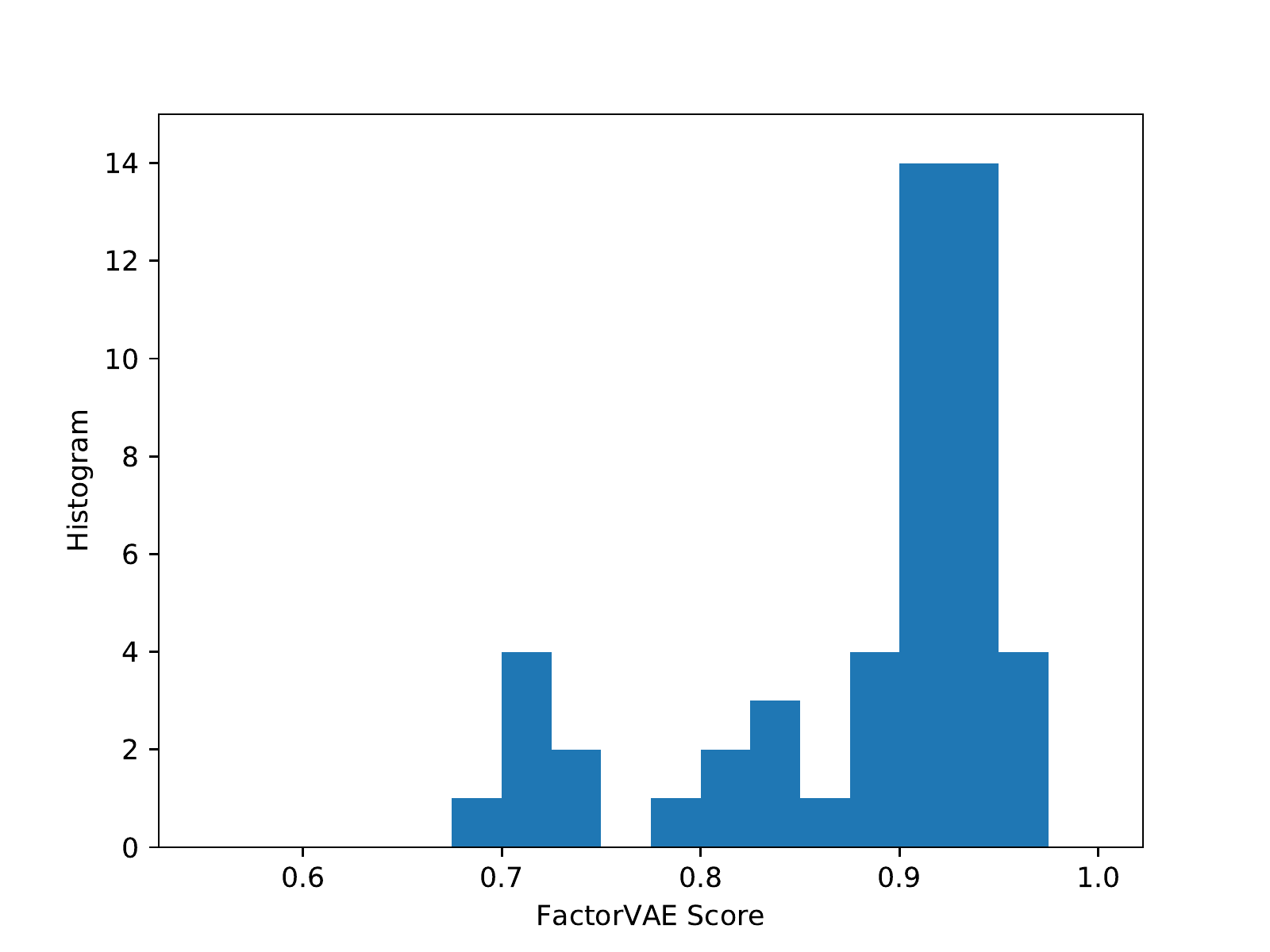}
	\caption{ Histogram of FactorVAE scores of InfoGAN-CR. 
	}
	\label{fig:factorvaehist-infogancr}
\end{figure}

\red{
\subsection{Discussion about the Reproducibility of Table \ref{tbl:comp}}
\label{app:dsprites-reproducibility}
An earlier version of out paper reports a slightly higher scores for both InfoGAN (modified) and \gan 
than the current Table \ref{tbl:comp} (e.g. the FactorVAE score for InfoGAN (modified) and \gan were $0.83 \pm 0.03$ and $0.90 \pm 0.01$).  
These results were averaged over 12 runs, which is not large enough. 
Later, we observed that 
rarely (but with a positive probability) the trained model has very bad disentanglement performance during the training of InfoGAN, and does not recover after CR is introduced. 
To make our results reproducible by anyone who uses our code from our github repo, we ran fresh 50 runs with exactly the same code and hyper-parameters. 
This is reported in the current Table~\ref{tbl:comp} and Table~\ref{tbl:detailed1}. Also, the saved models of these 50 runs are available in our github repo, 
to help anyone verify the reproducibility of our experiments.  
The histograms of FactorVAE scores of InfoGAN (modified) and InfoGAN-CR are in Figure \ref{fig:factorvaehist-infogan} and Figure \ref{fig:factorvaehist-infogancr}.
}

\subsection{Parameter Exploration} \label{app:parameters}
\begin{figure}[h]
	\centering
	\includegraphics[width=3in]{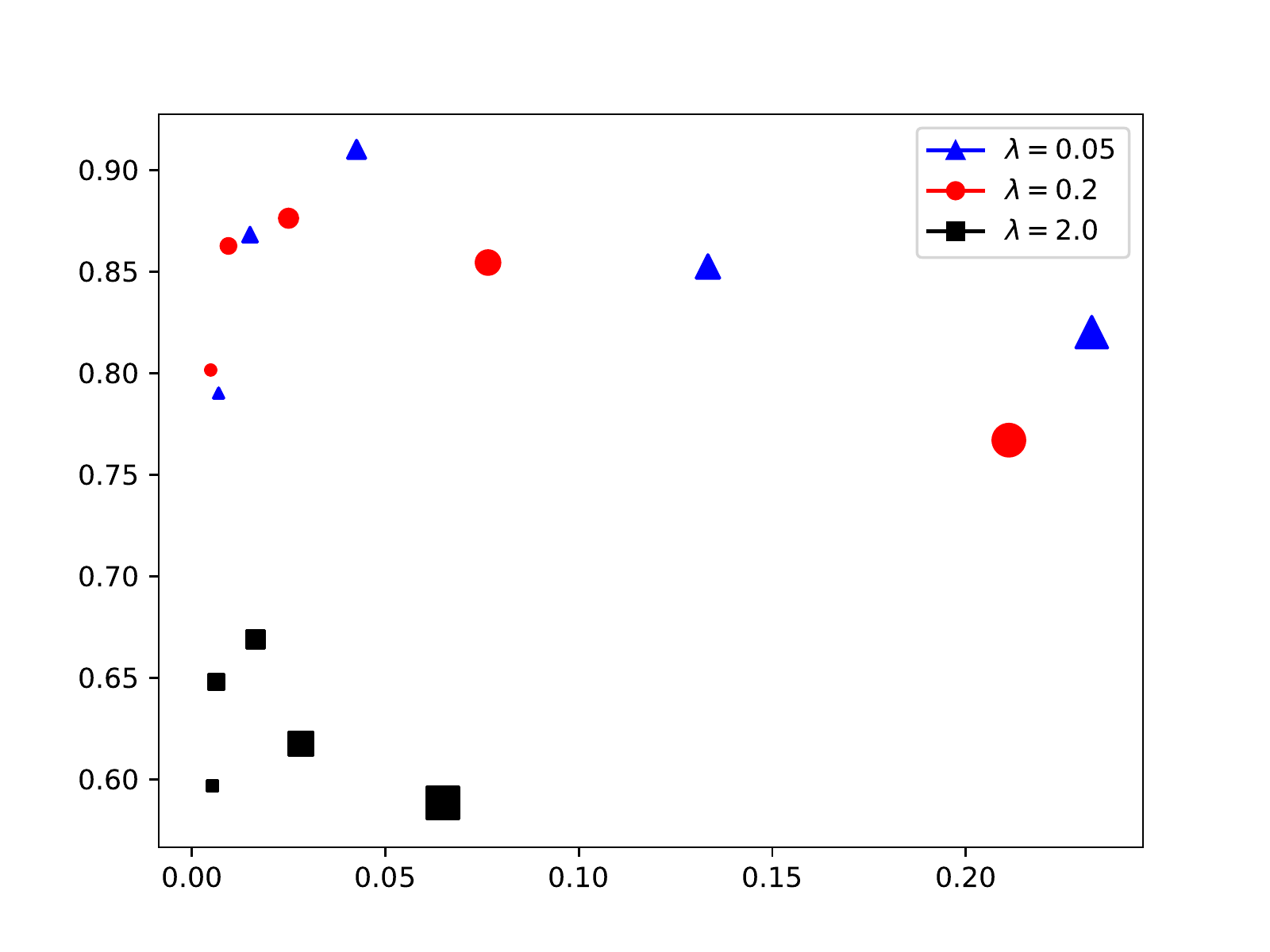}
	\put(-165,-5){Mode-Reversed KL-Divergence}
	\put(-220,50){\rotatebox{90}{Disentanglement}}
	\caption{ Achievable (mode-reversed KL-divergence, disentanglement) pairs for \gannosp, where disentanglement is measured as in \cite{KM18}. We vary the contrastive regularizer $\alpha$ and the InfoGAN regularizer.
	The size of each point denotes $\alpha$, ranging from smallest to largest for $\alpha \in \{0,1,2,4,8\}$. 
	}
	\label{fig:varying_alpha_kl}
\end{figure}

\begin{figure}[h]
	\centering
	\includegraphics[width=3in]{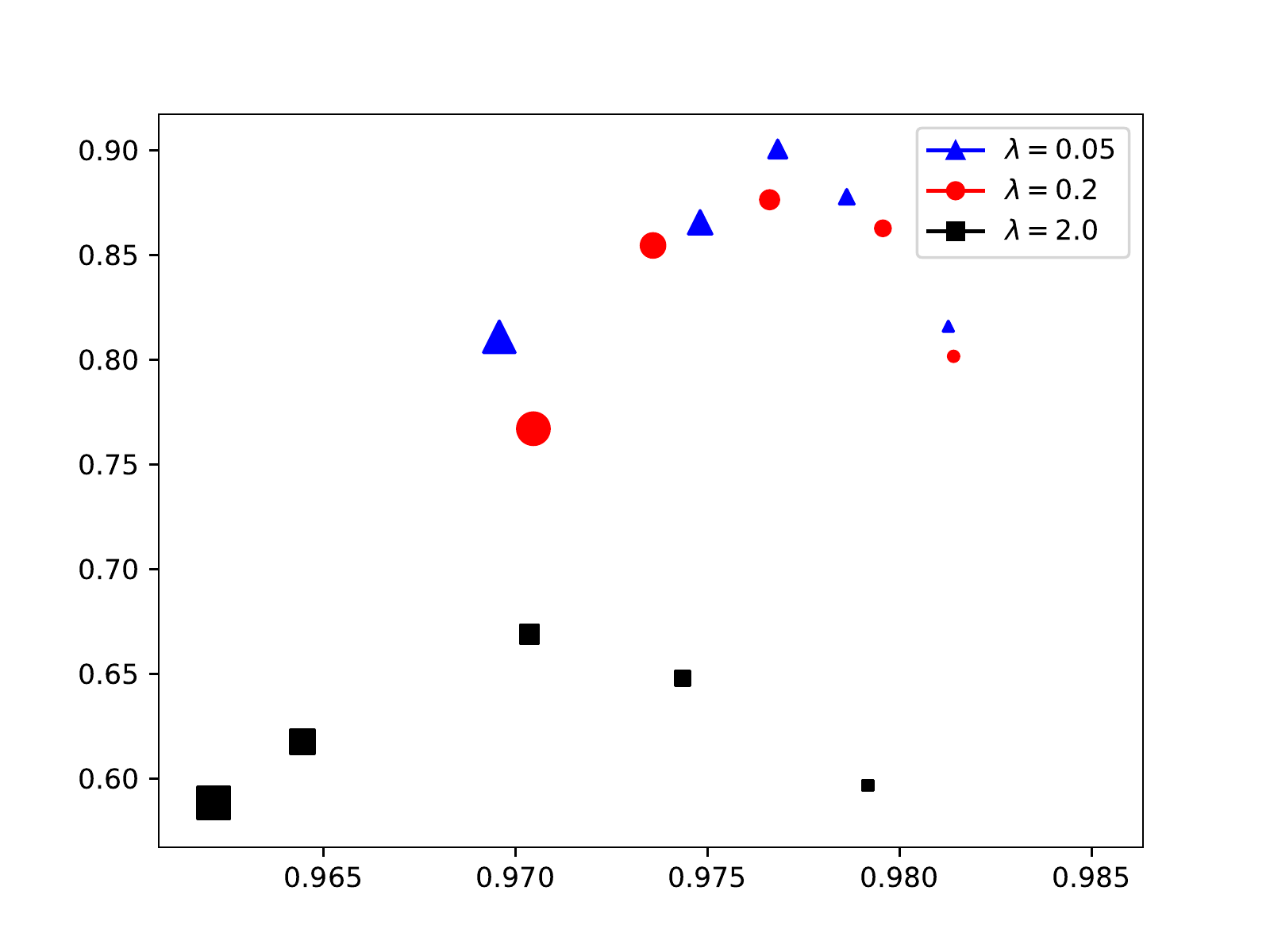}
	\put(-145,-5){Classifier confidence}
	\put(-220,50){\rotatebox{90}{Disentanglement}}
	\caption{ Achievable (classifier confidence, disentanglement) pairs for \gannosp, where disentanglement is measured as in \cite{KM18}. We vary the contrastive regularizer $\alpha$ and the InfoGAN regularizer.
	The size of each point denotes $\alpha$, ranging from smallest to largest for $\alpha \in \{0,1,2,4,8\}$. 
	}
	\label{fig:varying_alpha_confidence}
\end{figure}

To better understand the parameter exploration in Figure \ref{fig:varyalpha}, we generate similar plots, except representing image quality by mode-reversed KL-divergence (Figure \ref{fig:varying_alpha_kl}) and classifier confidence (Figure \ref{fig:varying_alpha_confidence}).

For both reverse KL-divergence and classifier confidence, we observe similar trends to Figure \ref{fig:varyalpha};
increasing $\alpha$ improves disentanglement to a point, whereas it appears to hurt both image quality metrics.
One observation is that for $\alpha\in\{0,1\}$, there is little noticeable change in the either image quality metric.
This suggests that CR does not introduce mode collapse or substantial reductions in image quality for small $\alpha$.

\subsection{Total Correlation}\label{app:tc}

\begin{figure}[h]
	\centering
		\includegraphics[width=2.5in]{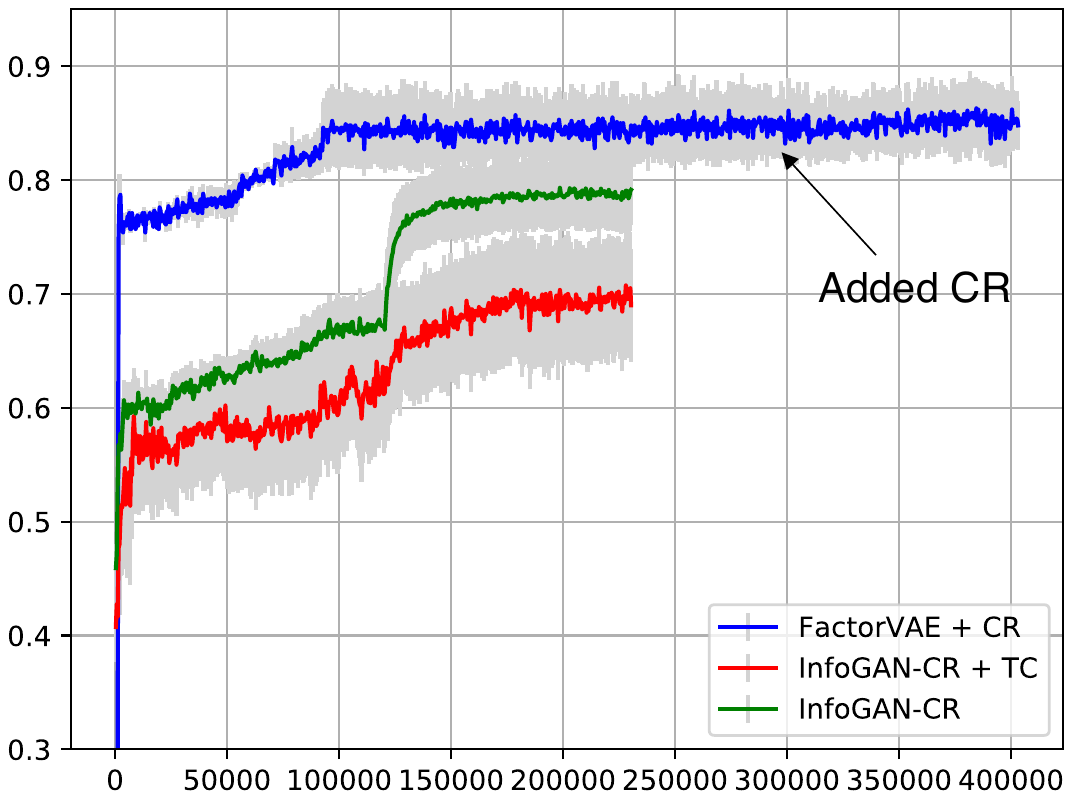}
		\put(-192,50){\rotatebox{90}{Disentanglement}}
		\put(-110,-7){Batch number}
		\hspace{0.6cm} 
	\includegraphics[width=3in]{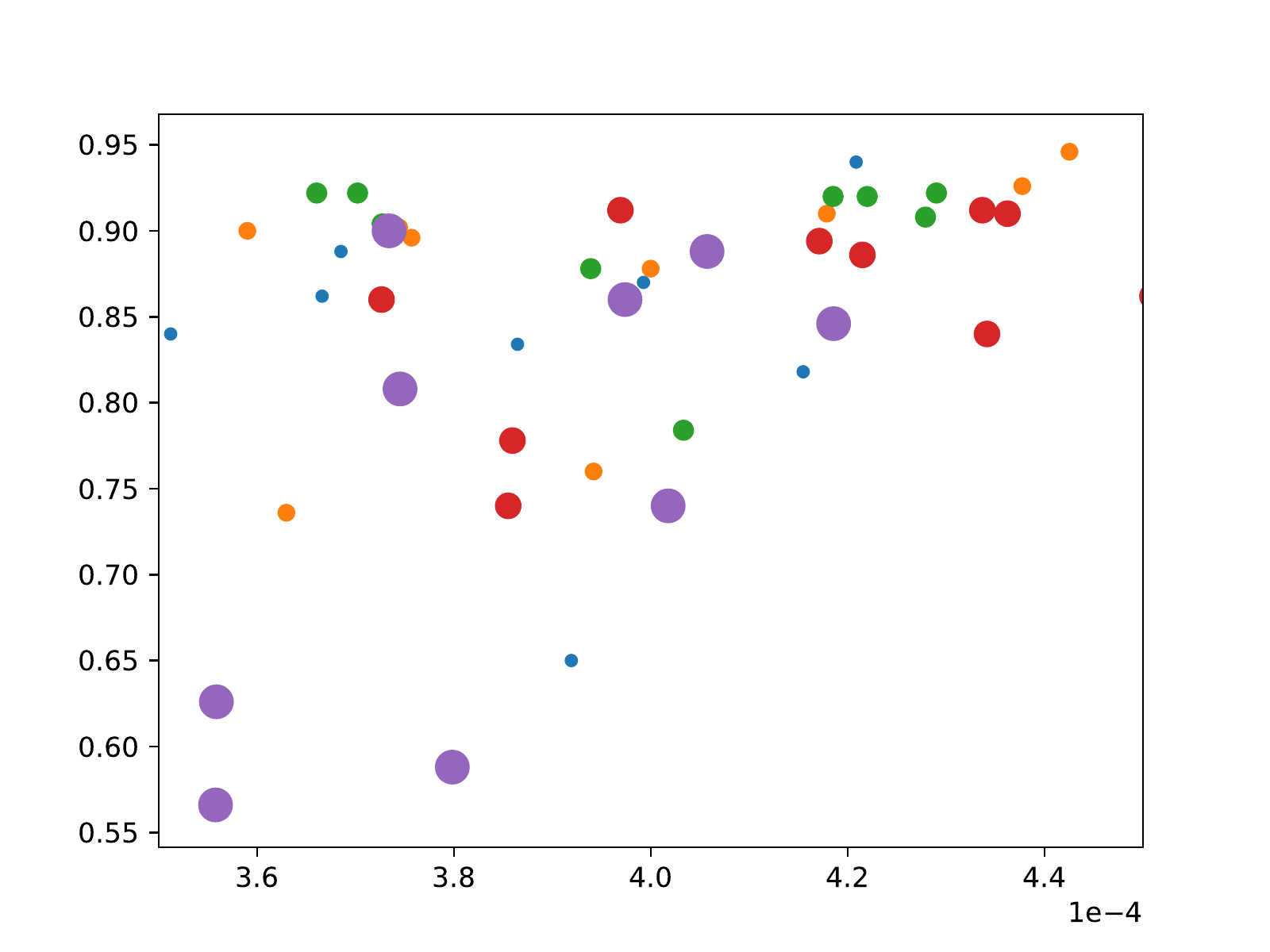}
	\put(-135,-5){dHSIC Score}
	\put(-220,50){\rotatebox{90}{Disentanglement}}
	\caption{[left panel] Adding TC regularization to InfoGAN-CR does not improve disentanglement; neither does adding CR to FactorVAE. [right panel] (dHSIC score, disentanglement) pairs for various \gan models. Each point represents a single model, and point size/color signifies the value of $\alpha \in \{0,1,2,4,8\}$. Larger points denote larger $\alpha$. 
	Disentanglement is measured as in \cite{KM18}. 
	We observe a positive correlation between disentanglement and dHSIC score, suggesting that TC regularization does not help GANs disentangle better. 
	}
	\label{fig:varying_alpha_dhsic}
\end{figure}

We claim that contrastive regularization is tailored to work well with GAN architectures. 
Similarly, we claim that total correlation regularization of FactorVAE is specifically tailored for VAE. 
To test this hypothesis, we have applied contrastive regularization (CR) to FactorVAE and total correlation (TC) regularization to InfoGAN-CR.
Figure \ref{fig:varying_alpha_dhsic} left panel  shows the disentanglement metric of each as a function of batch numbers. 
For FactorVAE, we introduce CR regularization of $\alpha = 20$ at batch 300,000. 
Note that the metric does not change perceptibly after adding CR. 
For InfoGAN-CR, we ran one set of trials with TC regularization from the beginning (red curve) and one set of trials without TC regularization (green curve). 
We use InfoGAN regularizer $\lambda = 0.7$ and TC coefficient $\beta = 1$ for the former.
Notice that \gan has a lower disentanglement score than FactorVAE in this plot because we did not use the optimal $\lambda$ for this dataset; 
this sensitivity is a weakness of \gan (as well as InfoGAN).
In Figure \ref{fig:varying_alpha_dhsic}, we observe that TC regularization actually reduces disentanglement compared to InfoGAN-CR without TC.
The jumps in disentanglement for the \gan curves are due to progressive training; we change the contrastive gap from 1.9 to 0.0 at batch 120,000.
The red line (\gan + TC) is averaged over 4 runs, the blue line (FactorVAE + CR) over 2 runs, and the green line (\gannosp) over ten.
These results support (but do not prove) our hypothesis that CR is better-suited to GAN architectures, whereas TC is better-suited to VAE architectures. 
To further confirm  this intuition, we show that disentanglement appears negatively correlated with a measure of total correlation in the following. 

To explore the relation between total correlation and disentanglement, Figure \ref{fig:varying_alpha_dhsic} plots the disentanglement score of \cite{KM18} as a function of dHSIC score while varying $\alpha$ for \gannosp.
Each point represents a single model, and point size/color signifies the value of $\alpha \in \{0,1,2,4,8\}$. Larger points denote larger $\alpha$. 
Since dHSIC approximates the total correlation between the latent codes, a lower dHSIC score implies a lower total correlation. 
Perhaps surprisingly, we find a noticeable positive correlation between dHSIC score and disentanglement. 
This suggests that TC regularization (i.e., encouraging small TC) actually hurts disentanglement for \gannosp. 
This may help to explain, or at least confirm, the findings in Figure \ref{fig:varying_alpha_dhsic}, which show that adding TC regularization to \gan reduces the disentanglement score.

\subsection{Contrastive Regularizer without InfoGAN Regularizer}
\label{app:onlycr}

\begin{figure}[h]
	\begin{center}
	\includegraphics[width=.3\textwidth]{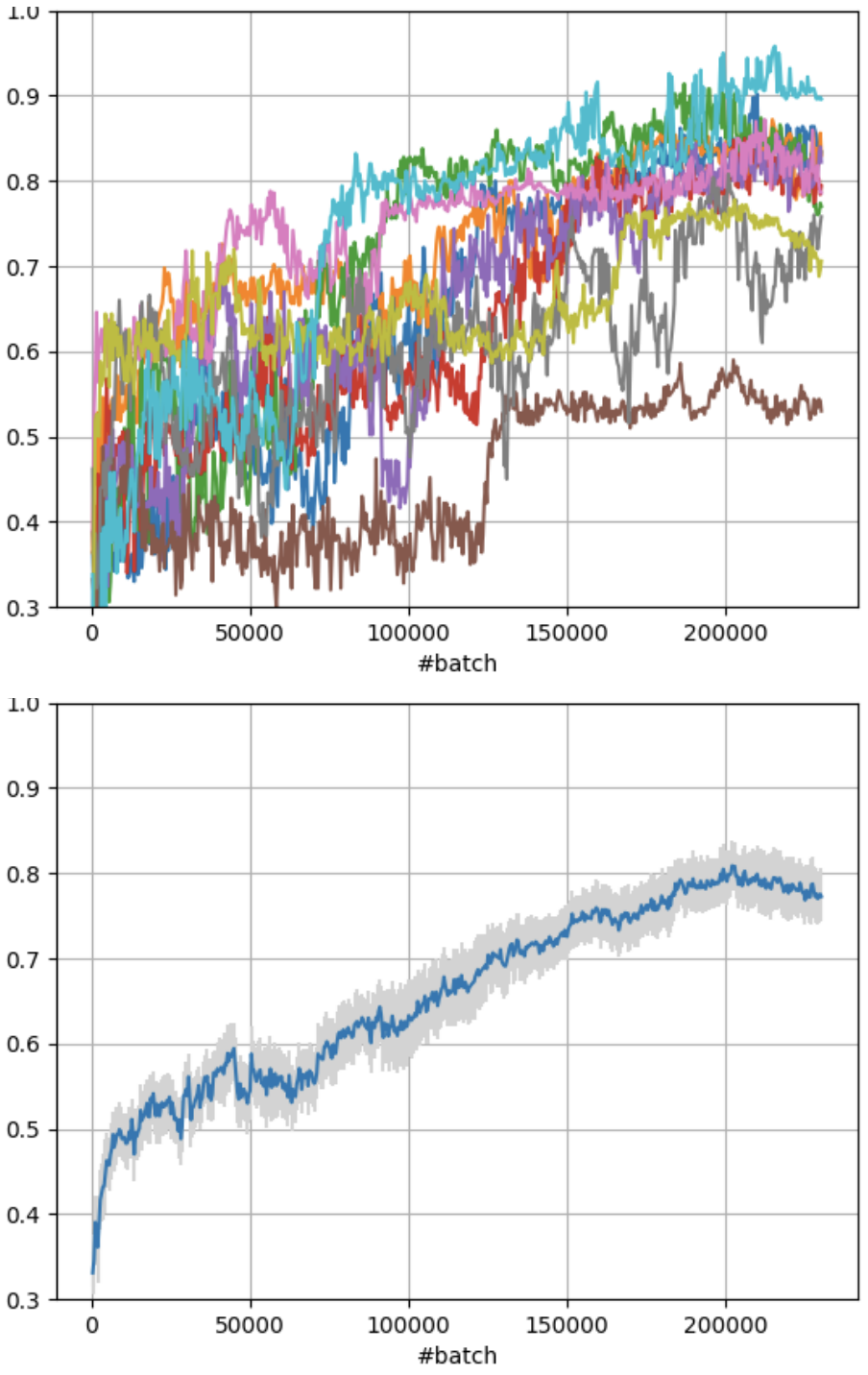}
	\includegraphics[width=.3\textwidth]{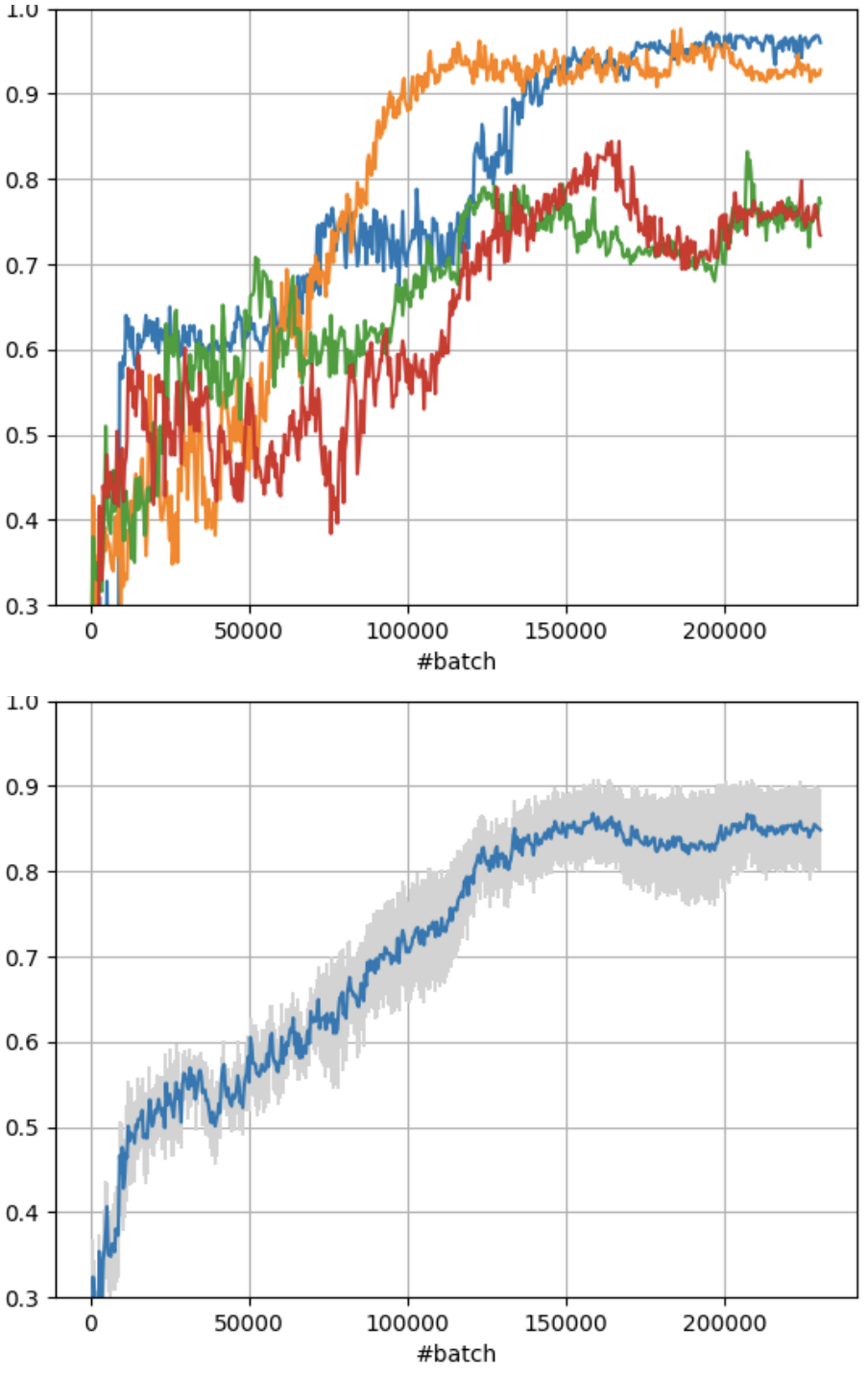}
	\includegraphics[width=.3\textwidth]{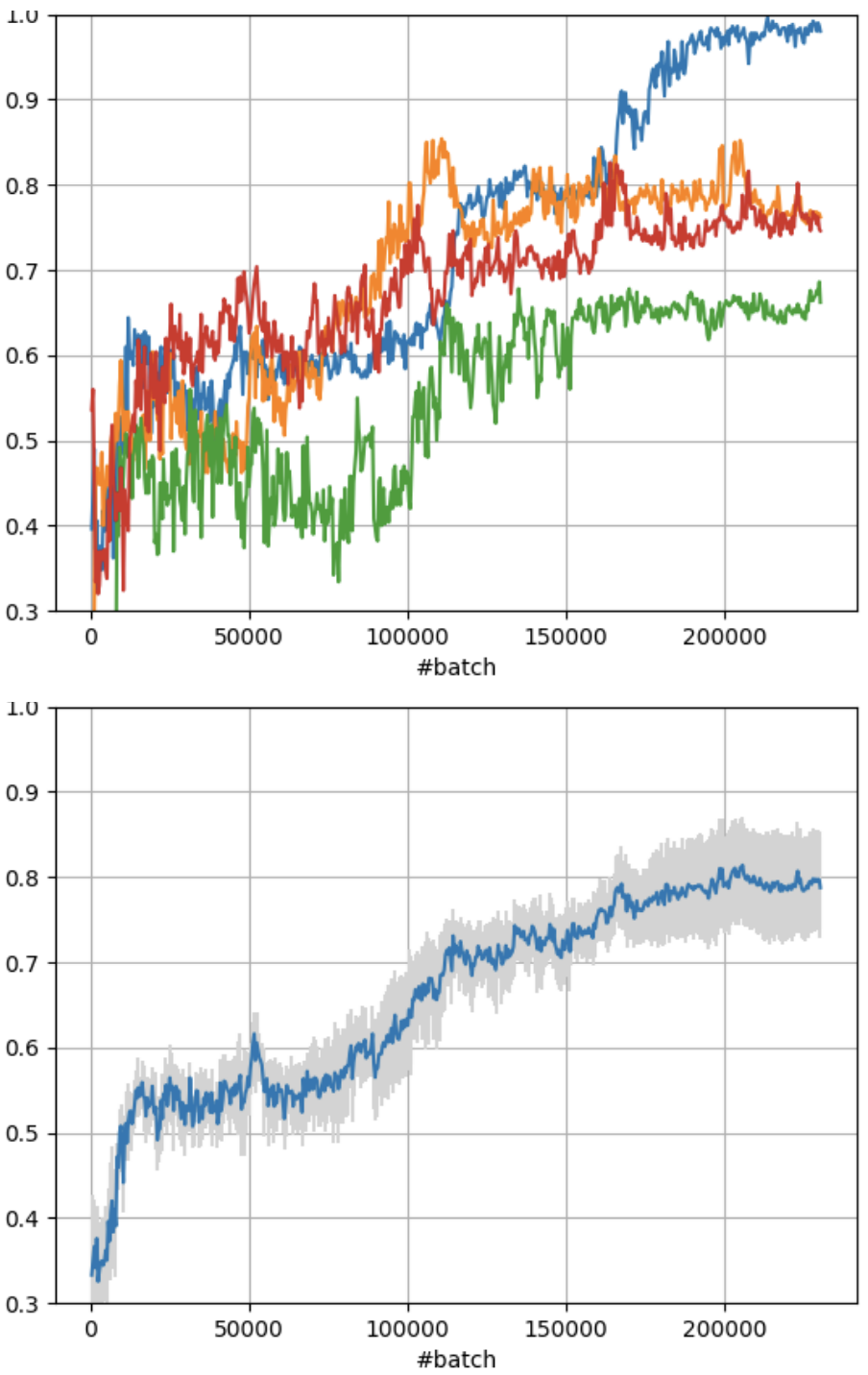}
	\put(-456,25){\rotatebox{90}{Disentanglement}}
	\put(-456,145){\rotatebox{90}{Disentanglement}}
	\put(-390,250){$\cL_{\rm Adv} - \alpha \cL_{\rm c}$}
	\put(-230,250){$\cL_{\rm Adv} - \alpha \cL_{\rm c}$}
	\put(-230,235){with PacGAN2}
	\put(-90,250){$\cL_{\rm Adv} - \alpha \cL_{\rm c}$}
	\put(-90,235){with PacGAN3}
	\end{center}
	\caption{Training without InfoGAN loss suffers from severe mode collapse, which in turn results in poor disentanglement score (left). 
	With PacGAN2 discriminator that takes 2 packed samples together each time, the mode collapse significantly decreases and disentanglement also improves (middle). 
	With PacGAN3 discriminator that takes 3 packed samples together each time, 
	the higher dimensionality of those packed samples results in poor performance (right). 
	The training trajectory of each instance is shown on the top, and the average is shown on the bottom. 
	}
	\label{fig:onlycr}
\end{figure}
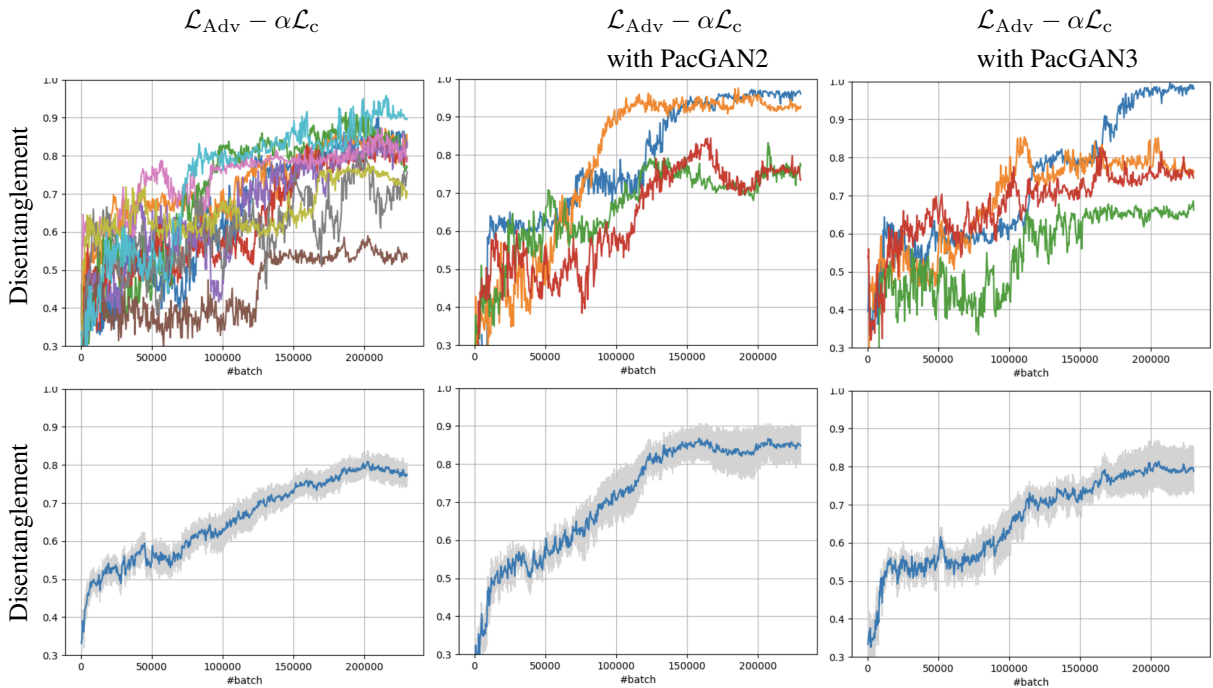

To test if InfoGAN regularizer is necessary, 
we trained \dsprites{} dataset without InfoGAN regularizer (i.e.~ $\lambda=0$) and progressively increasing $\alpha$. 
This new loss suffers from significant mode collapse, 
which can be significantly reduced with a recent technique for mitigating mode collapse known as PacGAN introduced in \cite{LKFO17}. 
The main idea is to life the adversarial discriminator to take $m$ samples packed together, all from either real data or generated data. 
This provably introduces an implicit inductive bias towards penalizing mode collapse, which is mathematically defined in 
\cite{LKFO17}, in terms of binary hypothesis testing and type I and type II errors. 
The resulting metric are shown in Figure~\ref{fig:onlycr}, where even with PacGAN we do not get the desired level of disentanglement 
without InfoGAN regularizer. We believe that InfoGAN and Contrastive regularizers play complementary roles in disentangling GANs. 

\subsection{Implementation Details}
\label{app:implementation}
In \dsprites{} experiments, we used a convolutional neural network for the FactorVAE encoder, InfoGAN discriminator, and CR discriminator, and a deconvolutional neural network for the decoder, and a multi-layer perceptron for total correlation discriminators. 
We used the Adam optimizer for all updates, whose learning rates are described below. 
For unmodified InfoGAN, we used the architecture described in Table 5 of \cite{KM18}. 
This architecture is reproduced in Table \ref{tab:infogan_arch} for completeness. 
As mentioned in Section \ref{sec:info}, we make several changes to the training of InfoGAN to improve its disentanglement, including changing the Adam learning rate to 0.001 for the generator and 0.002 for the InfoGAN and CR discriminators ($\beta_1$ is still 0.5).  
The architectural changes are included in Table \ref{tab:infogan_mod_arch}. 
We include in Table \ref{tab:cr_disc_dsprites} the architecture of our CR discriminator, which is similar to the InfoGAN discriminator.
Finally, Table \ref{tab:factorvae_arch} contains the architecture of FactorVAE, reproduced from \cite{KM18}.  
We use a batch size of 64 for all experiments.

\begin{table*}[h]
\centering
\begin{tabular}{|l|l|}
\hline
\textbf{Discriminator $D$ / Encoder $Q$}                        & \textbf{Generator $G$}                             \\ \hline
Input 64 $\times$ 64 binary image                              & Input $\in \mathbb R^{10}$                           \\ \hline
$4 \times 4$ conv. 32 lReLU. stride 2                          & FC. 128 ReLU. batchnorm                            \\ \hline
$4 \times 4$ conv. 32 lReLU. stride 2. batchnorm               & FC. $4 \times 4 \times 64$ ReLU. batchnorm         \\ \hline
$4 \times 4$ conv. 64 lReLU. stride 2. batchnorm               & $4 \times 4$ upconv. 64 lReLU. stride 2. batchnorm \\ \hline
$4 \times 4$ conv. 64 lReLU. stride 2. batchnorm               & $4 \times 4$ upconv. 32 lReLU. stride 2. batchnorm \\ \hline
FC. 128 lReLU. batchnorm (*)                                      & $4 \times 4$ upconv. 32 lReLU. stride 2. batchnorm \\ \hline
From *: FC. 1 sigmoid. (output layer for D)                                      & $4 \times 4$ upconv. 1 sigmoid. stride 2           \\ \hline
From *: FC. 128 lReLU. batchnorm. FC 5 for $Q$ &                                                    \\ \hline
\end{tabular}
\caption{InfoGAN architecture for \dsprites{} experiments from \cite{KM18}. We used 5 continuous codes and 5 noise variables.}
\label{tab:infogan_arch}
\end{table*}

\begin{table*}[h]
\centering
\begin{tabular}{|l|l|}
\hline
\textbf{Discriminator $D$ / Encoder $Q$}                        & \textbf{Generator $G$}                             \\ \hline
Input 64 $\times$ 64 binary (\dsprites{}) or color (\teapots{}) image                              & Input $\in \mathbb R^{10}$                           \\ \hline
$4 \times 4$ conv. 32 lReLU. stride 2. spectral normalization                          & FC. 128 ReLU. batchnorm                            \\ \hline
$4 \times 4$ conv. 32 lReLU. stride 2. spectral normalization                & FC. $4 \times 4 \times 64$ ReLU. batchnorm         \\ \hline
$4 \times 4$ conv. 64 lReLU. stride 2. spectral normalization                & $4 \times 4$ upconv. 64 lReLU. stride 2. batchnorm \\ \hline
$4 \times 4$ conv. 64 lReLU. stride 2. spectral normalization                & $4 \times 4$ upconv. 32 lReLU. stride 2. batchnorm \\ \hline
FC. 128 lReLU. spectral normalization (*)                                       & $4 \times 4$ upconv. 32 lReLU. stride 2. batchnorm \\ \hline
From *: FC. 1 sigmoid. (output layer for D)                                      & $4 \times 4$ upconv. 1 (\dsprites{}) or 3 (\teapots{}) sigmoid. stride 2           \\ \hline
From *: FC. 128 lReLU. spectral normalization  &                                                    \\ \hline
\hspace{3.3em} FC 5. spectral normalization (output layer for $Q$) & \\ \hline
\end{tabular}
\caption{InfoGAN (modified) architecture for \dsprites{} and \teapots{} experiments. We used 5 continuous codes and 5 noise variables.}
\label{tab:infogan_mod_arch}
\end{table*}

\begin{table*}[h]
	\centering
	\begin{tabular}{|l|}
		\hline
		\textbf{CR Discriminator }       \\ \hline
		Input 64 $\times$ 64 $\times$ 2 (2 binary images)             \\ \hline
		$4 \times 4$ conv. 32 lReLU. stride 2. spectral normalization   \\ \hline
		$4 \times 4$ conv. 32 lReLU. stride 2. spectral normalization              \\ \hline
		$4 \times 4$ conv. 64 lReLU. stride 2. spectral normalization      \\ \hline
		$4 \times 4$ conv. 64 lReLU. stride 2. spectral normalization       \\ \hline
		FC. 128 lReLU. spectral normalization     \\ \hline
		FC 5. softmax \\ \hline
	\end{tabular}
	\caption{CR discriminator architecture for \dsprites{} experiments. }
	\label{tab:cr_disc_dsprites}
\end{table*}

\begin{table*}[h]
	\centering
	\begin{tabular}{|l|}
		\hline
		\textbf{CR Discriminator}       \\ \hline
		Input 64 $\times$ 64 $\times$ 6 (2 color images)             \\ \hline
		$4 \times 4$ conv. 32 lReLU. stride 2.                             \\ \hline
		$4 \times 4$ conv. 32 lReLU. stride 2. batchnorm             \\ \hline
		$4 \times 4$ conv. 64 lReLU. stride 2. batchnorm     \\ \hline
		$4 \times 4$ conv. 64 lReLU. stride 2. batchnorm      \\ \hline
		FC. 128 lReLU. batchnorm            \\ \hline
		FC 5. softmax \\ \hline
	\end{tabular}
	\caption{CR discriminator architecture for \teapots{} experiments. }
	\label{tab:cr_teapots_arch}
\end{table*}

\begin{table*}[h]
\centering
\begin{tabular}{|l|l|}
\hline
\textbf{Encoder}                       & \textbf{Decoder}                        \\ \hline
Input 64 $\times$ 64 binary image      & Input $\in \mathbb R^{10}$              \\ \hline
$4 \times 4$ conv. 32 ReLU. stride 2   & FC. 128 ReLU.                           \\ \hline
$4 \times 4$ conv. 32 ReLU. stride 2. & FC. $4 \times 4 \times 64$ ReLU.        \\ \hline
$4 \times 4$ conv. 64 ReLU. stride 2. & $4 \times 4$ upconv. 64 ReLU. stride 2. \\ \hline
$4 \times 4$ conv. 64 ReLU. stride 2. & $4 \times 4$ upconv. 32 ReLU. stride 2. \\ \hline
FC. 128.            & $4 \times 4$ upconv. 32 ReLU. stride 2. \\ \hline
FC. $2 \times 10$.                                        & $4 \times 4$ upconv. 1. stride 2        \\ \hline
\end{tabular}
\caption{FactorVAE architecture for \dsprites{} and \teapots{} experiments, taken from \cite{KM18}.}
\label{tab:factorvae_arch}
\end{table*}

We implemented both FactorVAE and InfoGAN  using the architectures described in \cite{KM18}. 
Although InfoGAN exhibits a reported disentanglement score of $0.59 \pm 0.70$ in \cite{KM18},
we find that InfoGAN can exhibit substantially higher disentanglement scores ($0.83 \pm 0.03$) through some basic changes to the architecture and loss function. 
In particular, in accordance with \cite{MKKY18}, we changed the loss function from Wasserstein GAN to the traditional JSD loss. 
We also changed the generator's Adam learning rate to $0.001$ and the InfoGAN and CR discriminators learning rates to $0.002$;
we used 5 continuous input codes, whereas \cite{KM18} reported using four continuous codes and one discrete one.
We also used batch normalization in the generator, and spectral normalization in the discriminator. 
The effects of these changes are shown in the line `InfoGAN (modified)' in Table \ref{tbl:comp}.
The progressive scheduling of InfoGAN-CR is to run InfoGAN ($\alpha=0,\lambda = 0.05$) for 288000 batches, and then continue running it with CR ($\alpha=2.0,\lambda = 0.05$, gap=0) for additional 34560 batches (so that the total number of epoches is 28).
For FactorVAE, we used the architecture of \cite{KM18}, which uses $k=10$ latent codes in their architecture.

\section{\teapots{} Dataset}
\label{app:teapots}

We ran \gan on the \teapots{} dataset from  \cite{EW18},
with images of \teapots{} in various orientations and colors generated by the renderer in \cite{MW16}. 
Images have five latent factors: color (red, blue, and green), rotation (vertical), and rotation (lateral).
Colors are randomly drawn from $[0,1]$. Rotation (vertical) is randomly drawn from $[0,\pi/2]$. Rotation (lateral) is  drawn from $[0, 2\pi]$.
We generated a dataset of 200,000 such images with each combination of latent factors represented. 
Table \ref{tbl:teapot} shows the disentanglement scores of FactorVAE and InfoGAN compared to \gannosp.
Since the \teapots{} dataset does not have classes, we do not compute inception score for this dataset; 
however, the images generated by \gan appear sharper than those generated by FactorVAE.

We now discuss implementation details and additional experiments on  the  \teapots{} dataset. 
We used an identical architecture to the \dsprites{} dataset for InfoGAN and FactorVAE. 
For \gannosp, the CR discriminator architecture is changed slightly and is listed in Table \ref{tab:cr_teapots_arch}.
As with \dsprites{}, FactorVAE used 10 latent codes, so we  chose the best five to compute the disentanglement metric.

For InfoGAN-CR, we train InfoGAN with $\lambda=0.2$ for 50000 batches, and then
InfoGAN-CR with $\lambda=0.2$, $\alpha=3.0$, gap=1.9 for 35000 batches, and then
InfoGAN-CR with $\lambda=0.2$, $\alpha=3.0$, gap=0.0 for 40000 batches. We use a batch size of 64 for all experiments. %

\begin{figure}[H]
	\includegraphics[width=\textwidth]{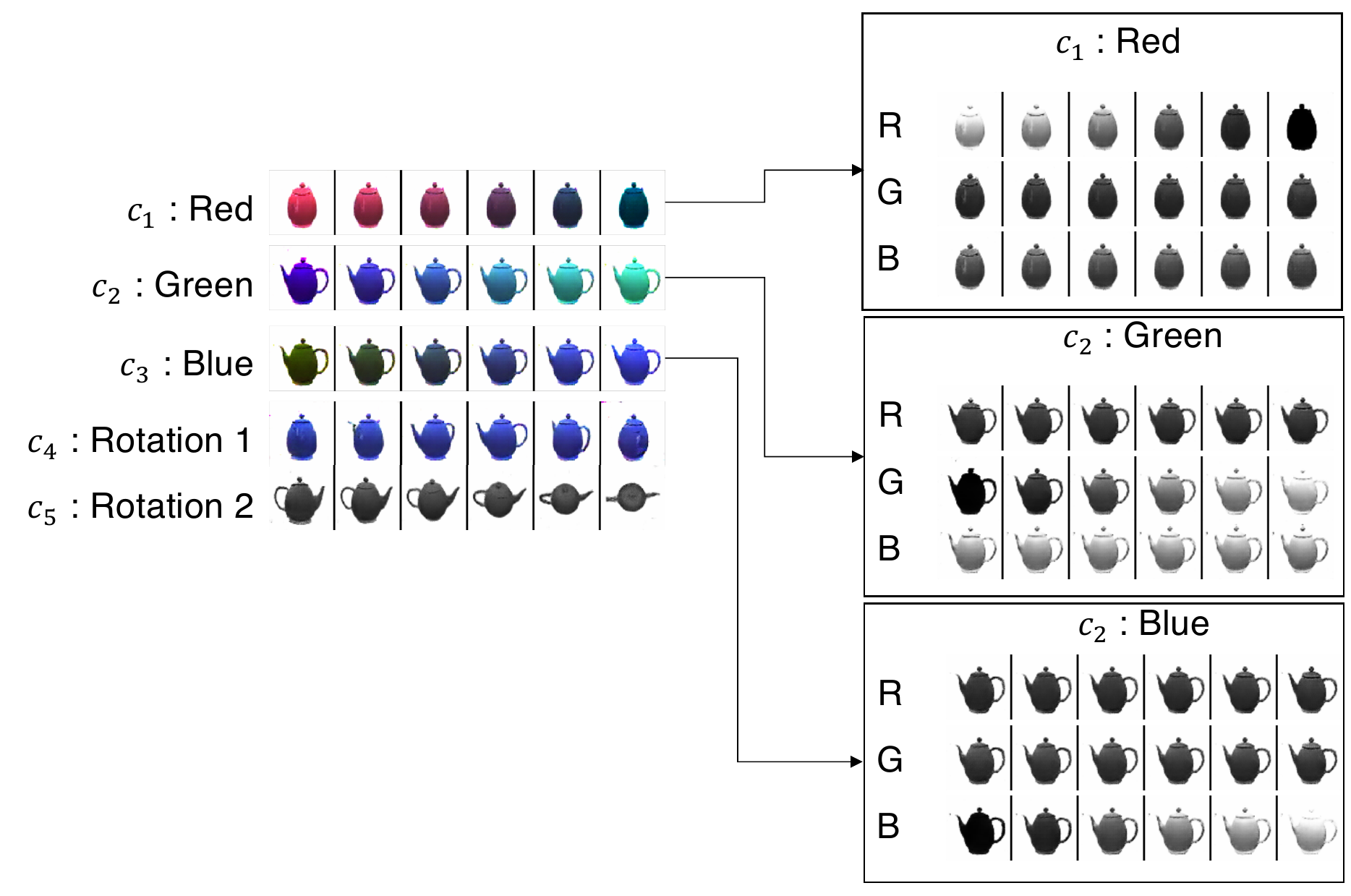}
	\caption{Latent traversal for \teapots{} dataset with \gannosp. Red (R), blue (B), and green (G) channels are shown separately for the latent factors that correspond to color.}
	\label{fig:teapot_traversal_infogan}
\end{figure}

\begin{figure}[H]
	\includegraphics[width=\textwidth]{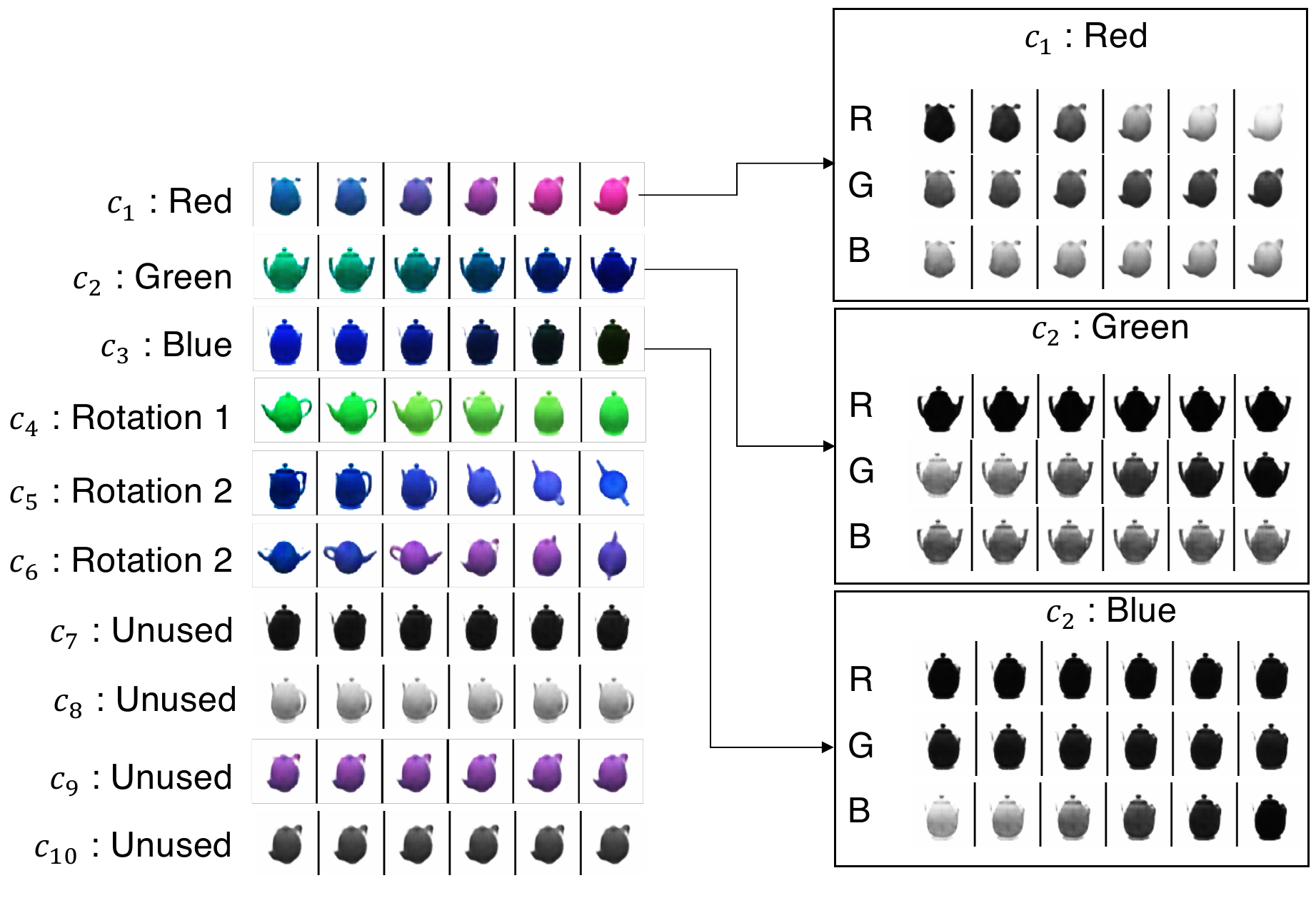}
	\caption{Latent traversal for \teapots{} dataset with FactorVAE. Red (R), blue (B), and green (G) channels are shown separately for the latent factors that correspond to color.}
	\label{fig:teapot_traversal_factorvae}
\end{figure}

We illustrate a latent traversal for the \teapots{} dataset under \gan in Figure \ref{fig:teapot_traversal_infogan}, which is from the best run of InfoGAN-CR with disentanglement metric of \cite{KM18} 1.0.
To make the traversal easier to interpret, we have separated the color channels for each  latent factor that captures color. 
This shows that each latent factor controls a single color channel, while the others are held fixed. 
A similar traversal is shown in Figure \ref{fig:teapot_traversal_factorvae} for FactorVAE, which is from the best run of FactorVAE with $\beta=40$ and disentanglement metric of \cite{KM18} 0.94.
Although it is difficult to draw conclusions from qualitative comparison, we found that the sharpness of images was reduced in the FactorVAE images, though FactorVAE is able to learn a meaningful disentanglement with three color factors and three (one duplicate) rotation factors. 

\begin{figure}[H]
    \centering
	\includegraphics[width=3.3in]{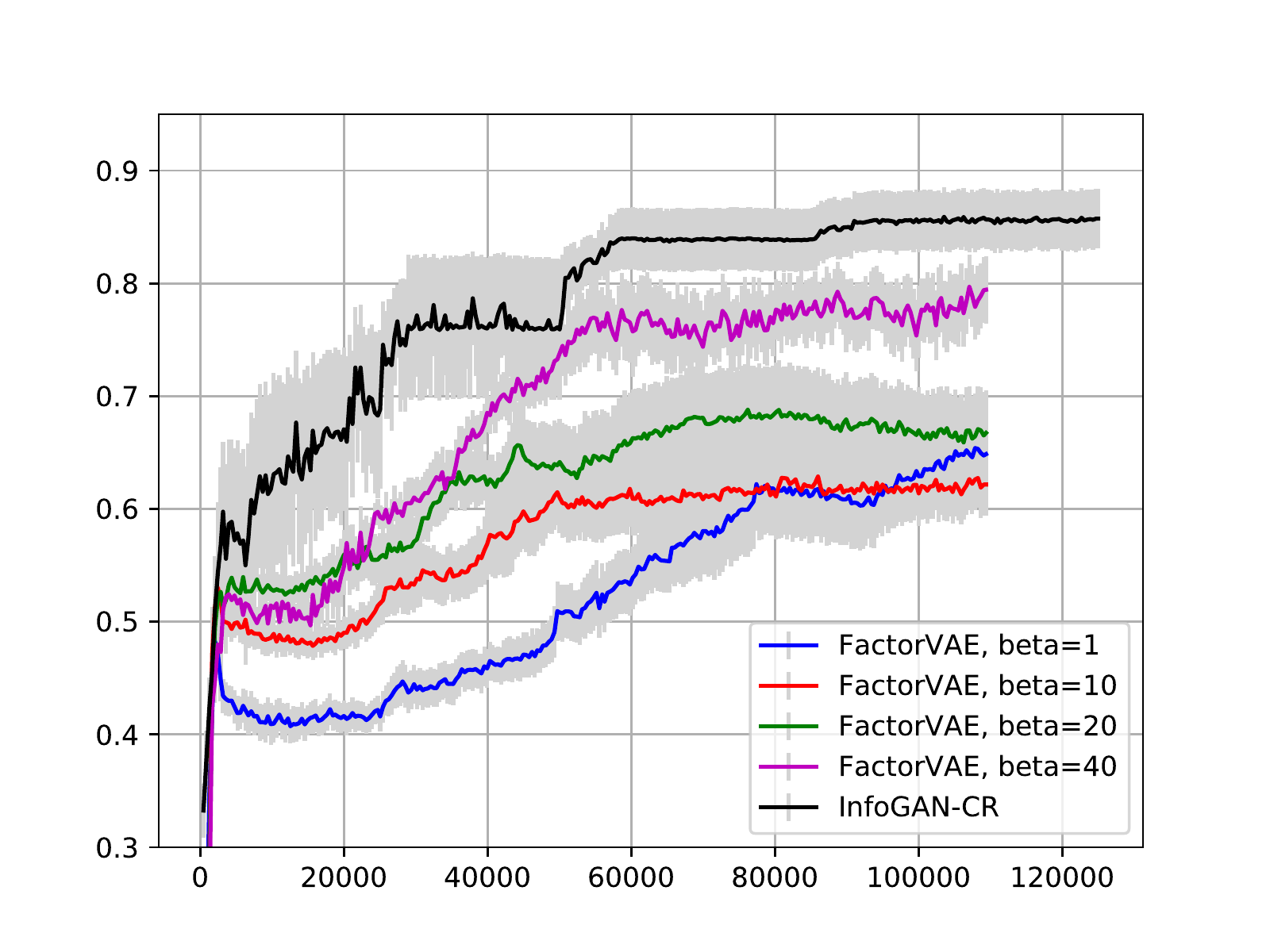}
	\put(-145,-5){Batch number}
	\put(-230,50){\rotatebox{90}{Disentanglement}}
	\caption{Disentanglement metric of \cite{KM18} as a function of batch number for \gan and FactorVAE on the \teapots{} dataset. The final models from this training were used to generate Table \ref{tbl:teapot}.}
	\label{fig:teapot_dis}
\end{figure}

Building on Table \ref{tbl:teapot}, we also plot the disentanglement metric of \cite{KM18} during the training of \gan and FactorVAE in Figure \ref{fig:teapot_dis}.
This plot shows that \gan achieves a consistently higher disentanglement score than FactorVAE throughout the training procedure, though FactorVAE comes close when $\beta=40$. 

\subsection{Non-commutative and Ambiguous Coordinate Systems}
\label{app:rotation}
To further push the boundaries of disentangled generation, it is important to understand when and why \gan fails;  the \teapots{} dataset gives a useful starting point. 
In particular,  \cite{higgins2018towards} observed that rotations about the canonical basis axes are  not commutative.
For example, suppose we apply two rotations of some angles, 
one about the $x$  and one about the $y$ axes in two different orders;
in general,  the resulting orientations of the object need not  be the same.

Because of this, a disentangled GAN or VAE trained on  a dataset where every possible orientation of teapot is represented should \emph{not} recover the canonical  basis for 3D space. 
Despite this fact, existing experiments (including our own) appear to recover the canonical axes of rotation.  
To explain this phenomenon, we observe that existing experiments on this dataset,  including \cite{EW18} and our own  initial experiments, do  not include all orientations of the object in the training data. 
For example, notice that none of the visualized images in Figure  \ref{fig:teapot_traversal_infogan}  show the bottom of the teapot.  
Because of the way the training data is selected, the canonical axes are indeed recovered. 

\begin{figure}[H]
\centering
	\includegraphics[width=2in]{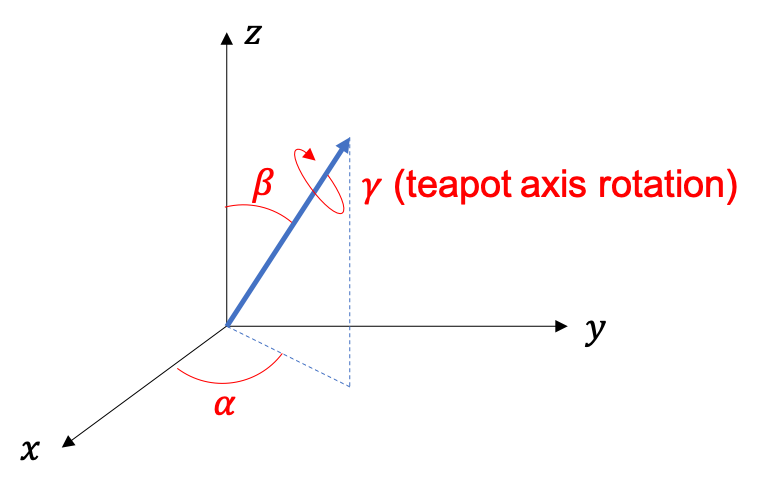}
	\caption{A commutative 3D rotational  coordinate system.}
	\label{fig:teapot_rotation}
\end{figure}

We next ask what would happen if all orientations were present in the dataset. 
To answer this  question,  first note that in general, 3D rotations can indeed commute. 
For example, the rotational coordinate system described in  Figure \ref{fig:teapot_rotation}  commutes. 
In that coordinate system, $\gamma$ represents the rotation along the axis of the teapot (orthogonal to the bottom of  the teapot), while $\alpha$ and $\beta$ represent  the orientation of the teapot's principal axis. 
However, this coordinate system is not unique; we could choose a different $z$-axis, for instance, and construct a different commutative rotational coordinate  system. 
Hence,  even if we were to represent all orientations of the teapots in our training data, it is unclear what orientation we would recover. 

\begin{figure}[H]
\centering
	\includegraphics[width=\textwidth]{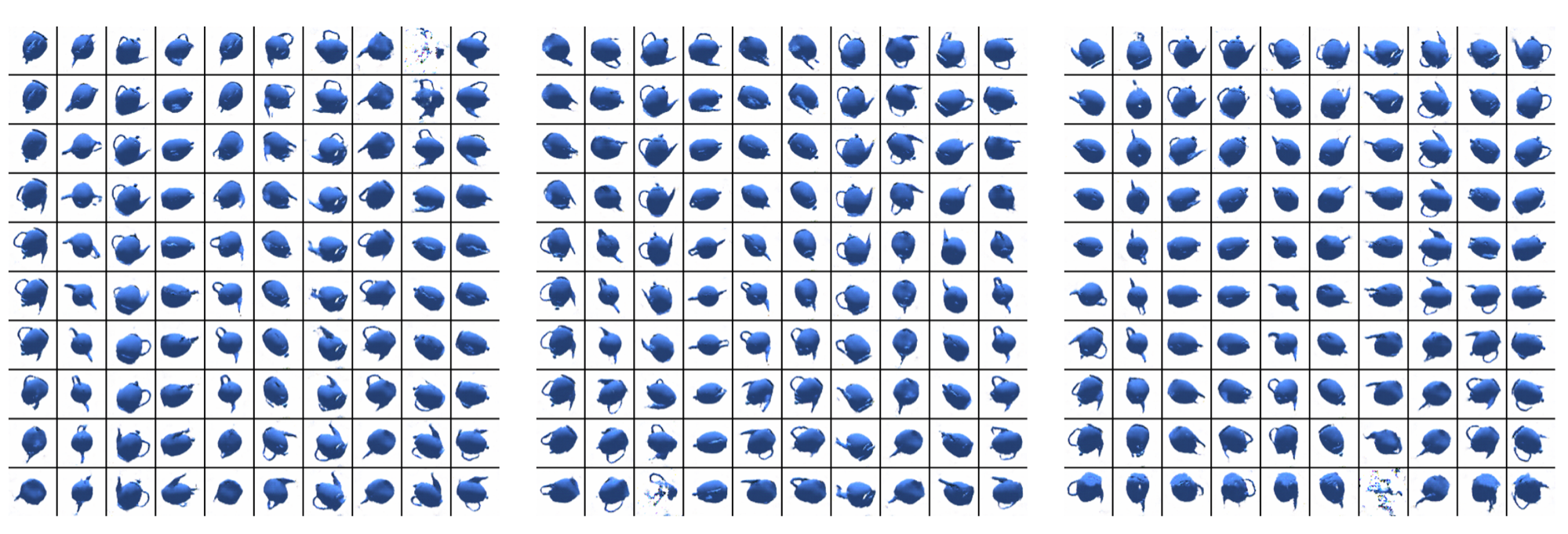}
	\put(-340,-10){$c_1$}
	\put(-200,-10){$c_2$}
	\put(-65,-10){$c_3$}
	\caption{Latent traversal for \gan over the \teapots{} dataset when every possible orientation of the teapot is included. 
	We find that \gan does appear to learn a consistent coordinate axis, even though there is not a unique disentangled representation.
	}
	\label{fig:teapot_traversal_overhead_light}
\end{figure}

Figure \ref{fig:teapot_traversal_overhead_light} shows the result from a single trial of an experiment where every possible orientation of the teapot is included in  the training data. 
We trained \gan with InfoGAN coefficient $\lambda=0.2$ and CR coefficient $\alpha=1.0$. 
As with our other experiments, we trained five latent factors and visualize the three most meaningful ones  in  Figure \ref{fig:teapot_traversal_overhead_light}.
To our surprise,  we find that the  system (roughly) recovers a similar coordinate system as the one depicted in  Figure \ref{fig:teapot_rotation}.
Here $c_1$ appears to capture $\gamma$, $c_2$ appears to recover $\beta$,  and $c_3$ recovers $\alpha$.
Even upon running multiple trials of this experiment, the \gan appears to learn (approximately) the same coordinate system  from  Figure \ref{fig:teapot_rotation}.
We hypothesize that this happens because of the illumination in the images. 
By default, the renderer from \cite{MW16} renders the teapots with an overhead light source. 
This may distinguish the vertical  axis from the others, causing \gan  to learn the vertical axis as a reference for the rotational coordinate system. 

\begin{figure}[H]
\centering
	\includegraphics[width=\textwidth]{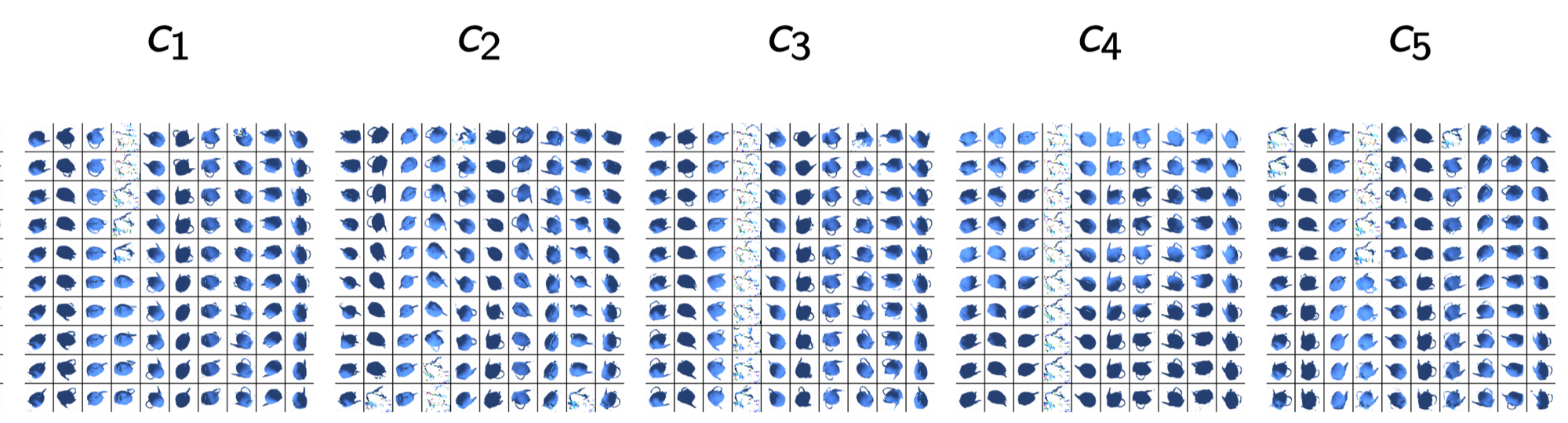}
	\caption{Latent traversal  over the \teapots{} dataset when the light source is chosen uniformly for each training image. 
	We find that \gan does not appear to learn a consistent coordinate axis.
	}
	\label{fig:teapot_traversal_random_light}
\end{figure}

To test this hypothesis, Figure \ref{fig:teapot_traversal_random_light} shows an experiment in which we randomized the light source for each image. 
For this experiment, we again  used an InfoGAN coefficient of $\lambda=0.2$, and a disentangling  coefficient of $\alpha=1.0$; the experimental  setup is identical  to that of Figure \ref{fig:teapot_traversal_overhead_light}, except for the light source in the training data.
We  find that in this case, \gan no longer appears to learn the vertical coordinate system.
Indeed, it  does not seem  to  learn \emph{any} disentangled representation. %
We observed similar results for FactorVAE for a range of total correlation coefficients on this dataset, so we do not believe  this effect is unique to \gannosp.
Instead, it suggests that in settings where there is no single disentangled representation, current disentanglement methods fail.

\section{\cdspriteslong{} (\cdsprites{}) Dataset} \label{sec:cdsprites}
To further study the failure modes of our InfoGAN-CR and other state-of-the-art architectures for disentangling, we 
introduce the following experiments. 
Towards this purpose we generate a new synthetic dataset which we call \cdspriteslong{} (\cdsprites{}).

\begin{figure}[H]
	\hspace{1.5cm}
	\includegraphics[width=0.5in]{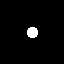}
	\hspace{0.95cm}
	\includegraphics[width=0.5in]{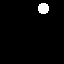}
	\hspace{0.95cm}
	\includegraphics[width=0.5in]{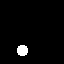}
	\hspace{3.5cm}
	\includegraphics[width=0.5in]{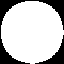}
	\put(-320,-10){\small $r=0$, $\gamma=0$}
	\put(-255,-10){\small $r=26$, $\gamma=33$}
	\put(-185,-10){\small $r=21$, $\gamma=13$}
	\put(-270,-22){(a) Samples (radius, angle)}
	\put(-70,-22){(b) Overlap of all samples}
	\caption{\cdspriteslong{} dataset.}
	\label{fig:cdsprites_dataset}
\end{figure}

It contains a set of 1080, 64 $\times$ 64 8-bit gray-scale images generated as follows. Each image has a white (pixel value 255) circular (disc) shape of radius 5 pixels on a black background (pixel value 0). For the placement of the shape we construct a polar (2D) coordinate system, whose co-ordinates are radius $r \in [0, 32]$ and angle $\gamma \in [0, 2\pi)$, and its origin is the center of the image canvas: (32, 32). Then the circular shape is placed on the on a point $(r, \gamma)$, such that radius $r$ is uniformly selected from $\{0, 1, \ldots, 26\}$ (pixel unit) and angle $\gamma$ is uniformly selected from $\{0, 2\pi\frac{1}{40}, 2\pi\frac{1}{40}, \ldots, 2\pi\frac{39}{40} \}$. Thus if the center of the circular shape is selected as $(r, \gamma)$ then it will be place on the pixel $(32 + r \cos \gamma , 32 + r \sin \gamma)$. Thus there are 27 $\times$ 40 (1080) total images in the dataset. Fig.~\ref{fig:cdsprites_dataset}(a) shows some sample and their corresponding radius ($r$) and angle ($\gamma$) indices. Fig.~\ref{fig:cdsprites_dataset}(b) shows the overlap of all the images in the dataset which shows the circular region where the shape can be placed. We expect that a good disentangling representation should disentangle the radius and angle latent factors. 

We train FactorVAE, InfoGAN and InfoGAN-CAR models on this dataset. We use the same architecture as the one we use for the \dsprites{} dataset for these models. However, we reduced number of latent factors to 2 for all the models, since there are only two factors to be learned. 

\begin{figure}[H]
		\centering
		\includegraphics[width=4.5in]{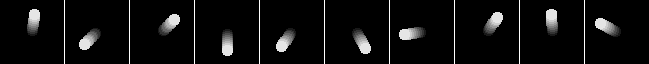}
	\put(-360,10){Radius}
	\put(-360,35){\bf Latent}
	\vspace{0.1cm}
	\includegraphics[width=4.5in]{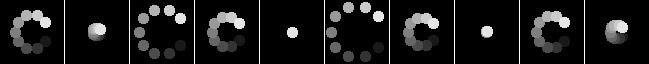}
	\put(-360,10){Angle}
	\caption{(\cdsprites{}) True latent traversal (see Appendix \ref{sec:cdsprites} for explanation). We see that the top row changes position radially and the bottom row changes the position angluarly.}
	\label{fig:cdsprites_traversal}
\end{figure}

\begin{figure}[H]
	\centering
	\includegraphics[width=4.5in]{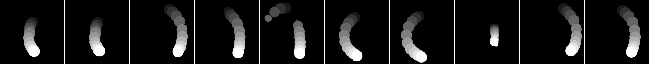}
	\put(-350,10){0}
	\put(-360,35){\bf Latent}
	\vspace{0.1cm}
	\includegraphics[width=4.5in]{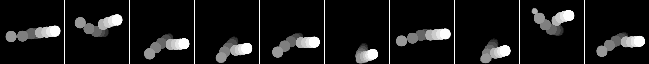}
	\put(-350,10){1}
	\caption{(\cdsprites{}) FactorVAE latent traversal (see Appendix \ref{sec:cdsprites} for explanation). We see that the model does not disentangle the true radius and angle factors and in fact they are mixed.}
	\label{fig:cdsprites_traversal_fvae}
\end{figure}

\begin{figure}[H]
	\centering
	\includegraphics[width=4.5in]{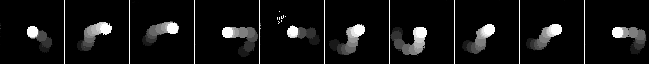}
	\put(-350,10){0}
	\put(-360,35){\bf Latent}
	\vspace{0.1cm}
	\includegraphics[width=4.5in]{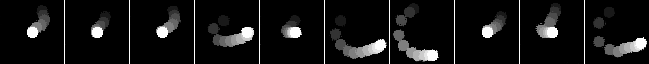}
	\put(-350,10){1}
	\caption{(\cdsprites{}) InfoGAN latent traversal (see Appendix \ref{sec:cdsprites} for explanation). We see that the model does not disentangle the true radius and angle factors and in fact they are mixed.}
	\label{fig:cdsprites_traversal_infogan}
\end{figure}

\begin{figure}[H]
	\centering
	\includegraphics[width=4.5in]{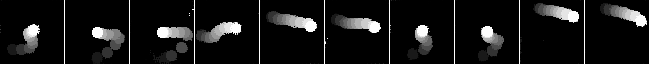}
	\put(-350,10){0}
	\put(-360,35){\bf Latent}
	\vspace{0.1cm}
	\includegraphics[width=4.5in]{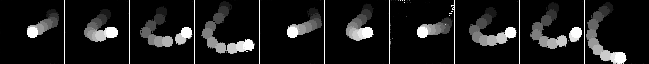}
	\put(-350,10){1}
	\caption{(\cdsprites{}) InfoGAN-CR latent traversal (see Appendix \ref{sec:cdsprites} for explanation). We see that the model does not disentangle the true radius and angle factors and in fact they are mixed.}
	\label{fig:cdsprites_traversal_infogan-cr}
\end{figure}

In Fig.~\ref{fig:cdsprites_traversal}, we show the traversal of the dataset through the true latents. Each row corresponds to one one latent's traversal while the other is fixed. Each column has a different fixed value for the other fixed latent. As we traverse through a latent keeping the other fixed, for easy visualization, we increase the shade of the shape from the darkest to the brightest. In Figs.~\ref{fig:cdsprites_traversal_fvae}, \ref{fig:cdsprites_traversal_infogan}, and \ref{fig:cdsprites_traversal_infogan-cr}, where show the traversal through the learned latents of FactorVAE, InfoGAN, and InfoGAN-CR respectively. We see that the none of the models truly disentangle the true radius and angle factors and in fact they are mixed in the learned learned latents. We believe this dataset is hard for any current models to disentangle, and thus could be used as good baseline for future research.

\section{CelebA Dataset}
\label{app:celeba}

\subsection{Implementation Details}
We crop the center 128$\times$128 pixels from original CelebA images, and resize the images to 32 $\times$ 32  for training.

The architecture of generator, InfoGAN discriminator, CR discriminator in this experiment are shown in Table \ref{tab:infogan_celeba_arch} and Table \ref{tab:cr_celeba_arch}, which are in part motivated by an independent implementation of InfoGAN\footnote{https://github.com/conan7882/tf-gans}. We used the Adam optimizer for all updates. Generator's learning rate  is 2e-3, InfoGAN discriminator's and CR discriminator's learning rates are 2e-4. $\beta_1$ is 0.5, batch size is 128 for all components. We train InfoGAN with $\lambda=2.0$ for 80000 batches, and then train InfoGAN-CR with $\lambda=2.0$, $\alpha=1.0$, $\text{gap}=0.0$ for another 10173 batches (so that total number of epoch is 57). The samples are generated from the end of training of one run.
\begin{table*}[h]
	\centering
	\begin{tabular}{|l|l|}
		\hline
		\textbf{Discriminator $D$ / Encoder $Q$}                        & \textbf{Generator $G$}                             \\ \hline
		Input 32 $\times$ 32 color image                             & Input $\in \mathbb R^{105}$                           \\ \hline
		$5 \times 5$ conv. 64 lReLU. stride 2. spectral normalization                          & FC. 1024 ReLU. batchnorm                            \\ \hline
		$5 \times 5$ conv. 128 lReLU. stride 2. spectral normalization                & FC. $4 \times 4 \times 256$ ReLU. batchnorm         \\ \hline
		$5 \times 5$ conv. 256 lReLU. stride 2. spectral normalization                & $5 \times 5$ upconv. 128 ReLU. stride 2. batchnorm \\ \hline
		$5 \times 5$ conv. 512 lReLU. stride 2. spectral normalization (*)                & $5 \times 5$ upconv. 64 ReLU. stride 2. batchnorm \\ \hline
		From *: FC. 1 sigmoid. (output layer for D)                                      & $5 \times 5$ upconv. 3 tanh. stride 2           \\ \hline
		From *: FC. 5. spectral normalization (output layer for $Q$) & \\ \hline
	\end{tabular}
	\caption{InfoGAN architecture for CelebA experiments. We used 5 continuous codes and 100 noise variables.}
	\label{tab:infogan_celeba_arch}
\end{table*}

\begin{table*}[h]
	\centering
	\begin{tabular}{|l|}
		\hline
		\textbf{CR Discriminator}       \\ \hline
		Input 32 $\times$ 32 $\times$ 6 (2 color images)             \\ \hline
		$5 \times 5$ conv. 64 lReLU. stride 2.                             \\ \hline
		$5 \times 5$ conv. 128 lReLU. stride 2. batchnorm             \\ \hline
		$5 \times 5$ conv. 256 lReLU. stride 2. batchnorm     \\ \hline
		$5 \times 5$ conv. 512 lReLU. stride 2. batchnorm      \\ \hline
		FC 5. softmax \\ \hline
	\end{tabular}
	\caption{CR discriminator architecture for CelebA experiments. }
	\label{tab:cr_celeba_arch}
\end{table*}

\section{Exploring Choices of CR}
In the design of CR, there are 4 choices, based on how the two input images are generated, when assuming gap is always zero:
\begin{enumerate}
	\item Same noise variables, and same latent codes except one
	\item Same noise variables, and random latent codes except one
	\item Random noise variables, and same latent codes except one
	\item Random noise variables, and random latent codes except one
\end{enumerate}

We tried all four settings in MNIST dataset, using architecture in \cite{CDH16}, with 1 10-category latent codes, 2 continuous latent codes, and 62 noise variables. We had 8 runs for each setting, and visually inspect whether it correctly disentangle digit, rotation, and width. The disentanglement successful rate are recorded in table \ref{tbl:mnist}.

\begin{table}[h]
	\centering
	\begin{tabular}{|c|c|}
		\hline
		Choice & Rate of Successful Disentanglement\\\hline
		1 & 5/8\\\hline
		2 & 8/8\\\hline
		3 & 2/8\\\hline
		4 & 6/8\\\hline
	\end{tabular}
	\caption{Rate of successful disentanglement using four choices of CR}	\label{tbl:mnist}
\end{table}

Based on this result, we choose to use option 2 as our starting point of designing CR.

\section{Model Centrality}
For all figure results of MC and UDR, each model is compared with all other models. 
For all numerical results of MC and UDR, in order to get more accurate comparison, each model is compared with randomly selected 80\% of other models. Such random selection is done 100 times, and the mean and standard error are shown.

\label{app:model_centrality}
\subsection{Our Model Selection Approach on InfoGAN-CR Models (\dsprites{} Dataset)}
We run 8 independent trials for each of the following 9 hyperparameter settings $\{\alpha=1,2,4\}\times \{\lambda=0.05,0.2,2.0\}$, and an additional 4 independent trials of $(\alpha=2,\lambda=0.05)$. There are 76 models in total. The progressive scheduling of those 76 runs is the same as supervised experiments: run InfoGAN ($\alpha=0$) for 288000 batches, and then continue running it with CR (gap=0) for additional 34560 batches (so that the total number of epoches is 28).

\begin{figure}[t]
    \centering
    \includegraphics[width=0.5\linewidth]{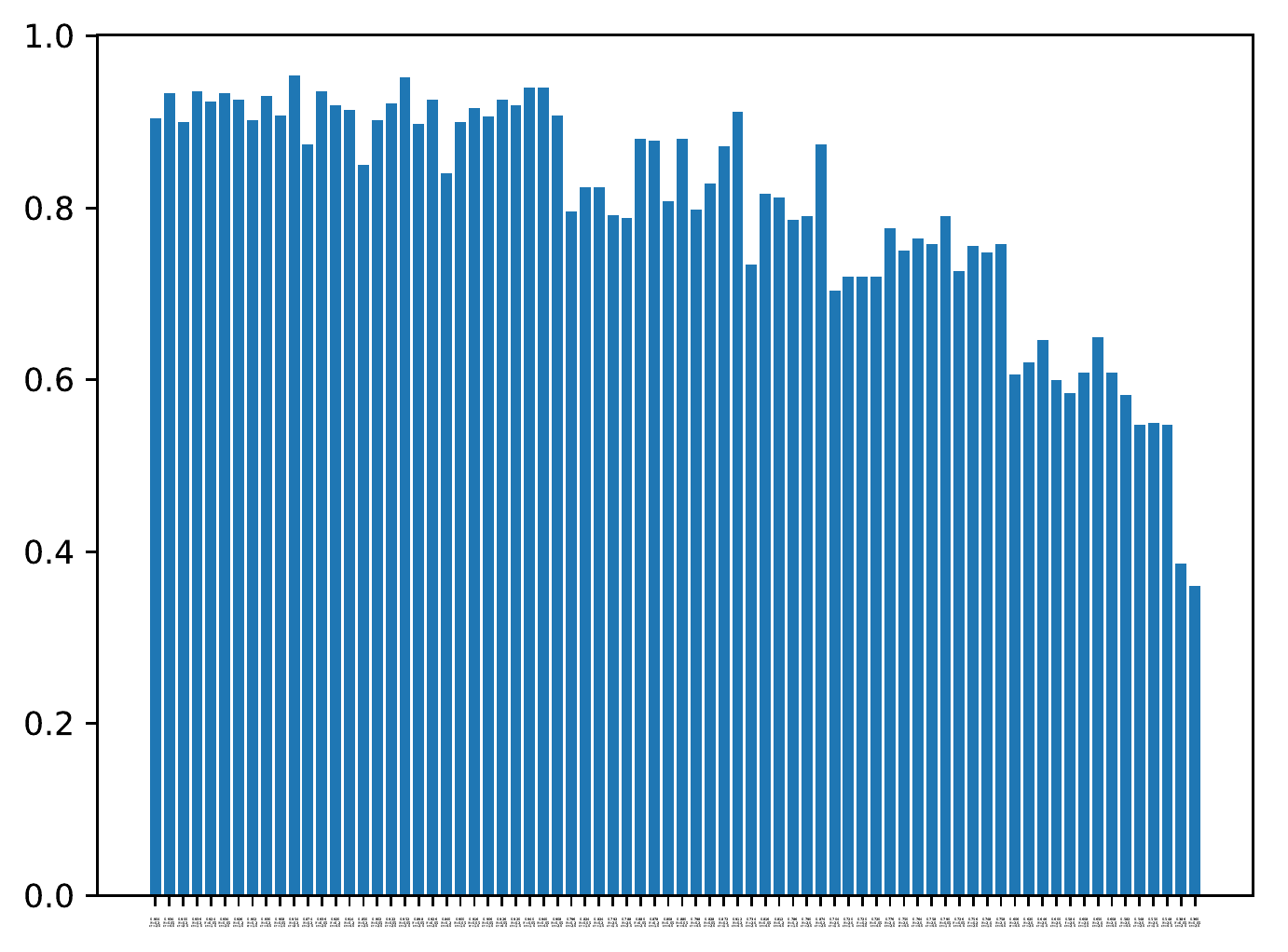}
    \caption{FactorVAE scores of InfoGAN-CR models (\dsprites{} dataset). The models are sorted according to our model selection score.}
    \label{fig:factorvae_bar_dSprites_infogancr_ours}
\end{figure}

\begin{figure}[t]
    \centering
    \includegraphics[width=0.5\linewidth]{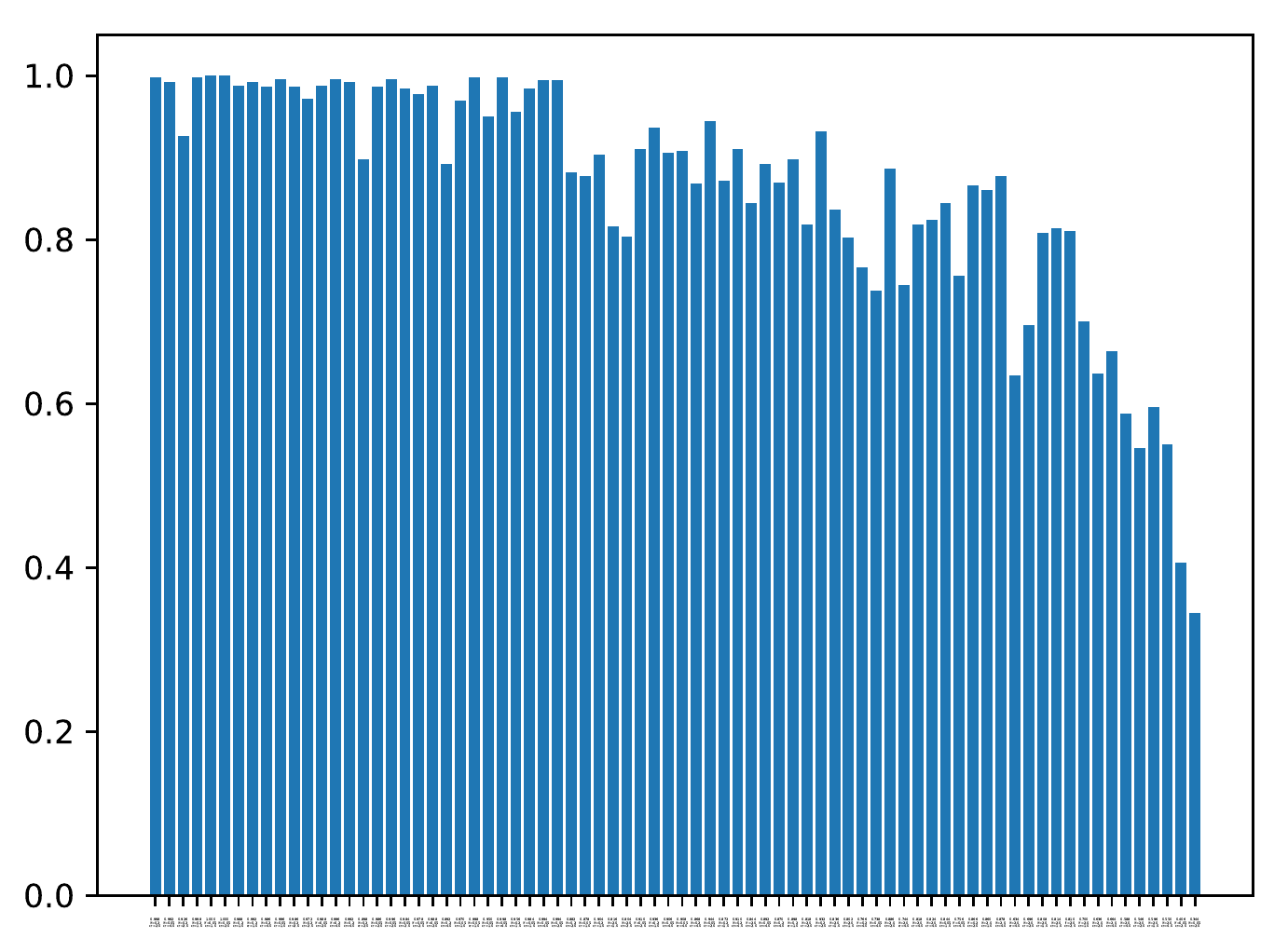}
    \caption{BetaVAE scores of InfoGAN-CR models (\dsprites{} dataset). The models are sorted according to our model selection score.}
    \label{fig:betavae_bar_dSprites_infogancr_ours}
\end{figure}

\begin{figure}[t]
    \centering
    \includegraphics[width=0.5\linewidth]{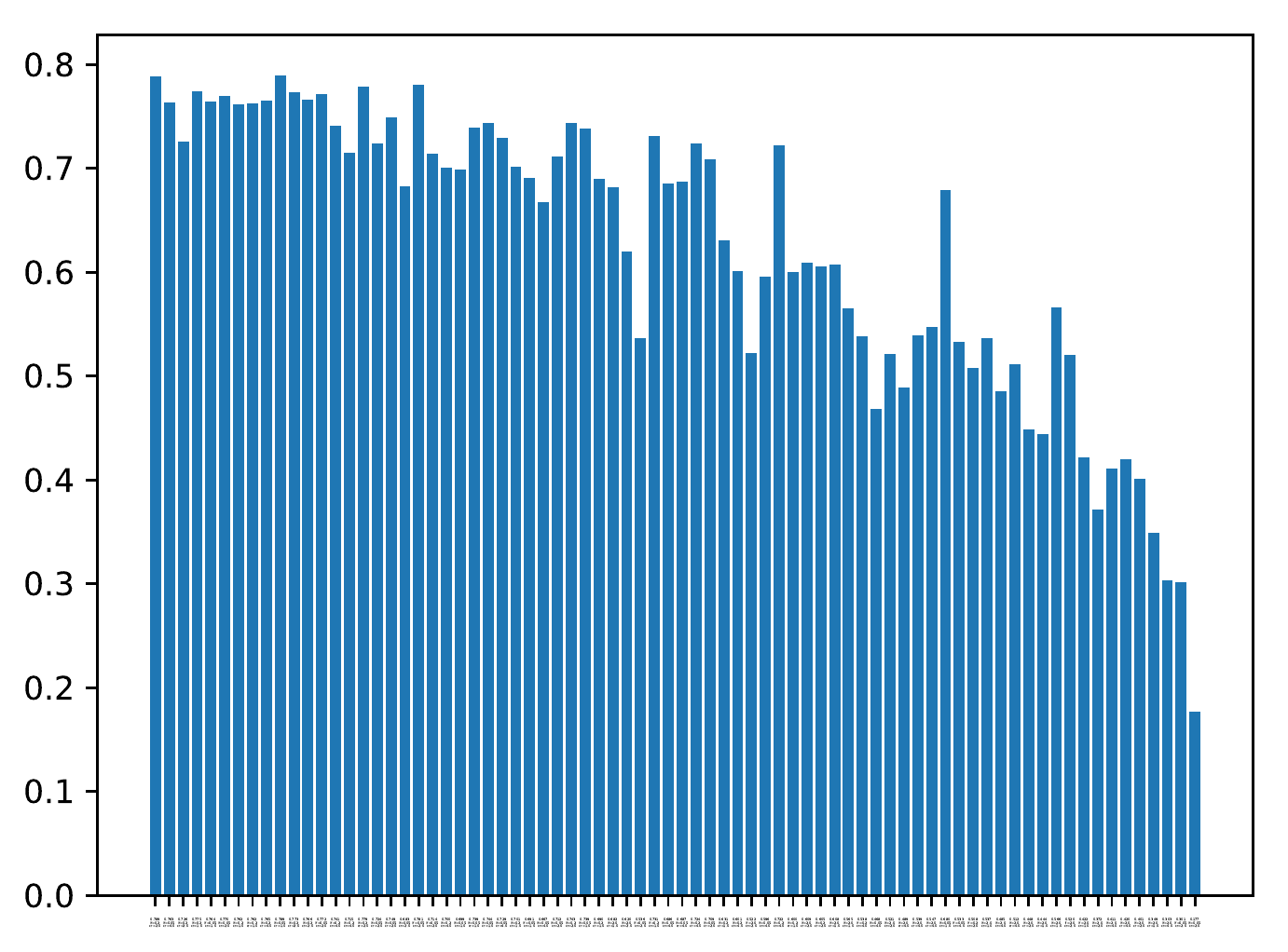}
    \caption{DCI scores of InfoGAN-CR models (\dsprites{} dataset). The models are sorted according to our model selection score.}
    \label{fig:dci_bar_dSprites_infogancr_ours}
\end{figure}

\begin{figure}[t]
    \centering
    \includegraphics[width=0.5\linewidth]{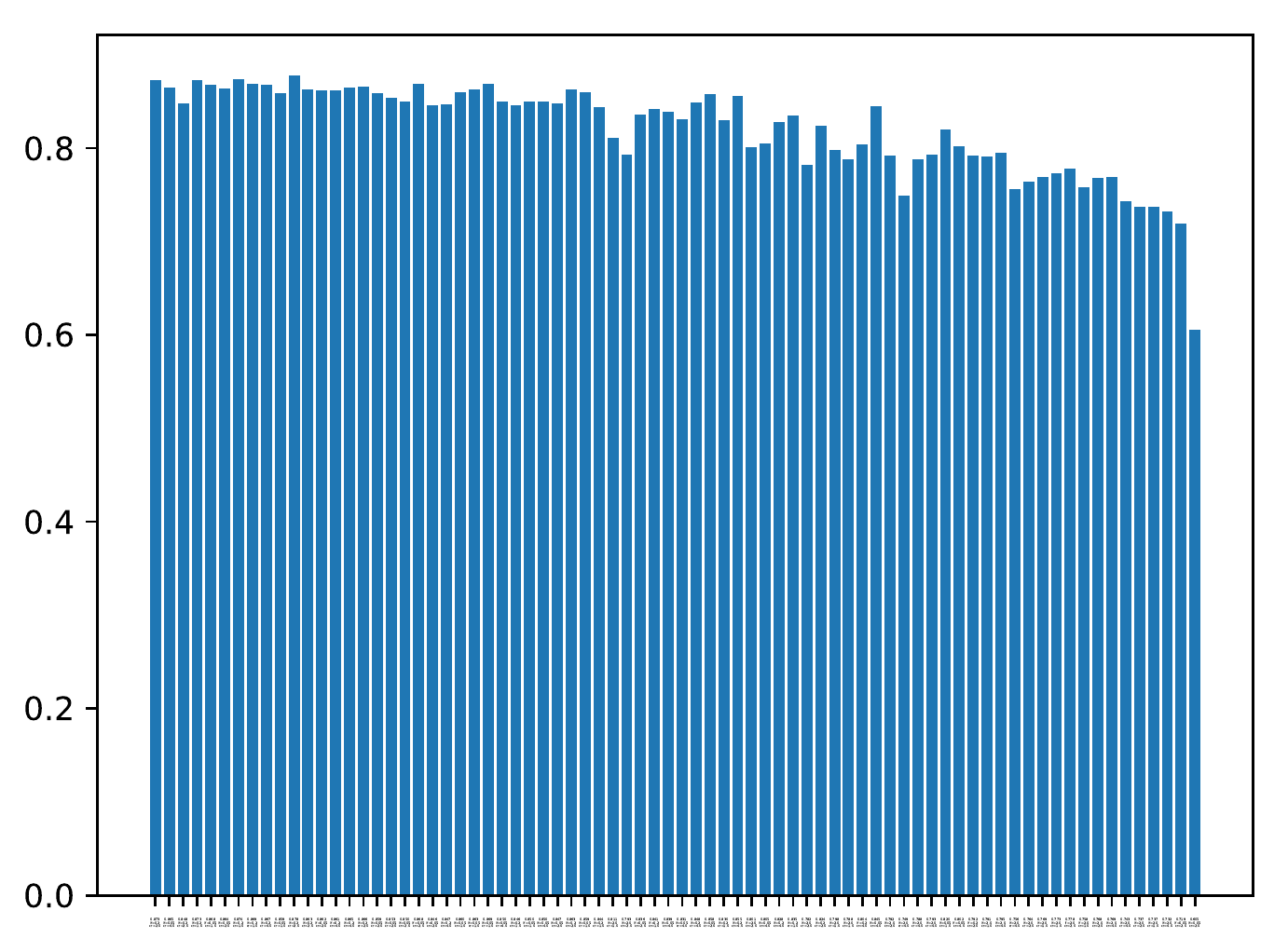}
    \caption{Explicitness scores of InfoGAN-CR models (\dsprites{} dataset). The models are sorted according to our model selection score.}
    \label{fig:expl_bar_dSprites_infogancr_ours}
\end{figure}

\begin{figure}[t]
    \centering
    \includegraphics[width=0.5\linewidth]{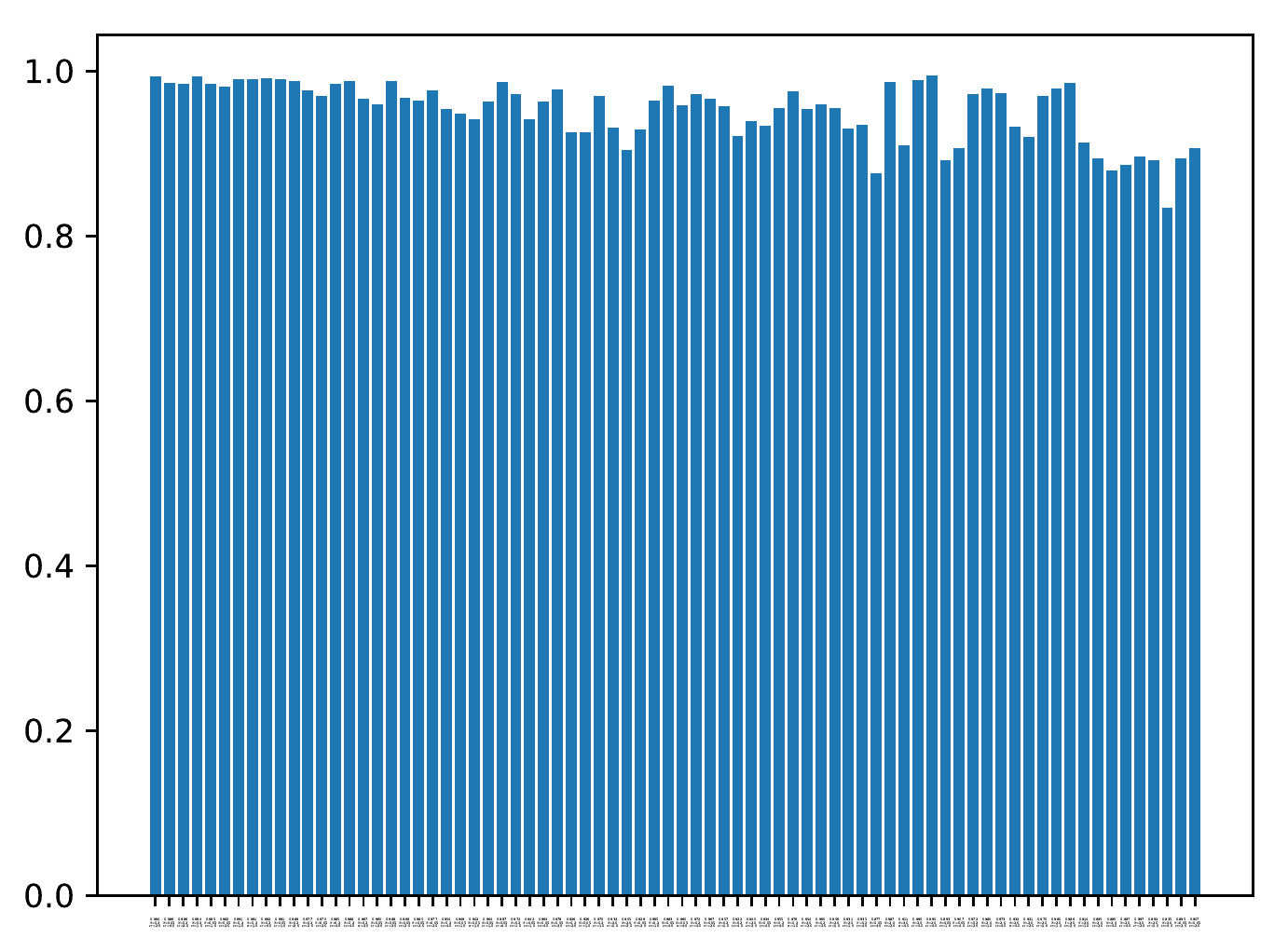}
    \caption{Modularity scores of InfoGAN-CR models (\dsprites{} dataset). The models are sorted according to our model selection score.}
    \label{fig:modu_bar_dSprites_infogancr_ours}
\end{figure}

\begin{figure}[t]
    \centering
    \includegraphics[width=0.5\linewidth]{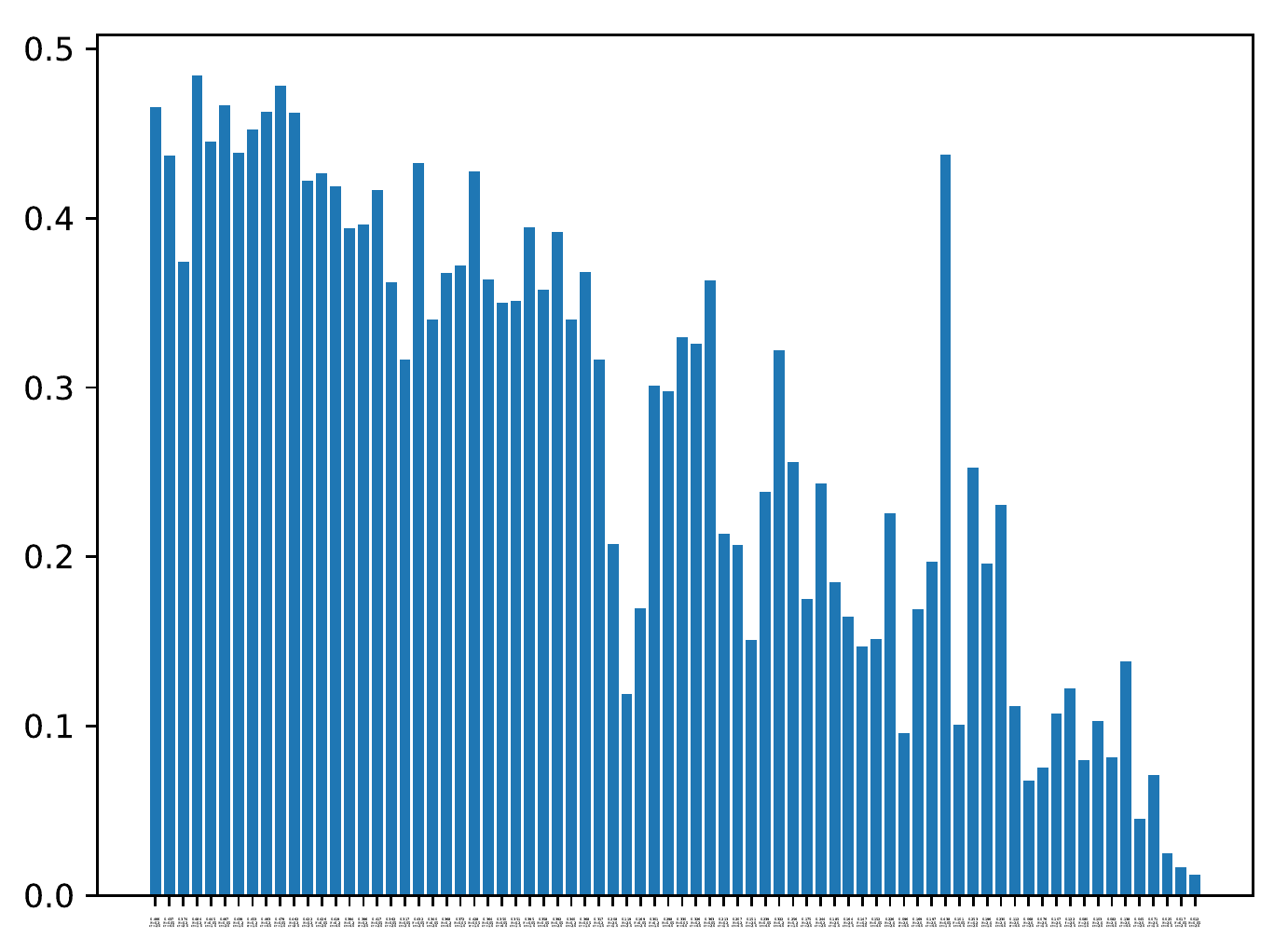}
    \caption{MIG scores of InfoGAN-CR models (\dsprites{} dataset). The models are sorted according to our model selection score.}
    \label{fig:mig_bar_dSprites_infogancr_ours}
\end{figure}

\begin{figure}[t]
    \centering
    \includegraphics[width=0.5\linewidth]{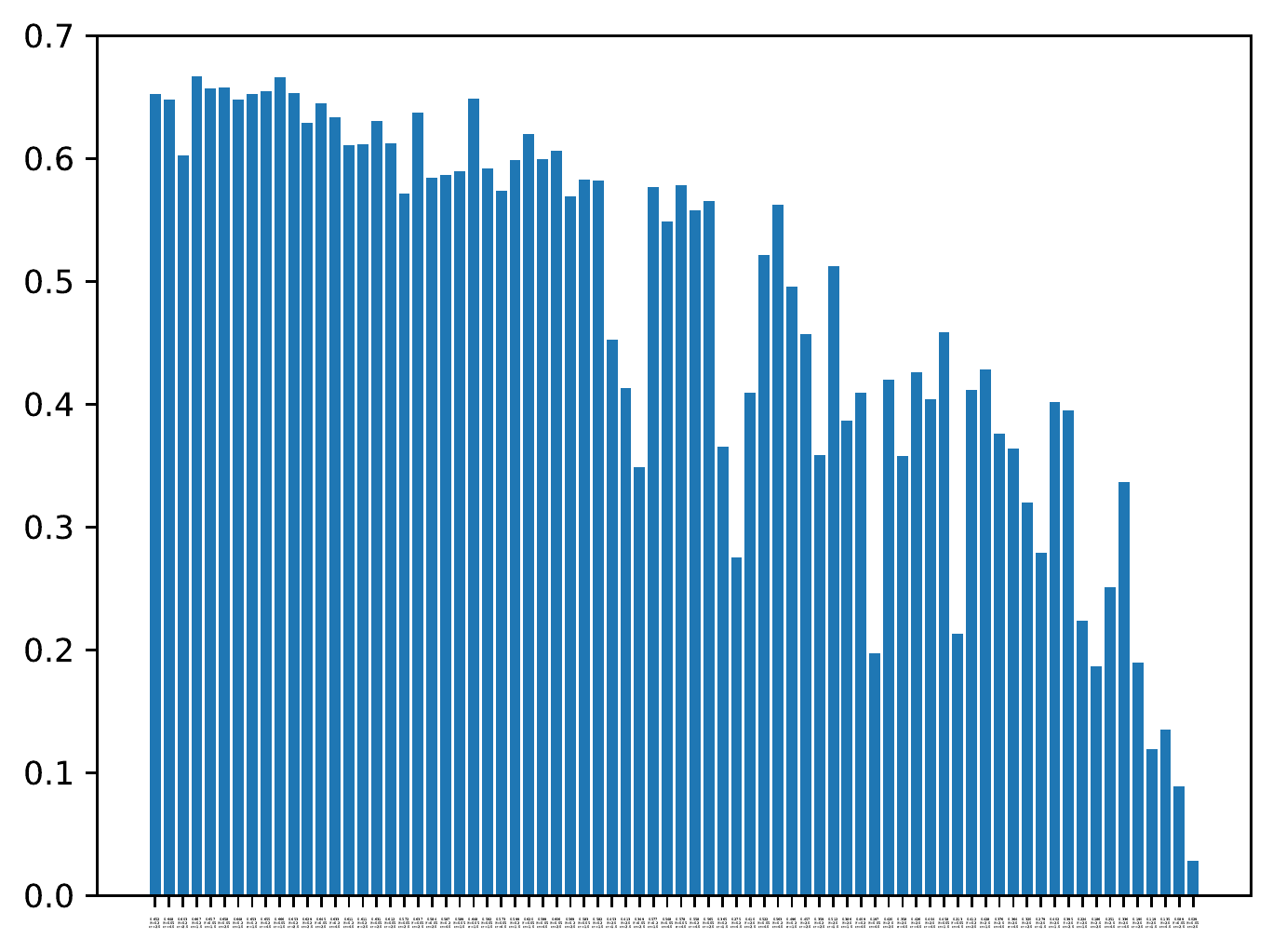}
    \caption{SAP scores of InfoGAN-CR models (\dsprites{} dataset). The models are sorted according to our model selection score.}
    \label{fig:sap_bar_dSprites_infogancr_ours}
\end{figure}

Figure \ref{fig:factorvae_bar_dSprites_infogancr_ours}, \ref{fig:betavae_bar_dSprites_infogancr_ours}, \ref{fig:dci_bar_dSprites_infogancr_ours},\ref{fig:expl_bar_dSprites_infogancr_ours},\ref{fig:modu_bar_dSprites_infogancr_ours},\ref{fig:mig_bar_dSprites_infogancr_ours},\ref{fig:sap_bar_dSprites_infogancr_ours} show the bar plots of supervised metrics, where the models are ordered according to our model selection approach. These figures show that our model selection approach correlates with other supervised metrics well.

\clearpage

\subsection{Our Model Selection Approach on FactorVAE Models (dSprites Dataset)}

We run 10 independent trials for each of the following 4 hyperparameter settings $tc=\{1,10,20,40\}$. There are 40 models in total. Each model is run for 27 epoches.

\begin{figure}[t]
    \centering
    \includegraphics[width=0.5\linewidth]{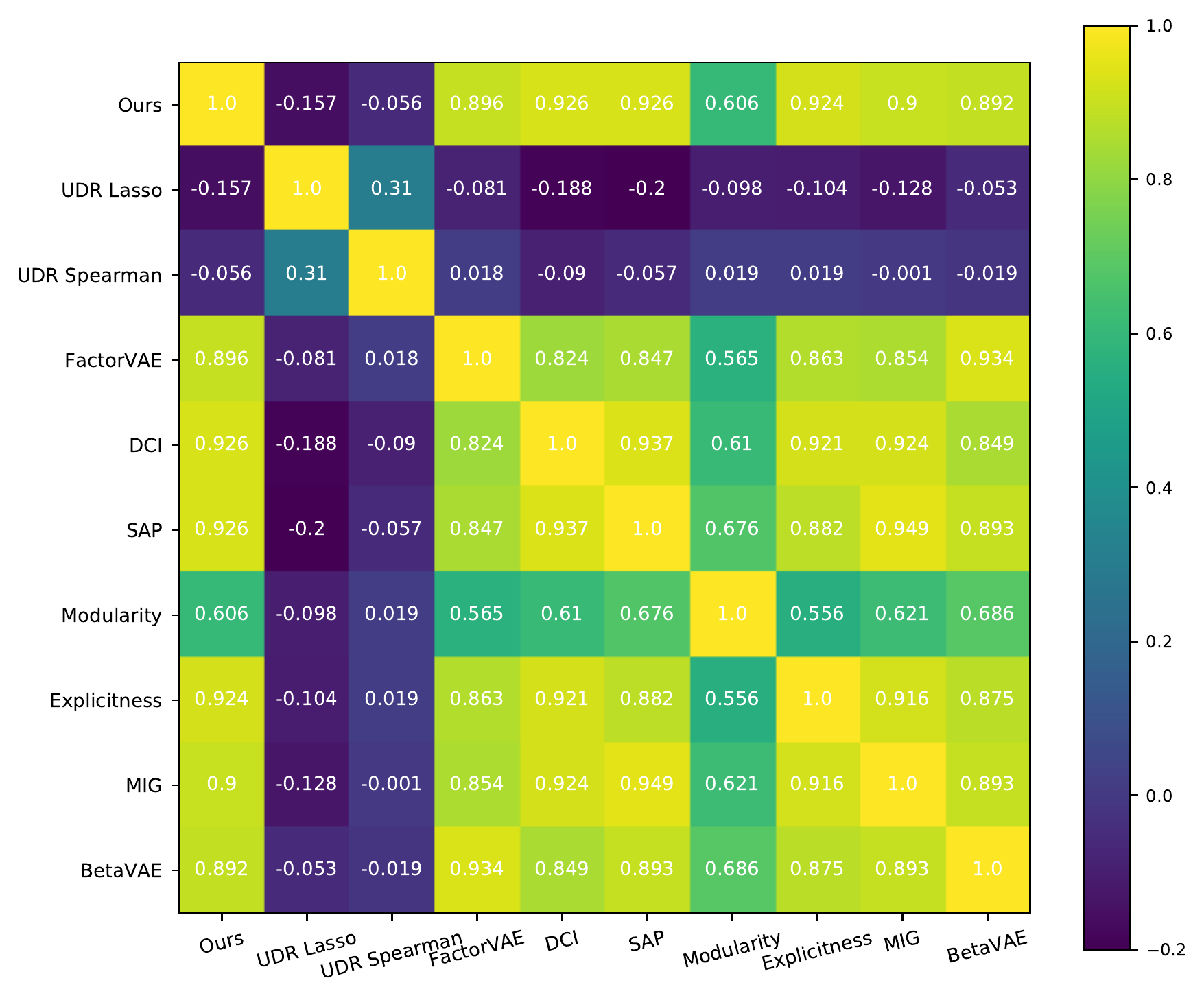}
    \caption{Spearman rank correlation of different metrics on FactorVAE models (\dsprites{} dataset).}
    \label{fig:metric_corr_dSprites_factorvae}
\end{figure}

Figure \ref{fig:metric_corr_dSprites_factorvae} shows the correlations between different metrics (including UDR and our model selection approach). It is clear that the score given by our model selection approach is strongly and positively correlated with other supervised metrics. On the other hand, we can see that UDR methods (Lasso or Spearman) are weakly correlated with other metrics. This suggests that our model selection approach is very effective for model selection, whereas UDR performs badly. 

\begin{figure}[t]
    \centering
    \includegraphics[width=0.5\linewidth]{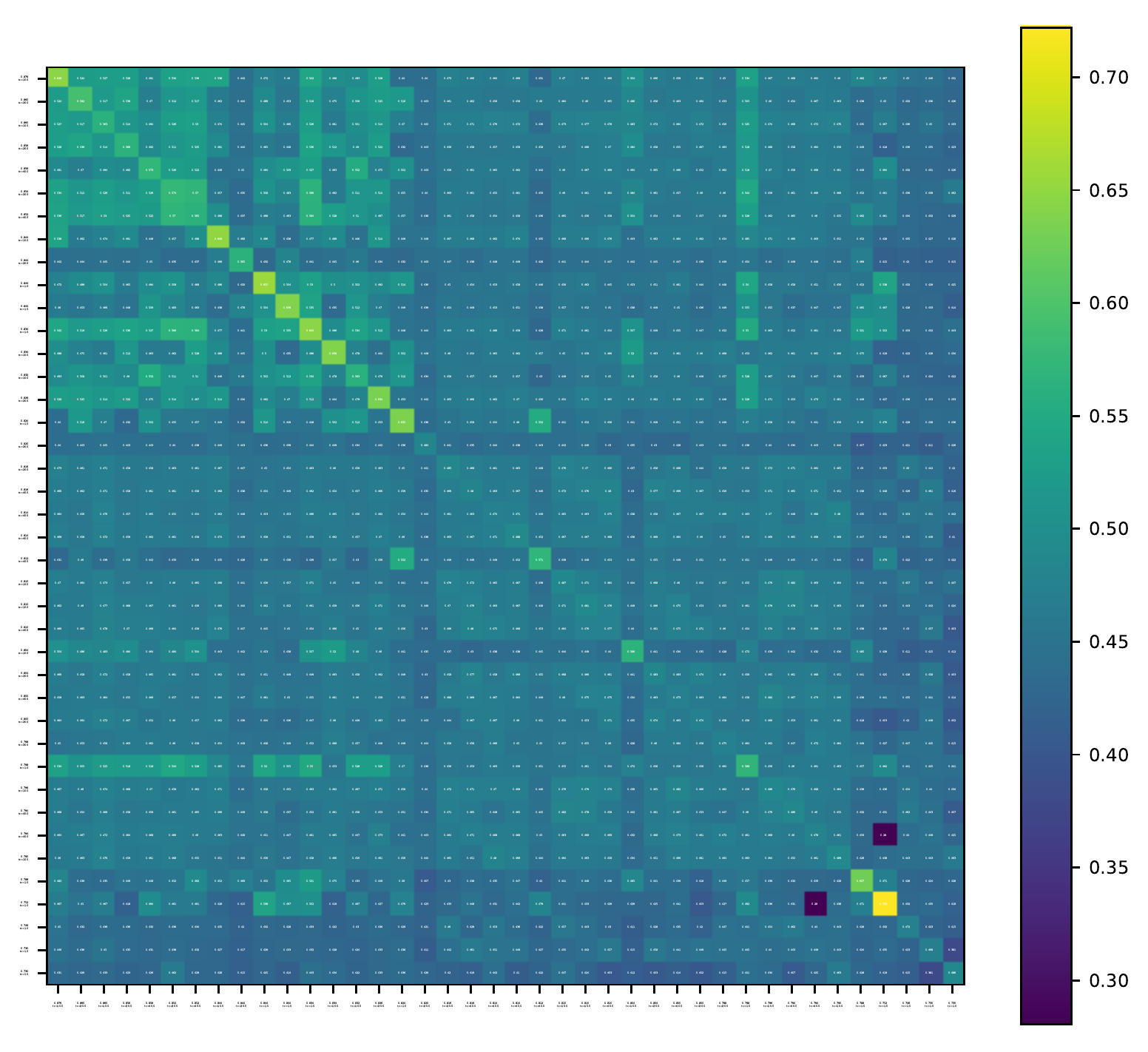}
    \caption{Our model selection approach on FactorVAE models (\dsprites{} dataset). Each row/column corresponds to an FactorVAE model. The models are sorted according to FactorVAE metric. $(i, j)$ entry represents the cross-evaluation score of $i$-th and $j$-th model using our model selection approach.}
    \label{fig:model_corr_dSprites_factorvae}
\end{figure}

Figure \ref{fig:model_corr_dSprites_factorvae} shows the cross-model score given by our model selection approach. Generally we can see that the score between good models are larger. This suggests that our model selection approach can effectively select the good model pairs.

\begin{figure}[t]
    \centering
    \includegraphics[width=0.5\linewidth]{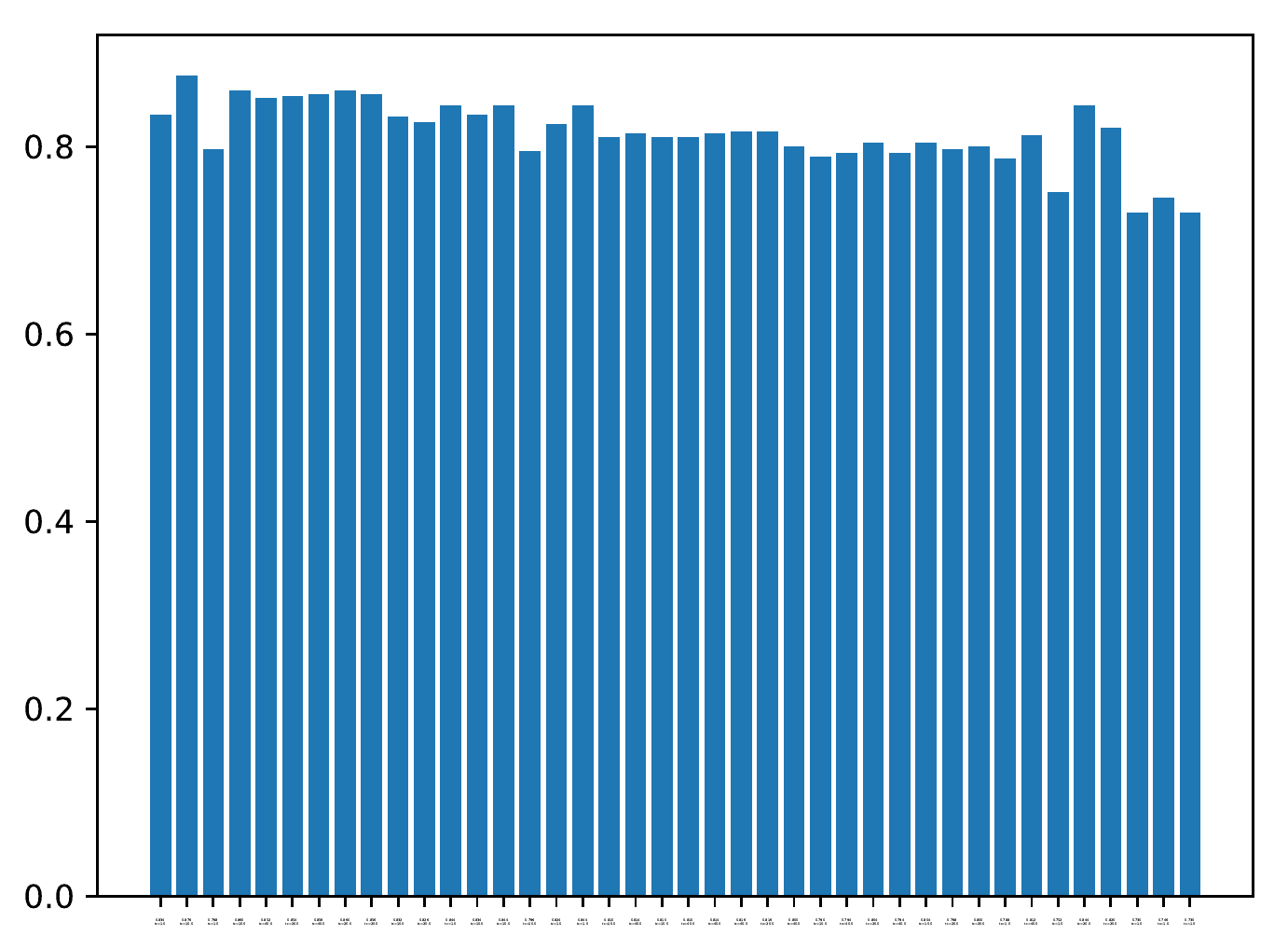}
    \caption{FactorVAE scores of FactorVAE models (\dsprites{} dataset). The models are sorted according to our model selection score.}
    \label{fig:factorvae_bar_dSprites_factorvae_ours}
\end{figure}

\begin{figure}[t]
    \centering
    \includegraphics[width=0.5\linewidth]{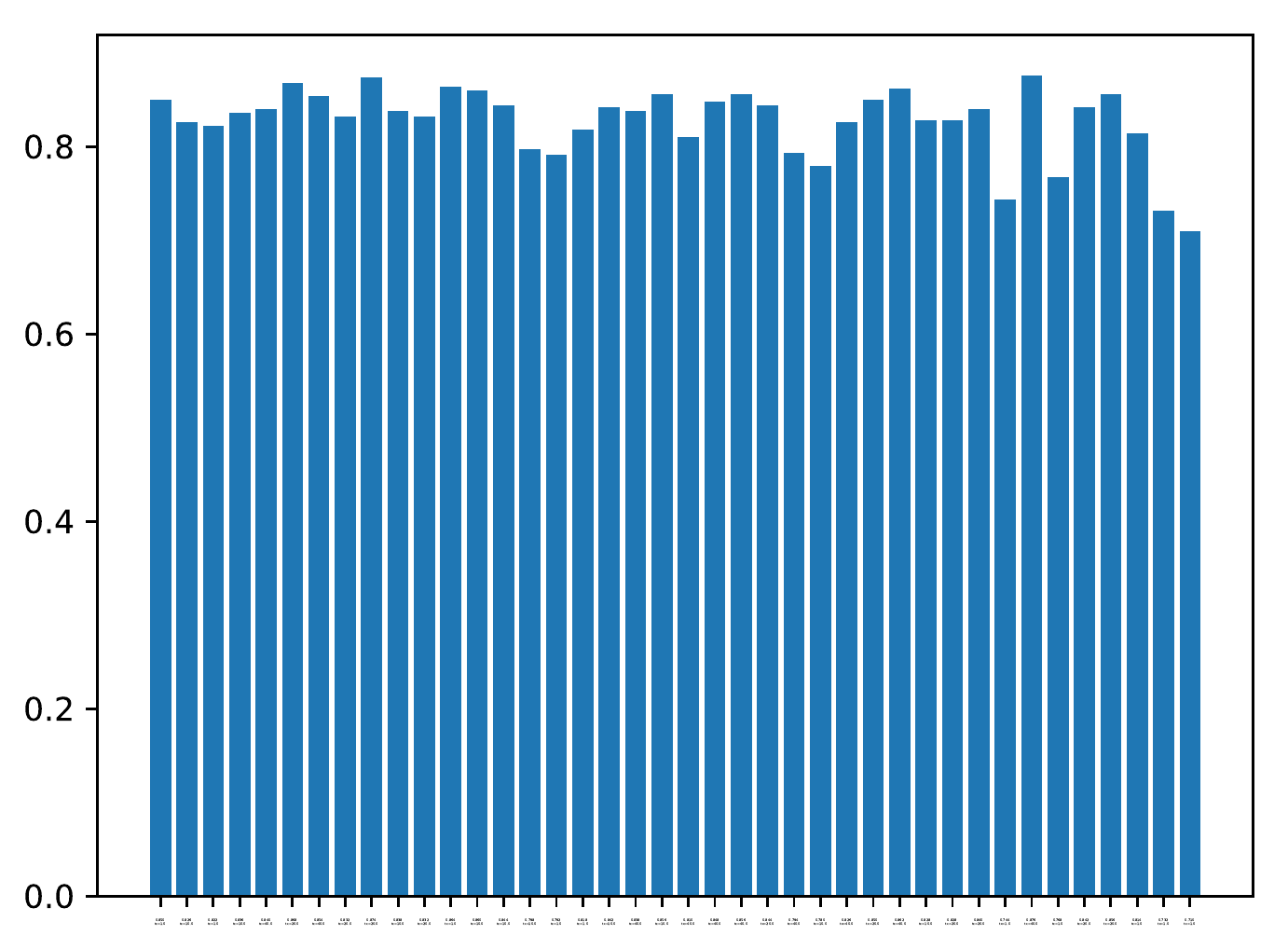}
    \caption{BetaVAE scores of FactorVAE models (\dsprites{} dataset). The models are sorted according to our model selection score.}
    \label{fig:betavae_bar_dSprites_factorvae_ours}
\end{figure}

\begin{figure}[t]
    \centering
    \includegraphics[width=0.5\linewidth]{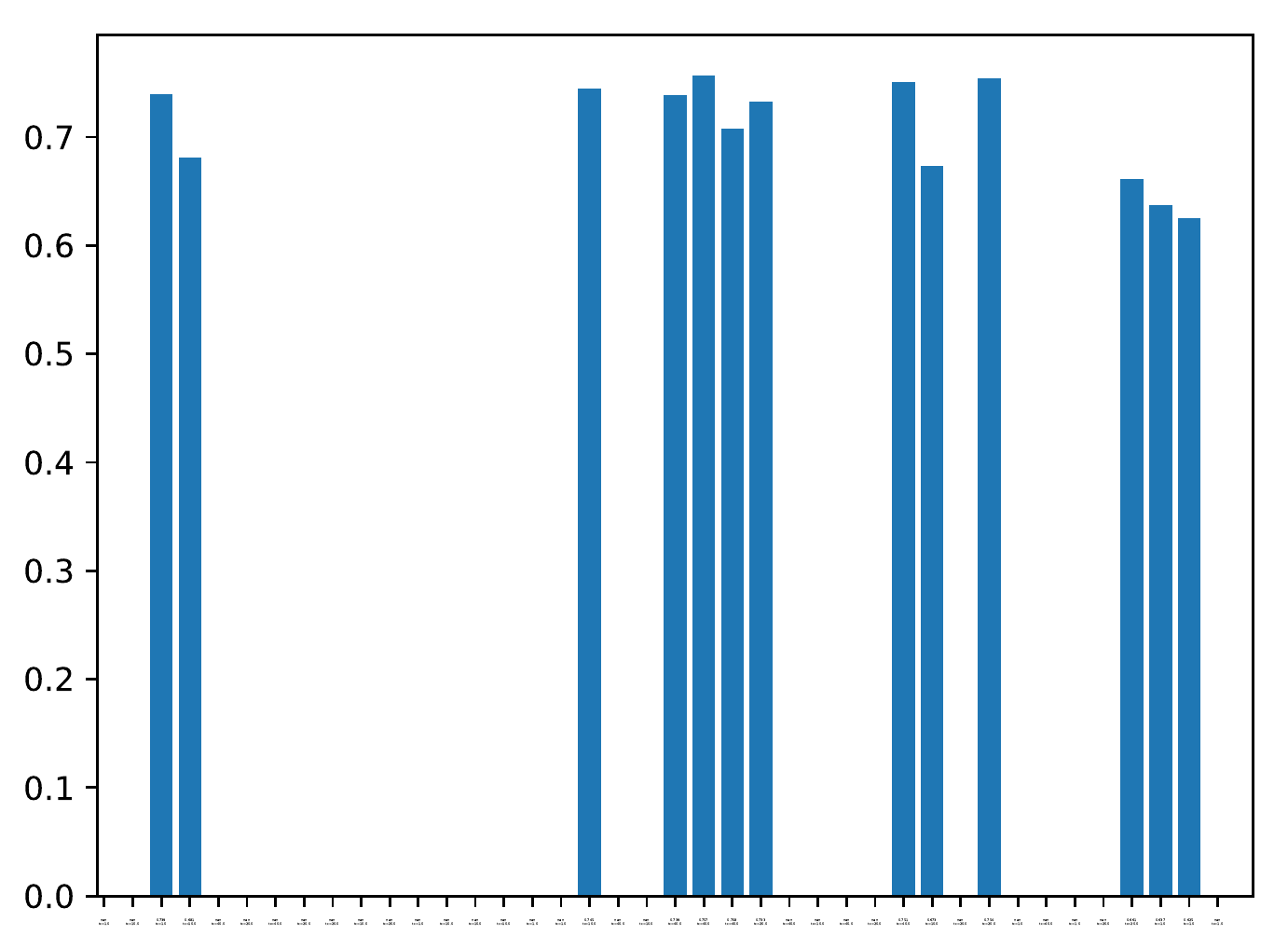}
    \caption{DCI scores of FactorVAE models (\dsprites{} dataset). The models are sorted according to our model selection score. Note that some of the DCI score gives NaN values and therefore some bars are missing.}
    \label{fig:dci_bar_dSprites_factorvae_ours}
\end{figure}

\begin{figure}[t]
    \centering
    \includegraphics[width=0.5\linewidth]{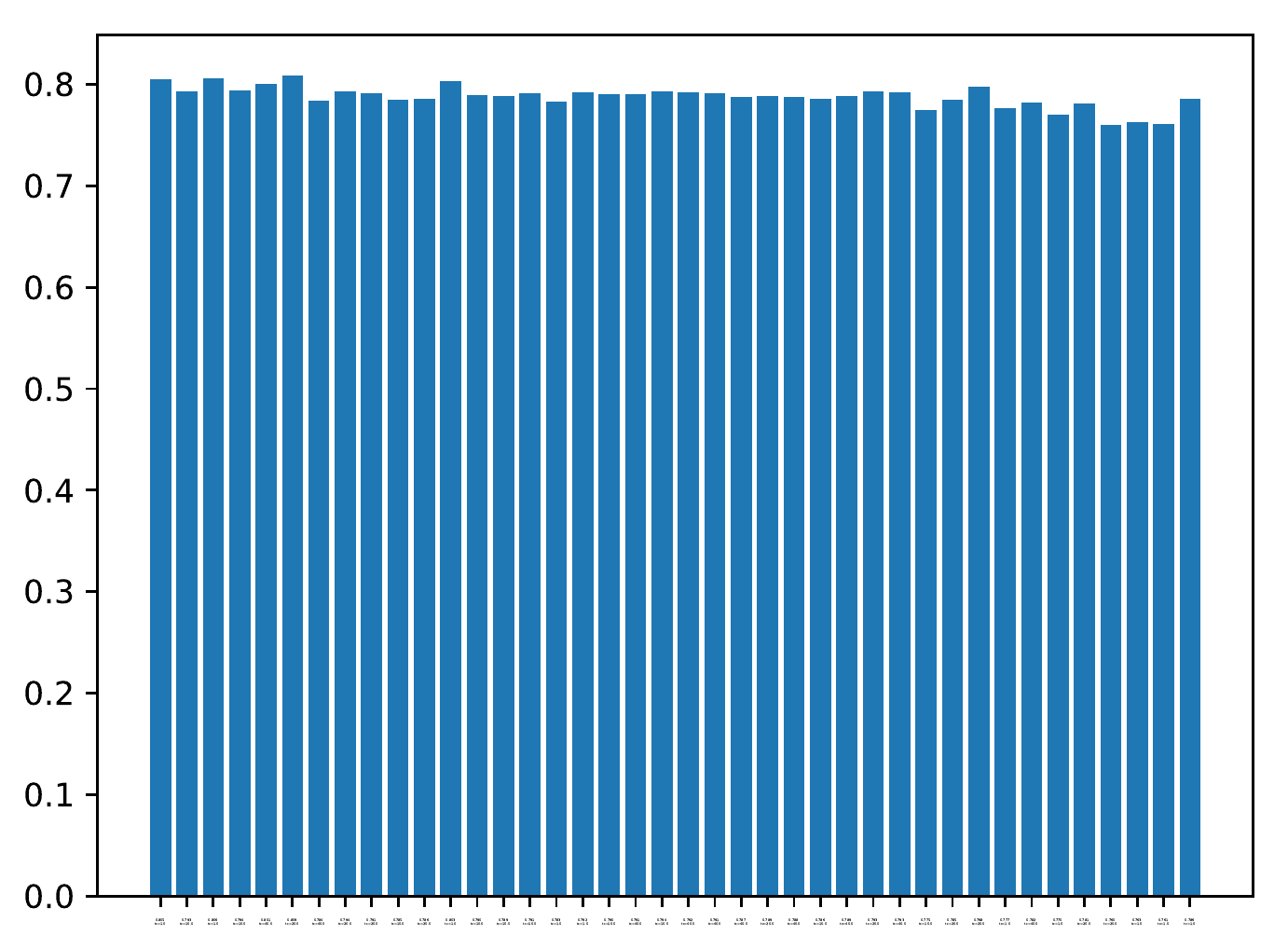}
    \caption{Explicitness scores of FactorVAE models (\dsprites{} dataset). The models are sorted according to our model selection score.}
    \label{fig:expl_bar_dSprites_factorvae_ours}
\end{figure}

\begin{figure}[t]
    \centering
    \includegraphics[width=0.5\linewidth]{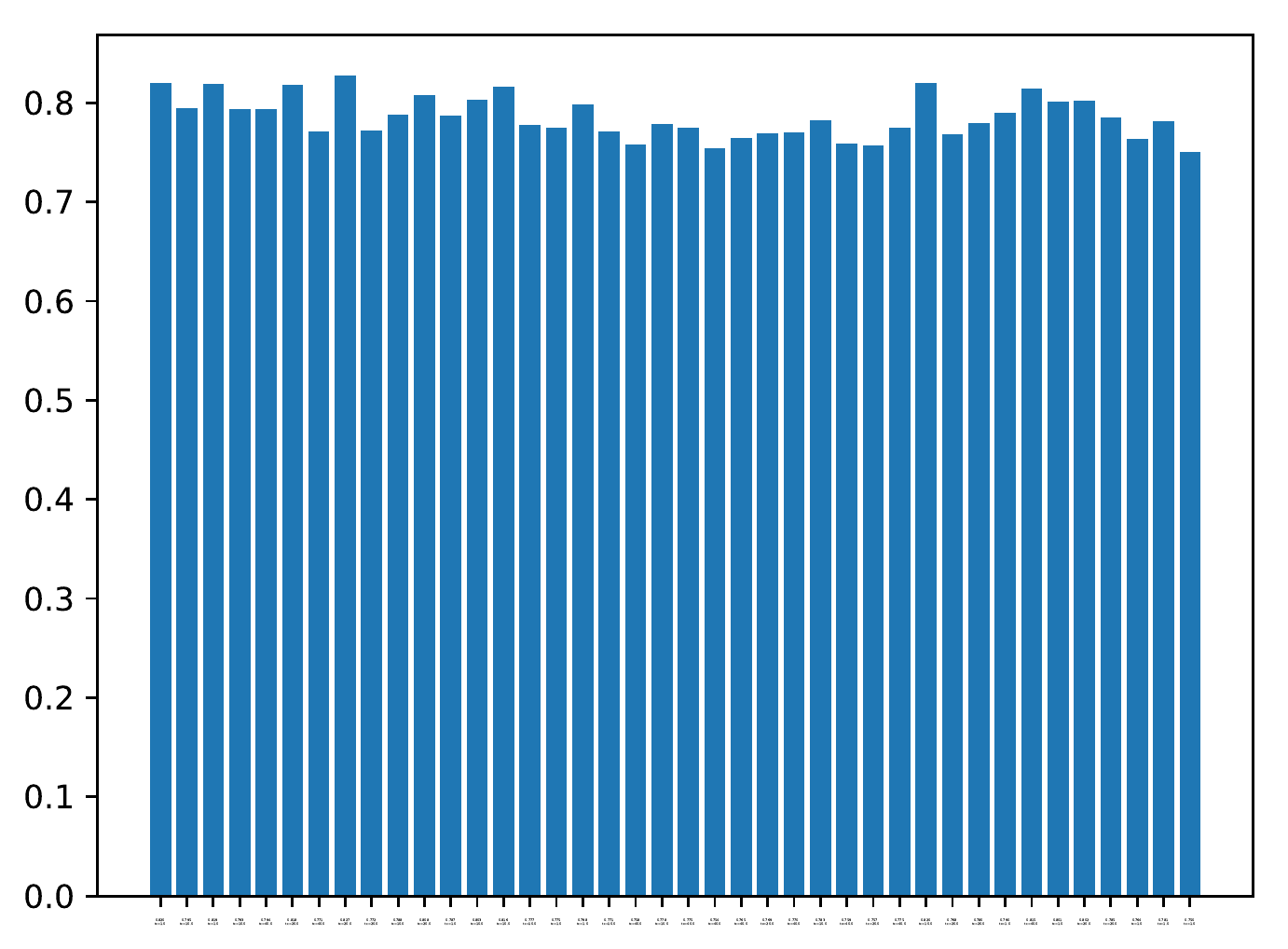}
    \caption{Modularity scores of FactorVAE models (\dsprites{} dataset). The models are sorted according to our model selection score.}
    \label{fig:modu_bar_dSprites_factorvae_ours}
\end{figure}

\begin{figure}[t]
    \centering
    \includegraphics[width=0.5\linewidth]{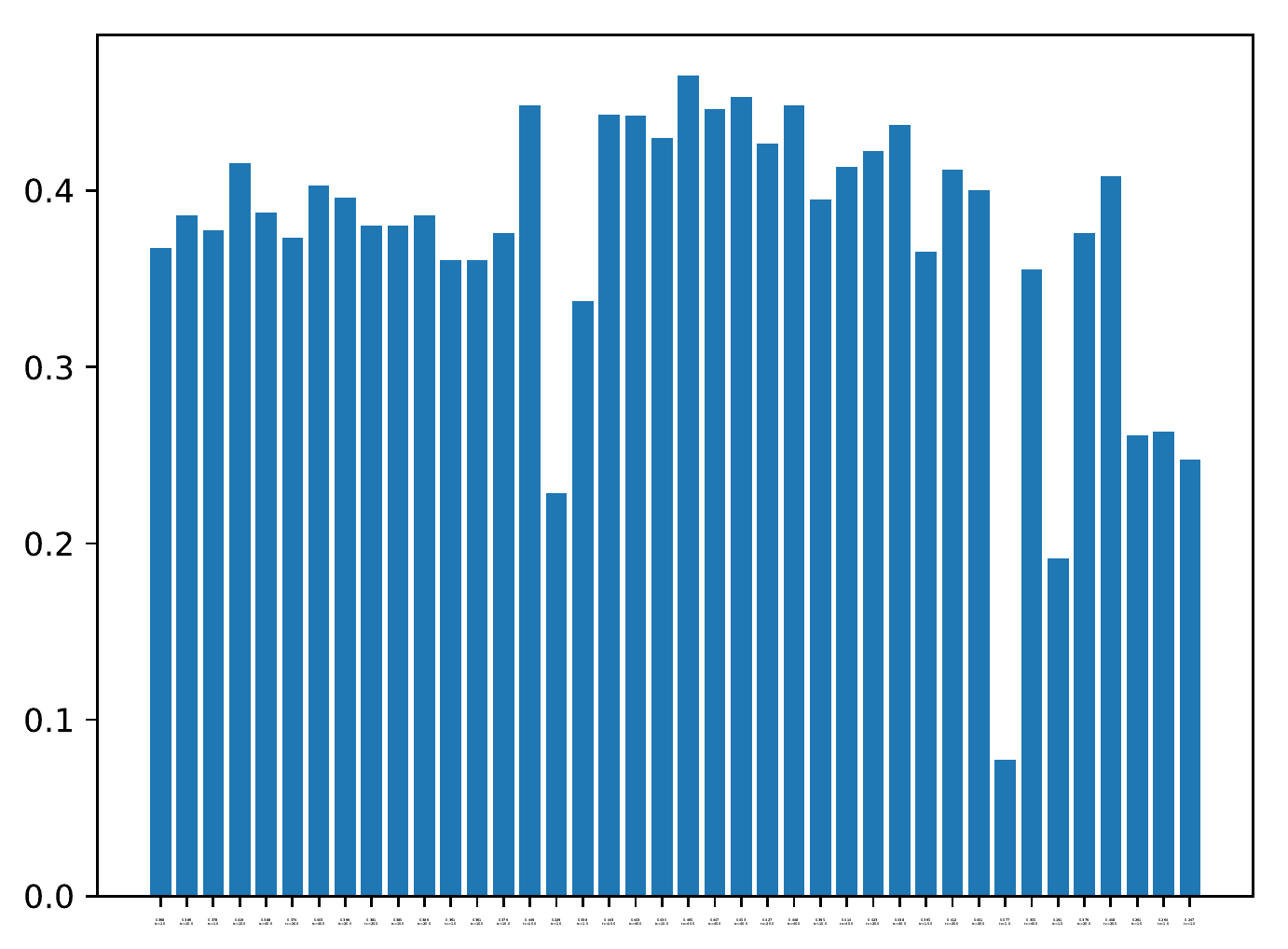}
    \caption{MIG scores of FactorVAE models (\dsprites{} dataset). The models are sorted according to our model selection score.}
    \label{fig:mig_bar_dSprites_factorvae_ours}
\end{figure}

\begin{figure}[t]
    \centering
    \includegraphics[width=0.5\linewidth]{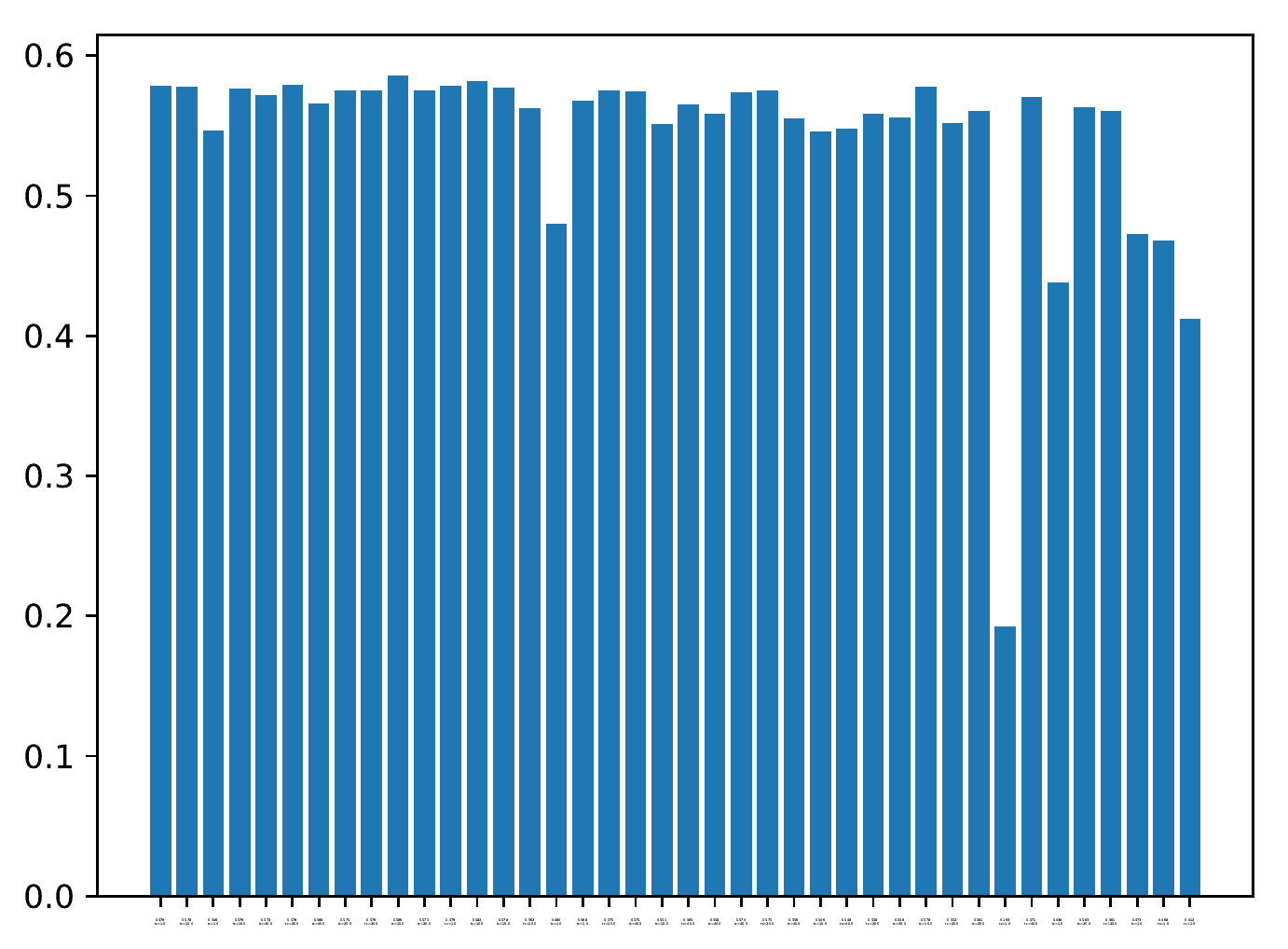}
    \caption{SAP scores of FactorVAE models (\dsprites{} dataset). The models are sorted according to our model selection score.}
    \label{fig:sap_bar_dSprites_factorvae_ours}
\end{figure}

Figure \ref{fig:factorvae_bar_dSprites_factorvae_ours}, \ref{fig:betavae_bar_dSprites_factorvae_ours}, \ref{fig:dci_bar_dSprites_factorvae_ours},\ref{fig:expl_bar_dSprites_factorvae_ours},\ref{fig:modu_bar_dSprites_factorvae_ours},\ref{fig:mig_bar_dSprites_factorvae_ours},\ref{fig:sap_bar_dSprites_factorvae_ours} show the bar plots of supervised metrics, where the models are ordered according to our model selection approach. These figures show that our model selection approach correlates with other supervised metrics well.

Table \ref{tbl:detailed1} shows that the model selected with ModelCentrality outperforms UDR in most metrics.

\clearpage

\subsection{Our Model Selection Approach on InfoGAN-CR Models (\teapots{} Dataset)}

We run 10 independent trials for each of the following 9 hyperparameter settings $\{\alpha=1.5,3.0,6.0\}\times \{\lambda=0.05,0.2,2.0\}$. There are 90 models in total. The progressive scheduling is the same as the supervised experiments: we train InfoGAN for 50000 batches, and then
InfoGAN-CR with gap=1.9 for 35000 batches, and then
InfoGAN-CR with gap=0.0 for 40000 batches.

\begin{figure}[t]
    \centering
    \includegraphics[width=0.5\linewidth]{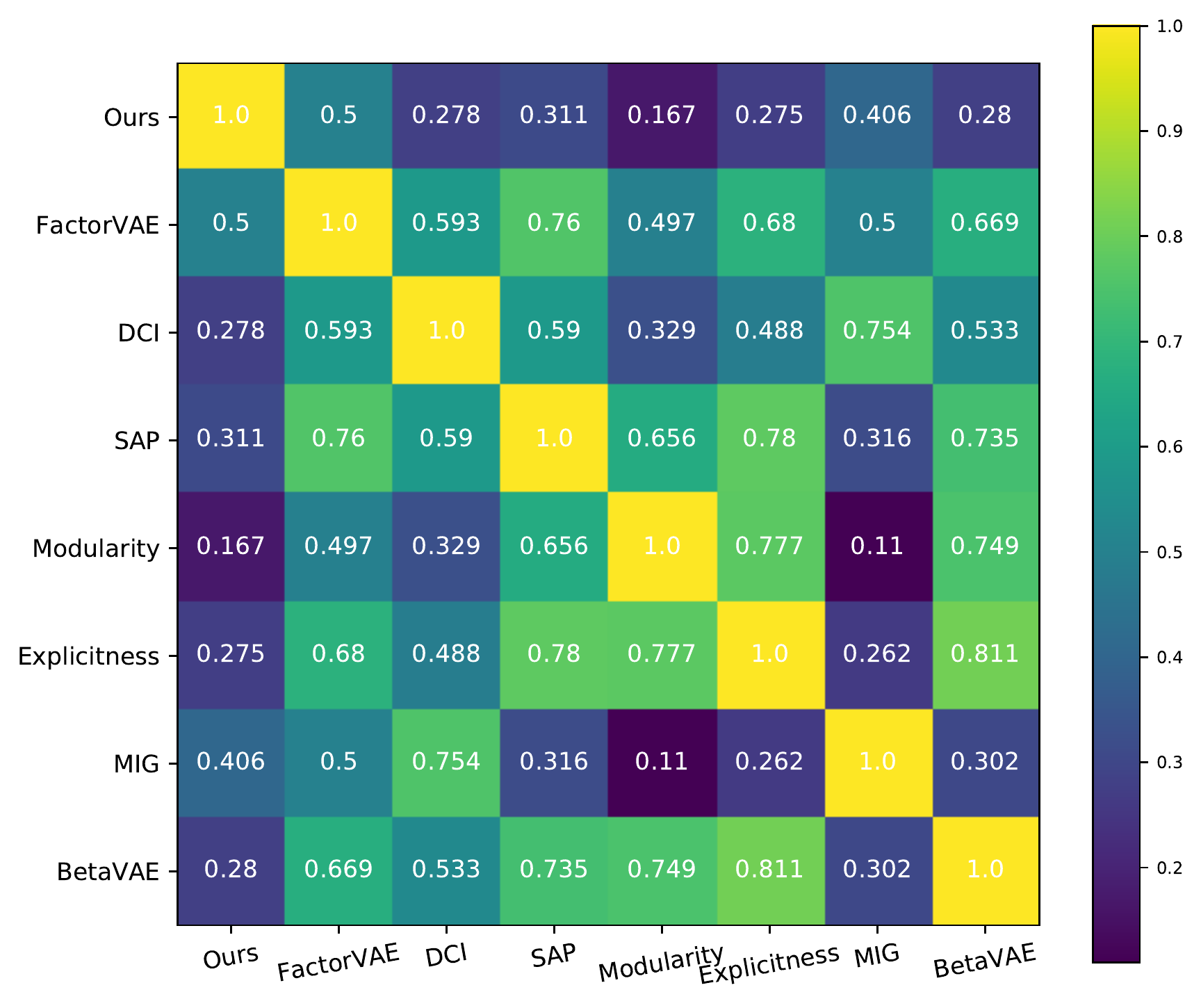}
    \caption{Spearman rank correlation of different metrics on InfoGAN-CR models (\teapots{} dataset).}
    \label{fig:metric_corr_teapots_infogancr}
\end{figure}

Figure \ref{fig:metric_corr_teapots_infogancr} shows the correlations between different metrics (including UDR and our model selection approach). It is clear that the score given by our model selection approach is positively correlated with other supervised metrics. Though the correlation is not as positive as in \dsprites{} dataset, we see that the correlations between other supervised metrics are also weaker than the ones in \dsprites{} dataset. This may because this dataset is harder to interpret and evaluate for those metrics than \dsprites{}.

\begin{figure}[t]
    \centering
    \includegraphics[width=0.5\linewidth]{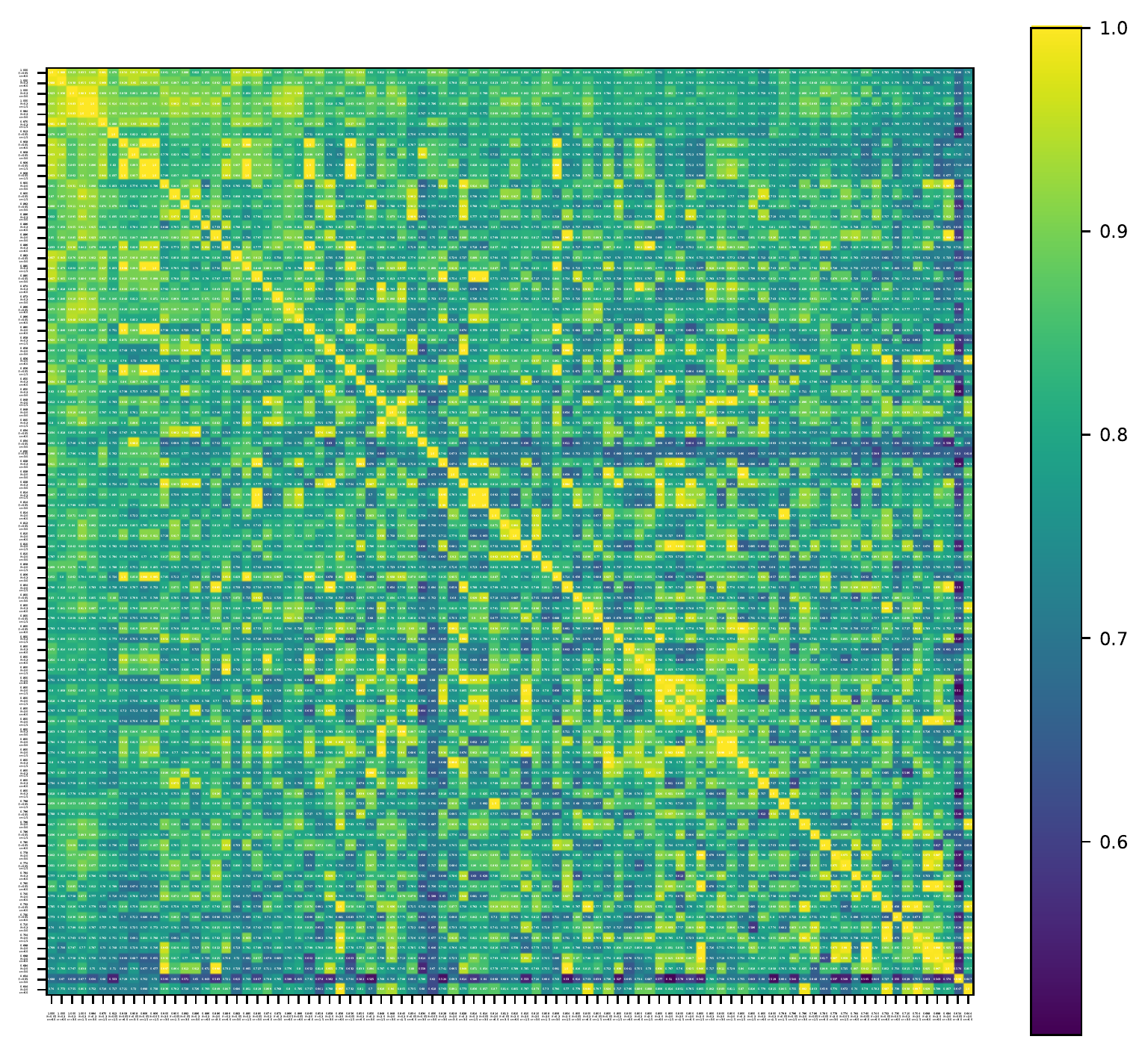}
    \caption{Our model selection approach on InfoGAN-CR models (\teapots{} dataset). Each row/column corresponds to an InfoGAN-CR model. The models are sorted according to FactorVAE metric. $(i, j)$ entry represents the cross-evaluation score of $i$-th and $j$-th model using our model selection approach.}
    \label{fig:model_corr_teapots_infogancr}
\end{figure}

Figure \ref{fig:model_corr_teapots_infogancr} shows the cross-model score given by our model selection approach. Generally we can see that the score between good models are larger, but the result is much more noisy than the one in \dsprites{}. Again, the reason might be that this dataset is more difficult.

\begin{figure}[t]
    \centering
    \includegraphics[width=0.5\linewidth]{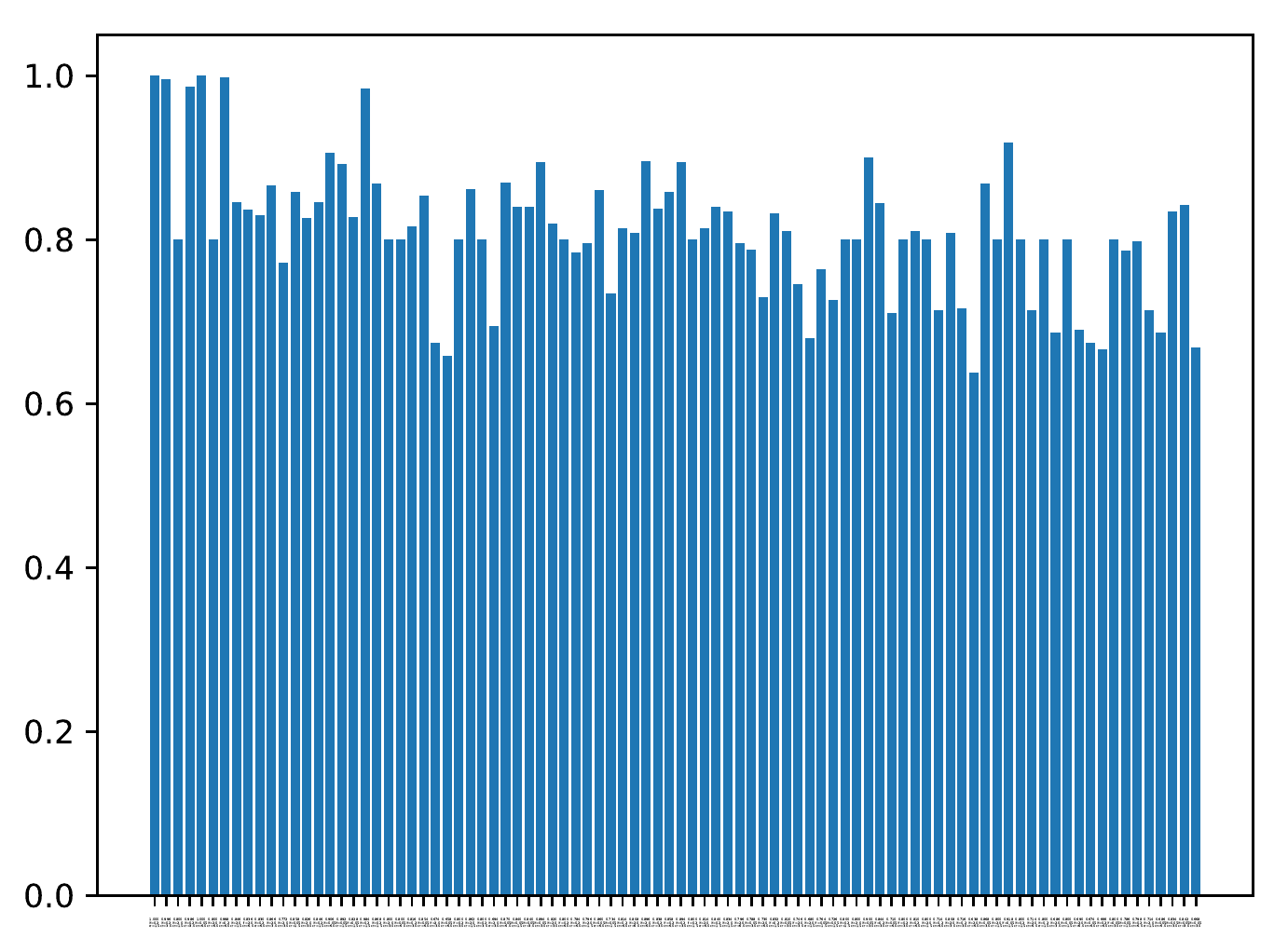}
    \caption{FactorVAE scores of InfoGAN-CR models (\teapots{} dataset). The models are sorted according to our model selection score.}
    \label{fig:factorvae_bar_teapots_infogancr_ours}
\end{figure}

\begin{figure}[t]
    \centering
    \includegraphics[width=0.5\linewidth]{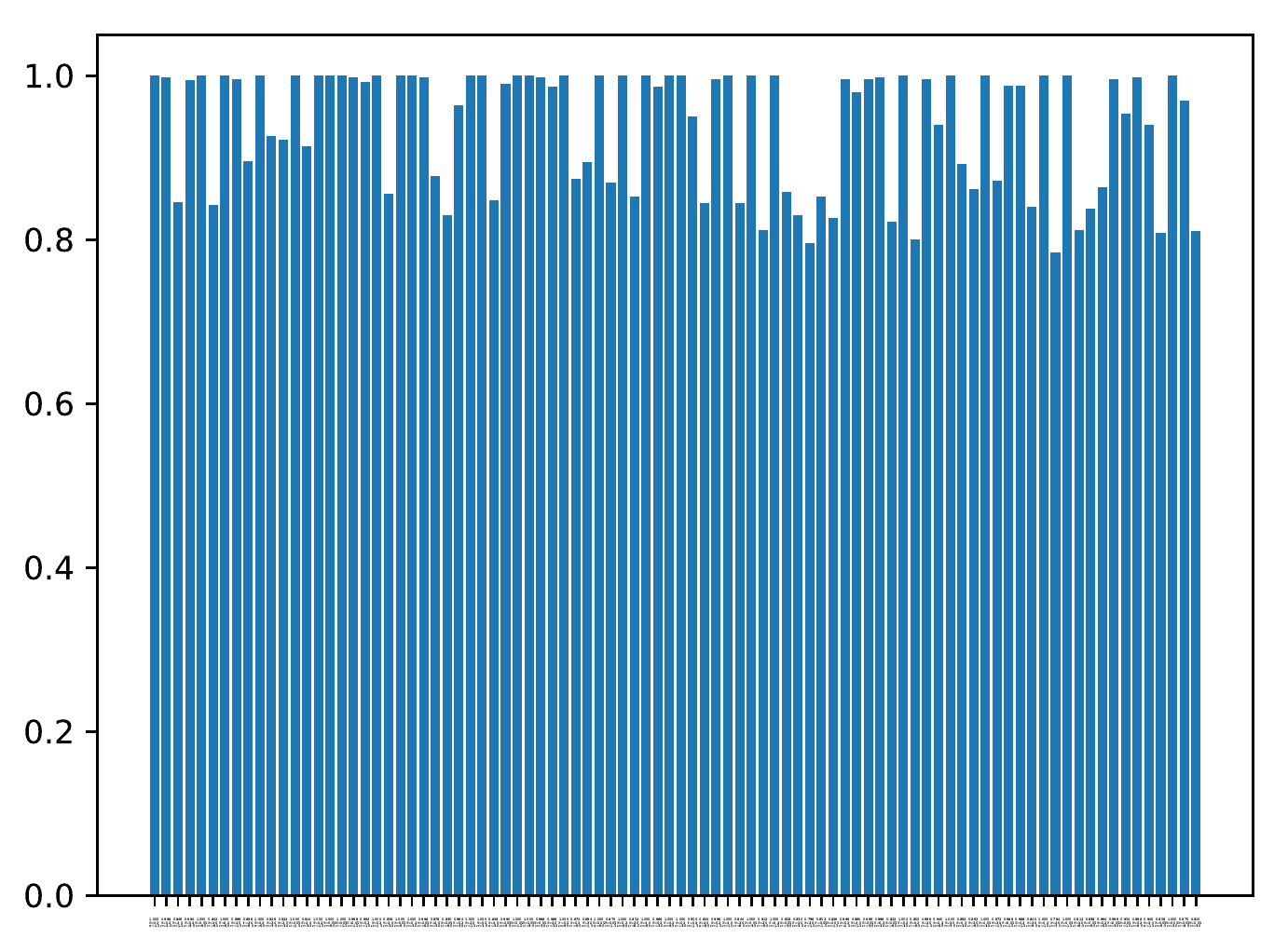}
    \caption{BetaVAE scores of InfoGAN-CR models (\teapots{} dataset). The models are sorted according to our model selection score.}
    \label{fig:betavae_bar_teapots_infogancr_ours}
\end{figure}

\begin{figure}[t]
    \centering
    \includegraphics[width=0.5\linewidth]{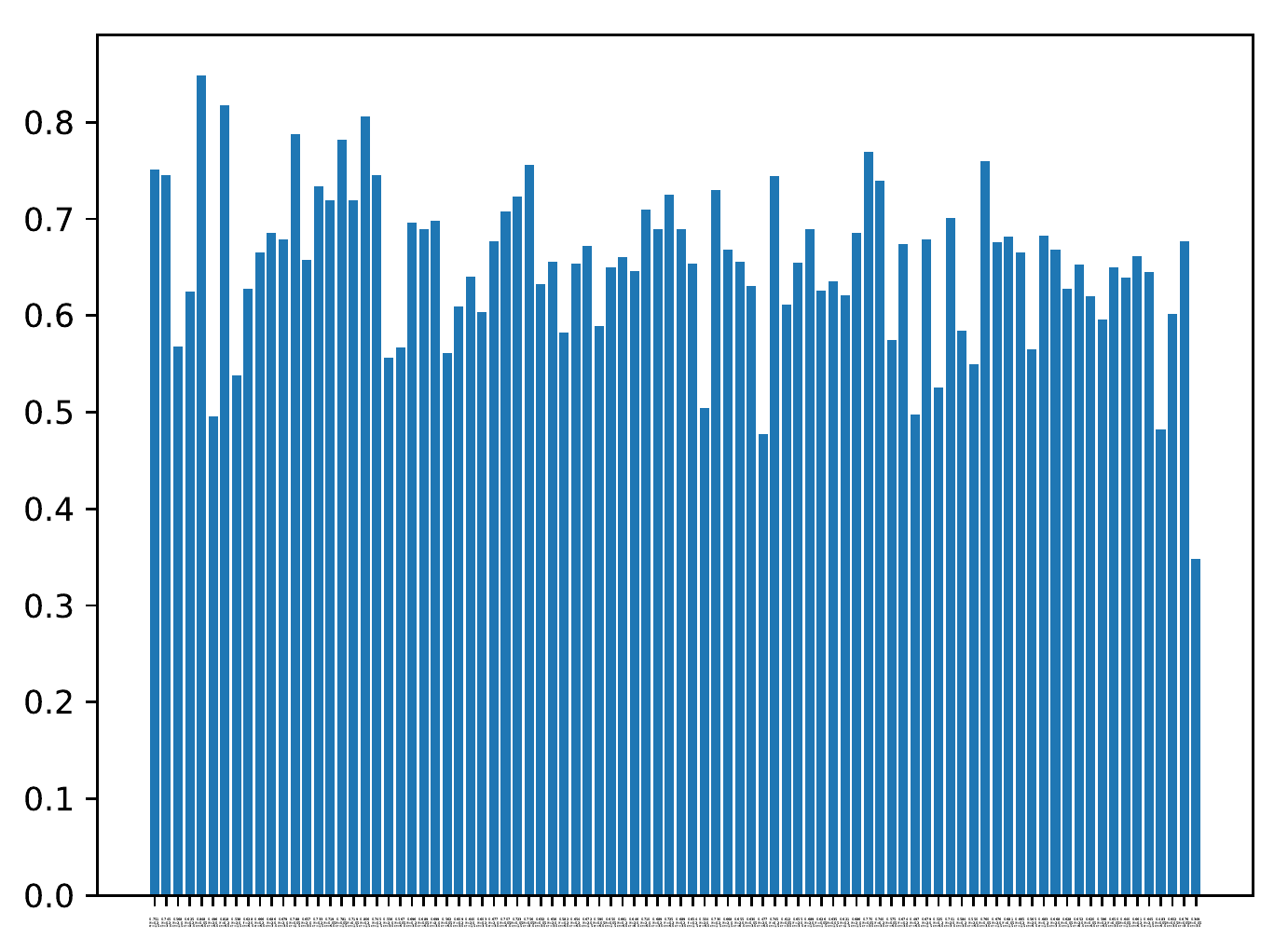}
    \caption{DCI scores of InfoGAN-CR models (\teapots{} dataset). The models are sorted according to our model selection score.}
    \label{fig:dci_bar_teapots_infogancr_ours}
\end{figure}

\begin{figure}[t]
    \centering
    \includegraphics[width=0.5\linewidth]{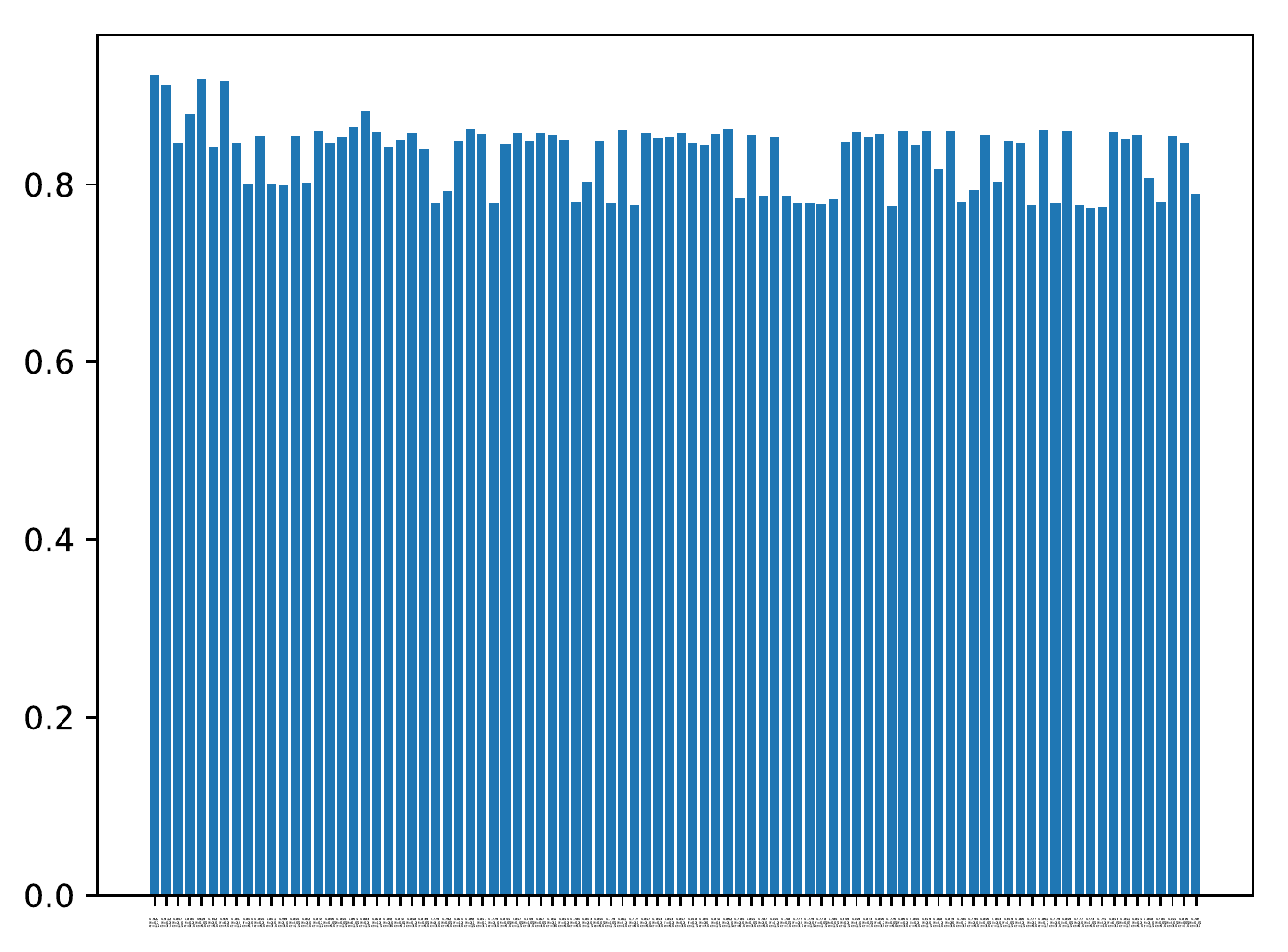}
    \caption{Explicitness scores of InfoGAN-CR models (\teapots{} dataset). The models are sorted according to our model selection score.}
    \label{fig:expl_bar_teapots_infogancr_ours}
\end{figure}

\begin{figure}[t]
    \centering
    \includegraphics[width=0.5\linewidth]{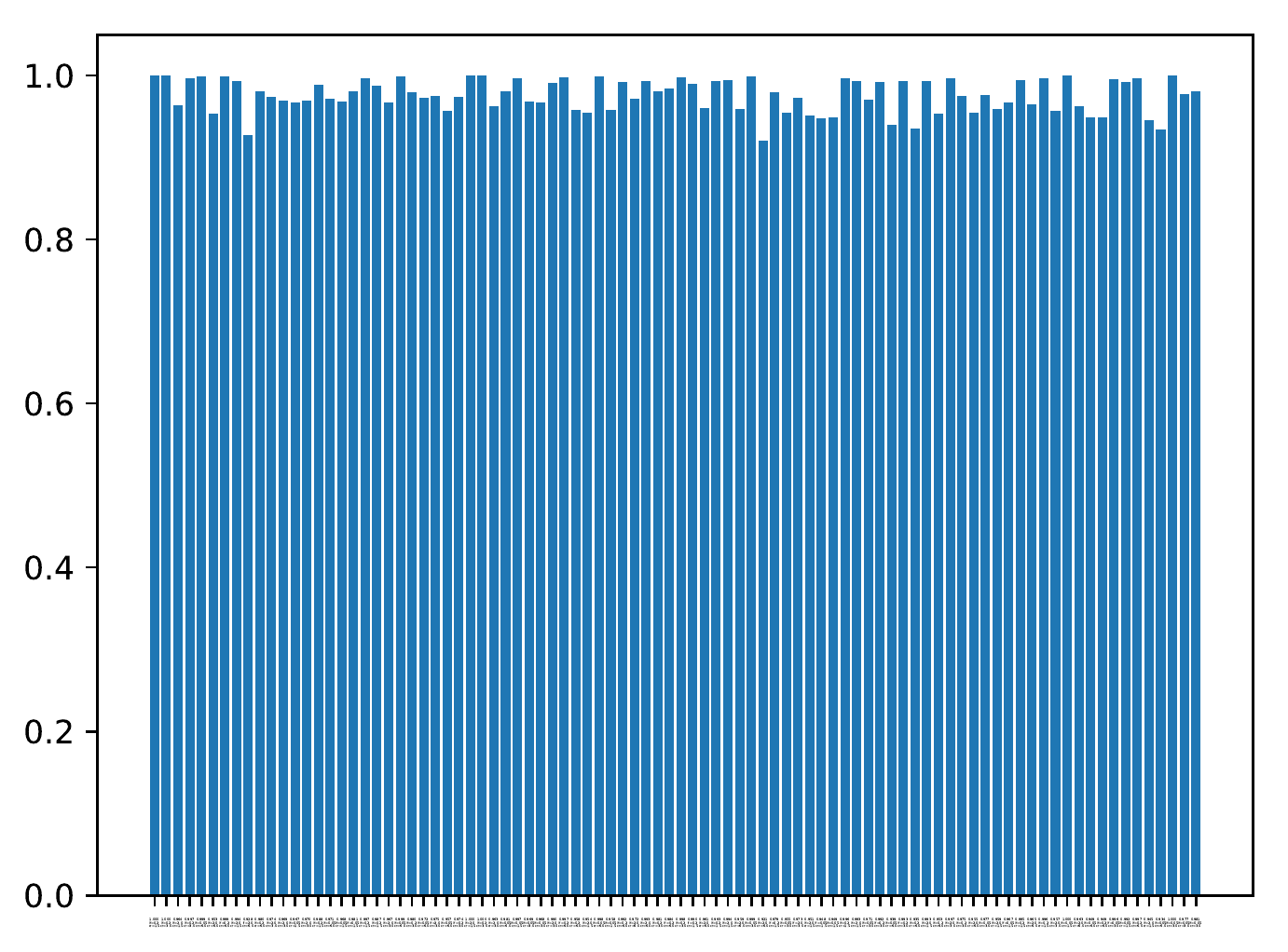}
    \caption{Modularity scores of InfoGAN-CR models (\teapots{} dataset). The models are sorted according to our model selection score.}
    \label{fig:modu_bar_teapots_infogancr_ours}
\end{figure}

\begin{figure}[t]
    \centering
    \includegraphics[width=0.5\linewidth]{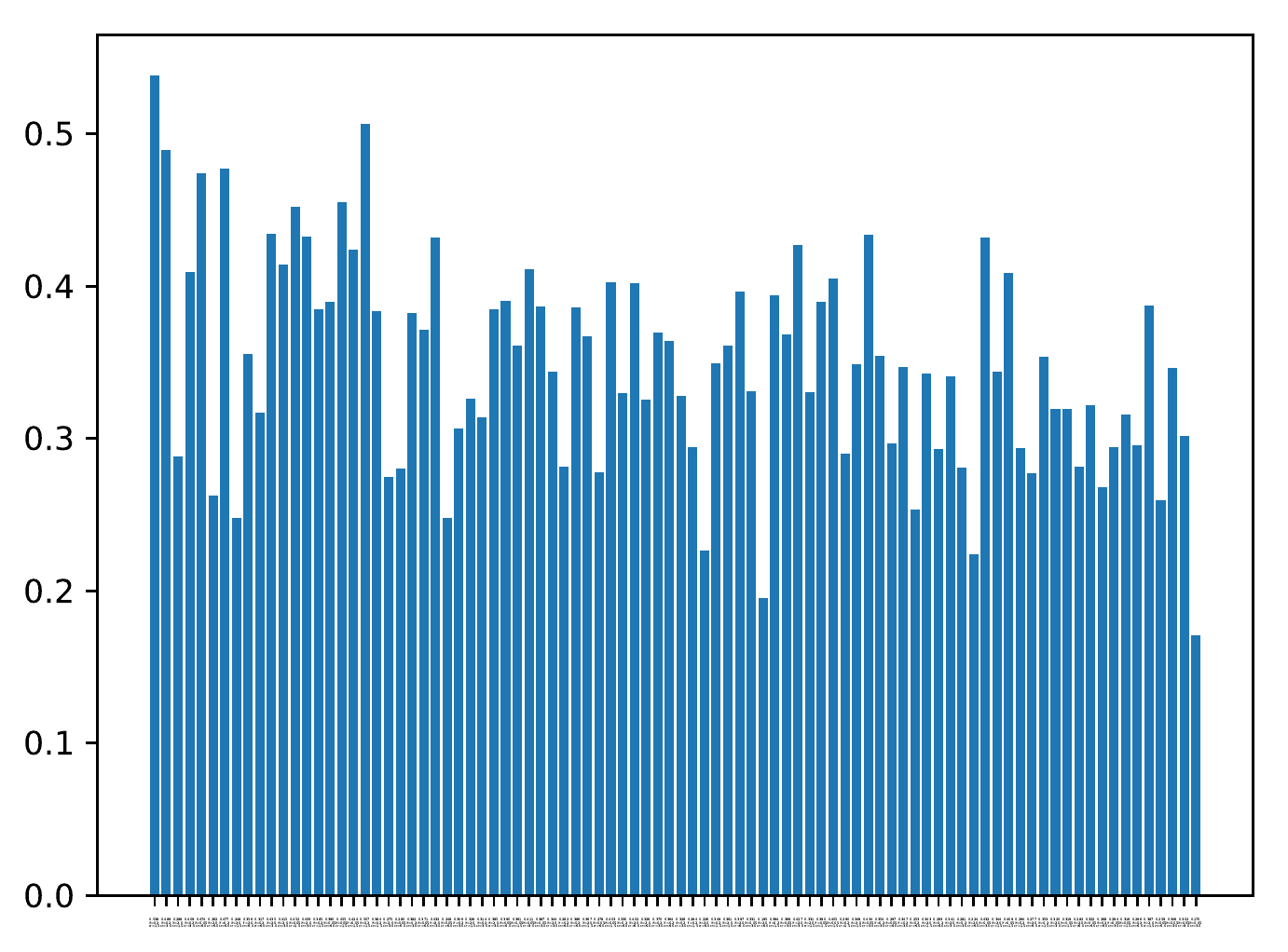}
    \caption{MIG scores of InfoGAN-CR models (\teapots{} dataset). The models are sorted according to our model selection score.}
    \label{fig:mig_bar_teapots_infogancr_ours}
\end{figure}

\begin{figure}[t]
    \centering
    \includegraphics[width=0.5\linewidth]{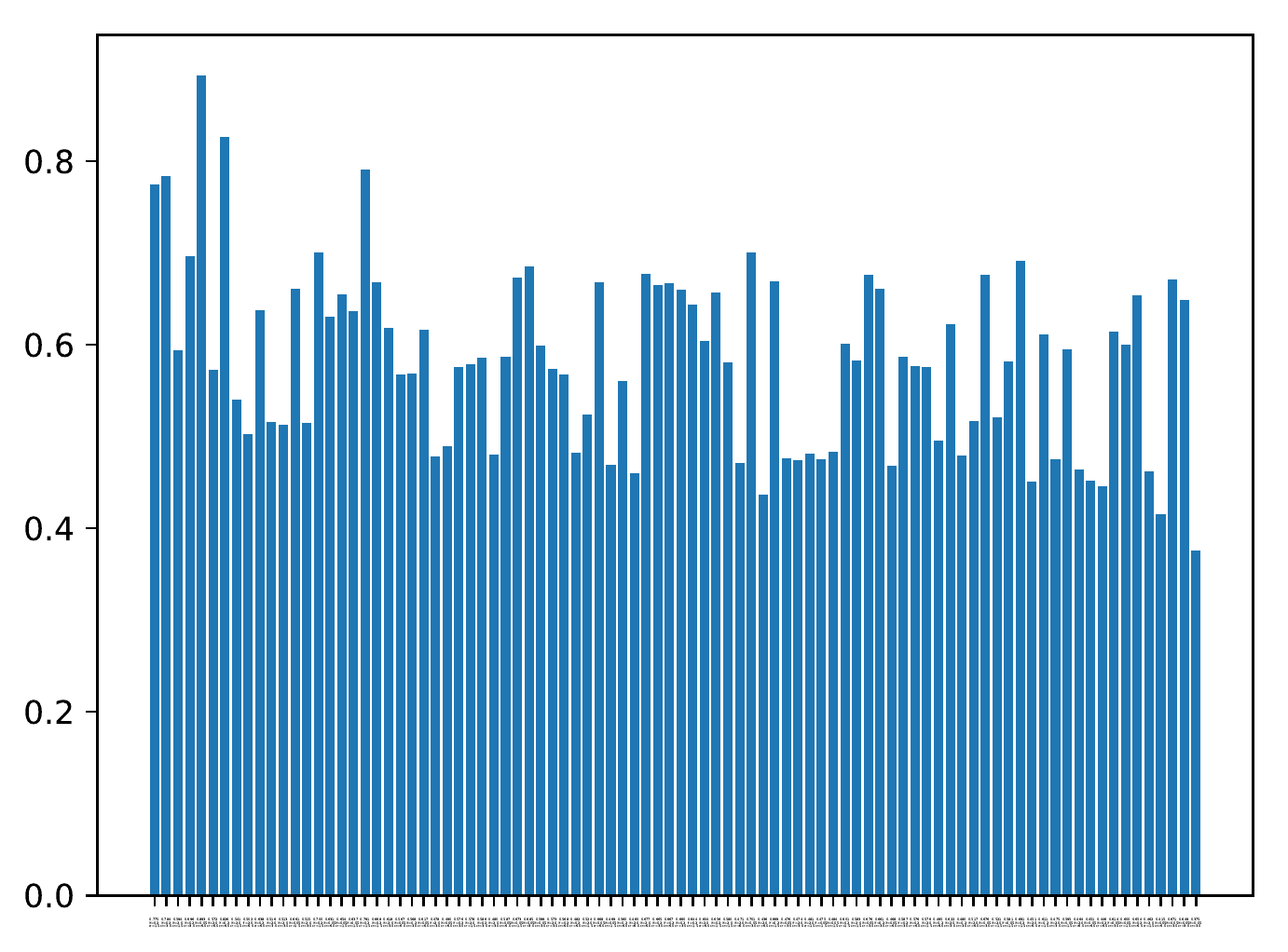}
    \caption{SAP scores of InfoGAN-CR models (\teapots{} dataset). The models are sorted according to our model selection score.}
    \label{fig:sap_bar_teapots_infogancr_ours}
\end{figure}

Figure \ref{fig:factorvae_bar_teapots_infogancr_ours}, \ref{fig:betavae_bar_teapots_infogancr_ours}, \ref{fig:dci_bar_teapots_infogancr_ours},\ref{fig:expl_bar_teapots_infogancr_ours},\ref{fig:modu_bar_teapots_infogancr_ours},\ref{fig:mig_bar_teapots_infogancr_ours},\ref{fig:sap_bar_teapots_infogancr_ours} show the bar plots of supervised metrics, where the models are ordered according to our model selection approach. These figures show that our model selection approach roughly correlates with other supervised metrics, but is more noisy than the results in \dsprites{}. Again, the reason might be that this dataset is more difficult.

\clearpage
\end{document}